\ifdefined\useorstyle

\documentclass[moor,sglanonrev]{informs4}
\RequirePackage{tgtermes}
\RequirePackage{newtxtext}
\RequirePackage{newtxmath}
\RequirePackage{bm}
\RequirePackage{endnotes}

\usepackage{macros/packages-or}
\usepackage{macros/editing-macros}
\usepackage{macros/statistics-macros-or}
\usepackage{macros/formatting}

\DoubleSpacedXI 

\usepackage{endnotes}
\let\footnote=\endnote

\usepackage[sort&compress]{natbib}
 \bibpunct[, ]{[}{]}{,}{n}{}{,}%
 \def\bibfont{\small}%
 \def\bibsep{\smallskipamount}%
\TheoremsNumberedThrough     
\ECRepeatTheorems

\EquationsNumberedThrough    

\usepackage{hyperref}

\usepackage{pgfplotstable}
\usepackage{graphicx}
\usepackage{subcaption}
\usepackage{float} 
\usepackage{tabularx}
\usepackage{overpic}
\usepackage{tikz}
\usepackage{rotating}
\usepackage{psfrag}
\usepackage{enumitem}
  
\begin{document}



\RUNAUTHOR{Li, Mittal, Namkoong, and Xia}
\RUNTITLE{Evaluating Model Performance Under Worst-case Subpopulations}

\TITLE{Evaluating Model Performance Under Worst-case Subpopulations }

\ARTICLEAUTHORS{%

\AUTHOR{Mike Li}
\AFF{Decision, Risk, and Operations Division, Columbia Business School, New York, NY 10027, \EMAIL{mli24@gsb.columbia.edu}} 
\AUTHOR{Daksh Mittal}
\AFF{Decision, Risk, and Operations Division, Columbia Business School, New York, NY 10027, \EMAIL{DMittal27@gsb.columbia.edu}} 

\AUTHOR{Hongseok Namkoong}
\AFF{Decision, Risk, and Operations Division, Columbia Business School, New York, NY 10027, \EMAIL{namkoong@gsb.columbia.edu}} 

\AUTHOR{Shangzhou Xia}
\AFF{Decision, Risk, and Operations Division, Columbia Business School, New York, NY 10027, \EMAIL{sxia24@gsb.columbia.edu}} 

} 

\ABSTRACT{The performance of ML models degrades when the training population is
different from that seen under operation. Towards assessing distributional
robustness, we study the worst-case performance of a model over \emph{all}
subpopulations of a given size, defined with respect to core attributes
$Z$. This notion of robustness can consider arbitrary (continuous) attributes
$Z$, and automatically accounts for complex intersectionality in disadvantaged
groups.  We develop a scalable yet principled two-stage estimation procedure
that can evaluate the robustness of state-of-the-art models. We prove that our
procedure enjoys several finite-sample convergence guarantees, including
\emph{dimension-free} convergence. Instead of overly conservative notions
based on Rademacher complexities, our evaluation error depends on the
dimension of $Z$ only through the out-of-sample error in estimating the
performance conditional on $Z$. On real datasets, we demonstrate that our
method certifies the robustness of a model and prevents deployment of
unreliable models.


}



\maketitle

%


\else

\documentclass[11pt]{article}
\usepackage[numbers]{natbib}
\usepackage{macros/packages}
\usepackage{macros/editing-macros}
\usepackage{macros/formatting}
\usepackage{macros/statistics-macros}


\begin{document}

\abovedisplayskip=8pt plus0pt minus3pt
\belowdisplayskip=8pt plus0pt minus3pt


\begin{center}
  {\LARGE  Evaluating Model Performance Under Worst-case Subpopulations} \\
  \vspace{.5cm}
  {\Large Mike Li ~~~ Daksh Mittal ~~~ Hongseok Namkoong ~~~
    Shangzhou Xia} \\
  \vspace{.2cm}
  {\large Decision, Risk, and Operations Division, Columbia Business School} \\
  \vspace{.2cm}
  \texttt{\{mli24, dmittal27, namkoong, sxia24\}@gsb.columbia.edu}
\end{center}


\begin{abstract}%
  
\end{abstract}

\fi

\section{Introduction}
\label{section:introduction}

Organizations increasingly deploy machine learning (ML) models to automate
decisions, yet these models often underperform when the operational
environment differs from the training environment. Model performance has been
observed to substantially degrade under distribution
shifts~\citep{BlitzerMcPe06, DaumeMa06, SaenkoKuFrDa10, TorralbaEf11,
  KohSaEtAl20} in domains ranging from healthcare
delivery~\citep{KoeneckeEtAl20} and financial services~\citep{AmorimCaVe18} to
environmental monitoring~\citep{BeeryCoGj20}.  Heavily engineered commercial
models are no exception~\citep{BuolamwiniGe18}.

Biases in data collection is a particularly prominent cause of distribution
shift. Data forms the infrastructure on which we build prediction
models~\citep{DentonHaAmSmNiSc20}, and they embody socioeconomic and political
inequities. For example, out of 10,000+ cancer clinical trials the National
Cancer Institute funds, less than 5\% of participants were
non-white~\citep{ChenLaDaPaKe14}.  Models trained on biased data replicate and
perpetuate bias: their performance drops significantly on underrepresented
speech recognition systems work poorly for Blacks~\citep{KoeneckeEtAl20} and
those with minority accents~\citep{AmodeiAnAnBaBaCaCaCaChCh16}.  More
generally, model performance degrades across demographic attributes such as
race, gender, or age, in facial recognition, video captioning, language
identification, and academic recommender systems~\citep{GrotherQuPh10,
  HovySo15, BlodgettGrOc16, SapiezynskiKaWi17, Tatman17, BuolamwiniGe18}.

It is crucial to \emph{rigorously certify} model robustness prior to
deployment for these heuristic approaches to bear fruit and transform
consequential applications. Ensuring that models perform uniformly well across
subpopulations is simultaneously critical for reliability, fairness,
satisfactory user experience, and long-term business goals.  While
practitioners often evaluate model performance across pre-defined demographic
segments, this approach fails to capture the complex operational reality where
disadvantaged groups are determined by multiple interacting factors---a
phenomenon known as intersectionality. The most adversely affected are often
determined by a complex combination of variables such as race, income, and
gender~\citep{BuolamwiniGe18}. For example, performance on summarization tasks
varies across demographic characteristics and document specific traits such as
abstractiveness, distillation, and location and dispersion of
information~\citep{GoelRaViTaWuZhXiBaRe21}.

To address these challenges, we study the worst-case subpopulation performance
across \emph{all} subpopulations of a given size. This conservative notion of
performance evaluates robustness to unanticipated distribution shifts in
$\worstcov$, and automatically accounts for complex intersectionality by
virtue of being agnostic to demographic groupings.  Formally, let $\worstcov$
be a set of core attributes that we wish to guarantee uniform performance
over. These may include protected demographic variables such as race, gender,
income, age, or domain-specific information such as length of the prompt or
metadata on the input; notably, it can contain any continuous or discrete
variables. We let $X \in \mathcal{X}$ be the input / covariate, and $Y \in \mathcal{Y}$
be the label.  In NLP and vision applications, $X$ is high-dimensional and
typically $\mbox{dim}(\worstcov) \ll \mbox{dim}(X)$.

For a fixed prediction model $\theta(X)$ and loss $\loss(\theta(x); y)$, our
goal is to ensure that the model $\theta$ performs well over all
subpopulations defined over $\worstcov$. We evaluate model losses on a mixture
component, which we call a subpopulation.  Postulating a lower bound
$\alpha \in (0, 1]$ on the demographic proportion (mixture weight), we
consider the set of subpopulations of the data-generating distribution
$P_\worstcov$
\begin{equation}
  \label{eqn:subpopulations}
  \mathcal{Q}_\alpha \defeq \left\{Q_\worstcov \mid
    P_\worstcov = a Q_\worstcov + (1-a) Q_\worstcov'~\mbox{for some}~a
    \ge \alpha,~\mbox{and subpopulation}~Q_\worstcov'
  \right\}.
\end{equation}
The demographic proportion (mixture weight) $a$ represents how
underrepresented the subpopulation is under the data-generating distribution
$P_Z$.

Before deploying the model $\theta$, we wish to evaluate the worst-case
subpopulation performance
\begin{equation}
  \label{eqn:cvar}
  \worstsub\opt \defeq \sup_{Q_\worstcov \in \mathcal{Q}_{\alpha}} \E_{\worstcov \sim
    Q_\worstcov} \left[
    \E[ \loss(\theta(X), Y) \mid \worstcov]
  \right].
\end{equation}
The worst-case subpopulation performance~\eqref{eqn:cvar} guarantees uniform
performance over subpopulations~\eqref{eqn:subpopulations} and has a clear
interpretation that can be communicated to diverse stakeholders.  The minority
proportion $\alpha$ can often be chosen from first principles, e.g., we wish
to guarantee uniformly good performance over subpopulations comprising at
least $\alpha = 20\%$ of the collected data. 
Alternatively, it is often informative to study the threshold level of
$\alpha\opt$ when $\alpha \mapsto \worstsub\opt$ crosses the \emph{maximum
  level of acceptable loss}. The threshold $\alpha\opt$ provides a
\emph{certificate of robustness} on the model $\theta(\cdot)$, guaranteeing
that all subpopulations larger than $\alpha\opt$ enjoy good performance.

We provide a principled and scalable procedure for estimating the worst-case
subpopulation performance~\eqref{eqn:cvar} and the certificate of robustness
$\alpha\opt$. A key technical challenge is that for each data point, we
observe the loss $\loss(\theta(X); Y)$ but never observe the conditional risk
evaluated at the attribute $\worstcov$
\begin{equation}
  \label{eqn:cond-risk}
  \condrisk(\worstcov) \defeq \E[\loss(\theta(X); Y) \mid \worstcov].
\end{equation}
In Section~\ref{section:approach}, we propose a two-stage estimation approach
where we compute an estimate $\what{\model}(\cdot) \in \modelclass$ of the conditional risk
$\condrisk(\cdot)$. Then, we compute a \emph{debiased} estimate of the
worst-case subpopulation performance under $\what{\model}(\cdot)$ using a
dual reformulation of the worst-case problem~\eqref{eqn:cvar}. We show several
theoretical guarantees for our estimator of the worst-case subpopulation
performance~\eqref{eqn:cvar}.  In particular, our first finite-sample result
(Section~\ref{section:convergence}) shows convergence at the rate {\scriptsize
  $O_p\left( \sqrt{\mathfrak{Comp}_n(\modelclass) /n}\right)$}, where
$\mathfrak{Comp}_n$ denotes a notion of complexity for the model class
estimating the conditional risk~\eqref{eqn:cond-risk}.

In some applications, it may be natural to define $\worstcov$ using images or
natural languages describing the input and use deep networks to predict the
conditional risk~\eqref{eqn:cond-risk}.  As the complexity term
$\mathfrak{Comp}_n(\modelclass)$ becomes prohibitively large in this
case~\citep{BartlettFoTe17, ZhangBeHaReVi17}, our second result
(Section~\ref{section:dim-free}) shows data-dependent \emph{dimension-free}
concentration of our two-stage estimator: our bound only depends on the
complexity of the model class $\modelclass$ through the out-of-sample error
for estimating the conditional risk~\eqref{eqn:cond-risk}. This
error can be made  small using overparameterized deep networks, allowing
us to estimate the conditional risk~\eqref{eqn:cond-risk} using even the
largest deep networks and still obtain a theoretically principled upper
confidence bound on the worst-case subpopulation performance. 
Leveraging these guarantees, we develop principled procedures for estimating
the certificates of robustness $\alpha\opt$ in
Section~\ref{section:certificate}.

In Section~\ref{section:experiments}, we demonstrate the effectiveness of our
procedure on real data. By evaluating model robustness under subpopulation
shifts, our methods allow the selection of robust models before deployment as
we illustrate using the recently proposed CLIP model~\citep{RadfordKiEtAl21}.
Finally, we generalize the notion of worst-case subpopulation performance we
study in Section~\ref{section:extension}. We note that these measures in fact
form an equivalence with coherent risk measures and distributionally robust
losses that are classical in the OR/MS literature. At a high level, our result
uncovers a deeper connection between classical ideas in risk measures and the
more recent ML fairness literature.

\paragraph{Related work.}
Our notion of worst-case subpopulation performance is also related to the by
now vast literature on fairness in ML. We give a necessarily
abridged discussion and refer readers to~\citet{BarocasHaNa19}
and~\citet{Corbett-DaviesGo18} for a comprehensive treatment. A large body of
work studies \emph{equalizing} a notion of performance over fixed, pre-defined
demographic groups for \emph{classification tasks}~\citep{Chouldechova17a,
  FeldmanFrMoScVe15, BarocasSe16, HardtPrSr16, KleinbergMuRa16,
  WoodworthGuOhSr17}.~\citet{KearnsNeRoWu18, KearnsNeRoWu19, HebertKiReRo17}
consider finite subgroups defined by a structured class of functions over $Z$,
and study methods of equalizing performance across them. By contrast, our
approach instantiates Rawls’ theory of distributive justice~\citep{Rawls01,
  Rawls09}, where we consider the allocation of the loss $\loss(\cdot; \cdot)$
as a resource. Rawls' difference principle maximizes the welfare of the
worst-off group and provides incentives for groups to maintain the status
quo~\citep{Rawls01}. Similarly, \citet{HashimotoSrNaLi18} studied negative
feedback loops generated by user retention---they use a more conservative
notion of worst-case loss than ours---as poor performance on a currently
underrepresented user group can have long-term consequences.

The long line of works on distributionally robust
optimization (DRO) aims to \emph{train models} to perform well under
distribution shifts. Previous approaches considered finite-dimensional
worst-case regions such as constraint sets~\citep{DelageYe10, GohSi10,
  KohSaLi19} and those based on notions of distances for probability measures
such as $f$-divergences and likelihood ratios~\citep{Ben-TalHeWaMeRe13,
  WangGlYe15, BertsimasGuKa18, LamZh15, Lam19, MiyatoMaKoNaIs15, LaguelMaHa20,
  DuchiGlNa21, DuchiNa21, LaguelPiMaHa21}, Levy-Prokhorov~\citep{ErdouganIy06},
Wasserstein distances~\citep{EsfahaniKu18, Shafieezadeh-AbadehEsKu15,
  BlanchetKaZhMu17,BlanchetMu19, BlanchetKaMu19,LiHuSo19, GaoKl22,
  BlanchetKaZhMu17, VolpiNaSeDuMuSa18, BlanchetMuSi22, ChenLiQiXu22,
  GaoChKl17, Gao20, BlanchetHeMu20, GaoChKl24, WangSiBlZh23}, and integral
probability metrics based on reproducing kernels~\citep{StaibJe19, ZhuJiDiSc21,
  WangGaXi21, XuLeChXi24}. The distribution shifts considered in these
approaches are modeled after mathematical convenience and are often difficult
to interpret. As a result, optimizing models under worst-case performance
often results in overly conservative models and these approaches do not
currently scale to modern large-scale NLP or vision applications, as those
models could have hundreds of thousands of parameters, posing great
computational challenges when DRO methods are applied. In this work, we
approach distributional robustness from a different angle: instead of robust
training, we study the problem of \emph{evaluation} -- given a model, can we
certify its robustness properties in any way?

In particular, our work is most closely related to~\citet{DuchiHaNa20}, who
proposed algorithms for \emph{training} models with respect to the worst-case
subpopulation performance~\eqref{eqn:cvar}, a more ambitious goal than our
narrower viewpoint of \emph{evaluating} model performance pre-deployment.
Their (full-batch) training procedure requires solving a convex program with
$n^2$ variables per gradient step, which is often prohibitively
expensive. Furthermore, training with respect to the worst-case conditional
risk $\E[\loss(\theta(X); Y) \mid \worstcov]$ does not scale to deep networks
that can overfit to the training data~\citep{SagawaKoHaLi19}. By contrast, our
evaluation perspective aims to take advantage of the rapid progress in deep
learning. We build scalable evaluation methods that apply to arbitrary models,
which allows leveraging state-of-the-art engineered approaches for training
$\theta(\cdot)$. Our narrower focus on evaluation allows us to provide
convergence rates that scale advantageously with the dimension of $Z$,
compared to the nonparametric $O_p(n^{-1/d})$ rates for
training~\citep{DuchiHaNa20}. Recently,~\citet{JeongNa20} studied a similar
notion of worst-case subpopulation performance in causal inference.

As we note later, our worst-case subpopulation performance gives the usual
conditional value-at-risk for $\E[\loss(\theta(X); Y) \mid Z]$, a classical
tail risk measure. Tail-risk estimation has attracted great interest:
\citet{WozabalWo09, PflugWo10} derive asymptotic properties of plug-in
estimates of coherent, law-invariant risk functionals, and
\citet{BelomestnyKr12, GuiguesKrSh18} derive central limit results. A
distinguishing aspect of our work is the unobservability of the conditional
risk $\E[\loss(\theta(X),Y)\mid\worstcov]$, which necessitates a shift to a
\emph{semiparametric estimation} paradigm. To tackle this challenge, we derive
a debiased approach to tail-risk estimation and provide both asymptotic and
finite-sample convergence guarantees. Concurrent to an earlier conference
version of this work,~\citet{SubbaswamyAdSa21} study a similar problem and
propose another estimator different from ours. They claim their estimator is
debiased and cite recent work~\citep{JeongNa20} as inspiration; but to our
knowledge, their estimator is not debiased as they incorrectly apply
\citet{JeongNa20}'s main insights and we note important errors in their proof
(Section~\ref{section:estimation} for a detailed discussion). In addition to
this difference, all finite-sample convergence guarantees and connections to
coherent risk measures are new in this work.

Our work significantly expands on the earlier conference
version~\citep{anon21}, in additino to a complete revision for the OR/MS
audience (e.g., background on how state-of-the-art OpenAI models address
distribution shift).  We develop a debiased estimator instead of a plug-in
estimator, using tools from the semiparametric statistics literature to
correct for the first-order error in estimating the nuisance parameter
$z \mapsto \E[\loss(\theta(X); Y) \mid Z = z]$.  We derive a central limit
result for our debiased estimator in Section~\ref{section:asymptotics},
showing that it is possible to achieve standard $\sqrt{n}$-rates of
convergence even when the fitted $z \mapsto \what{\mu}(z)$ converge at a
slower $n^{-1/3}$ rate.  Our new result allows computing confidence intervals
for the worst-case subpopulation performance. We also extend our prior
finite-sample concentration guarantees over uniformly bounded losses to
heavy-tailed losses in Section~\ref{sec:heavy-tail}. Finally, we propose a
natural extension of our worst-case subpopulation performance where we allow
the subpopulation proportion $\alpha$ to be stochastic. We connect this
generalized worst-case subpopulation performance to the vast literature on
distributional robustness and coherent risk measures in
Section~\ref{section:extension}.



\section{Methodology}
\label{section:approach}

\begin{figure}[t]
\centering
\includegraphics[width=\textwidth]{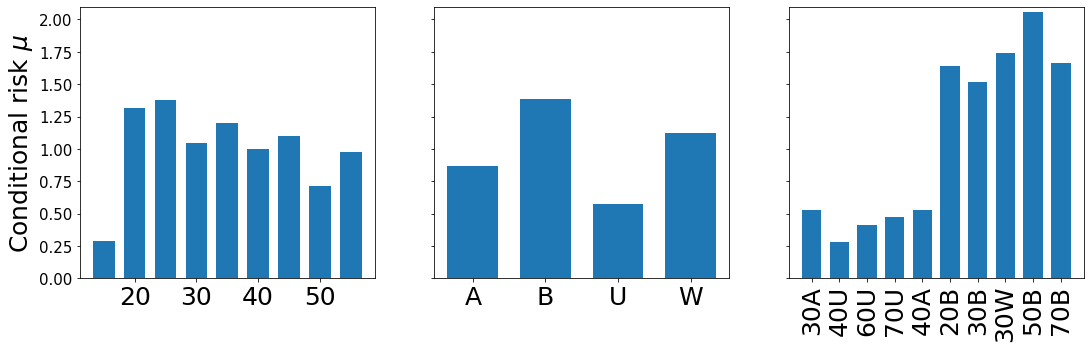}
     \caption{Conditional risk
       $\condrisk(\worstcov) = \E[(Y - \theta(X))^2 \mid \worstcov]$. Here
       $Z$ = \emph{age} on the left panel, $Z$ = \emph{race} in the center, and $Z$ = (\emph{age}, \emph{race}) on the right. A = Asian, B = Black, U = Unknown, W = White.}
     \label{fig:intersectionality-warfarin}
\end{figure}

We begin by contrasting our approach to standard alternatives that consider
pre-defined, fixed demographic
groups~\citep{MitchellWuZaBaVaHuSpRaGe19}. Identifying disadvantaged subgroups
a priori is often challenging as they are determined by \emph{intersections}
of multiple demographic variables.
To illustrate such complex intersectionality, consider a drug dosage
prediction problem for Warfarin~\citep{IWPC09}, a common anti-coagulant (blood
thinner). Taking the best prediction model for the optimal dosage on this
dataset based on genetic, demographic and clinical factors~\citep{IWPC09}, we
present the squared error on the root dosage. In
Figure~\ref{fig:intersectionality-warfarin}, when age and race are considered
\emph{simultaneously} instead of \emph{separately}, subpopulation performances
vary significantly across intersectional groups.

The worst-case subpopulation performance~\eqref{eqn:cvar} automatically
accounts for latent intersectionality.  It is agnostic to demographic
groupings and allows considering infinitely many subpopulations that represent
at least $\alpha$-fraction of the training population $P$. By allowing the
modeler to select arbitrary protected attributes $\worstcov$, we are able to
consider potentially complex subpopulations. For example, $\worstcov$ can even
be defined with respect to a natural language description of the input
$X$. The choice of $\worstcov$---and subsequent worst-case subpopulation
performance~\eqref{eqn:cvar} of the conditional risk
$\condrisk(\worstcov) = \E[\loss(\theta(X); Y) \mid \worstcov]$---interpolates
between the most conservative notion of subpopulations (when
$\worstcov = (X, Y)$) and simple counterparts defined over a single variable.

The choice of the subpopulation size $\alpha$ should be informed by domain
knowledge---desired robustness of the system---and the dataset size relative
to the complexity of $\worstcov$. Often, proxy groups can be used for
selecting $\alpha$. If we wish to ensure good performance over patients of all
races aged 50 years or older, we can choose $\alpha$ to be the proportion of
the least represented $(race, age \ge 50)$ group---this leads to
$\alpha = 5\%$ in the Warfarin data. The corresponding worst-case
subpopulation performance~\eqref{eqn:cvar} guarantees good performance over
all groups of similar size.

When it is challenging to commit to a specific subpopulation size, it may be
natural to postulate a \emph{maximum level of acceptable loss $\thresh$}. To
measure the robustness of a model, we define the smallest subpopulation size
$\alpha\opt$ for which the worst-case subpopulation performance is
acceptable
\begin{equation}
  \label{eqn:certificate}
  \alpha\opt \defeq \inf \{\alpha: \worstsub\opt \le \thresh\}.
\end{equation}
This provides a \emph{certificate of robustness}: if $\alpha\opt$ is large,
then $\theta$ is brittle against even majority subpopulations; if it is
sufficiently small, then the model $\theta(X)$ performs well on
underrepresented subpopulations.

We now derive estimators for the worst-case subpopulation
performance~\eqref{eqn:cvar} and the certificate of
robustness~\eqref{eqn:certificate}, based on i.i.d.\ observations
$(X_i, Y_i, Z_i)_{i=1}^n \sim P$. We assume our observations are independent
from the data used to train the model $\theta(\cdot)$.

\paragraph{Dual reformulation} The worst-case subpopulation
performance~\eqref{eqn:cvar} is unwieldy as it involves an infinite
dimensional optimization problem over probabilities. Instead, we use its dual
reformulation for tractable estimation.

We denote $\hinge{\cdot} =\max(\cdot, 0)$, and denote by $\worstsub(\model)$
the worst-case subpopulation performance for a function $\model(\worstcov)$
(so that $\worstsub\opt = \worstsub(\condrisk)$). Let $\aq{\model}$ denote the
$(1-\alpha)$-quantile of $\model(\worstcov)$.
\begin{lemma}[{\citet[Theorem 6.2]{ShapiroDeRu14} and~\citet{RockafellarUr00}}]
  \label{lemma:dual}
  If $\E[\model(\worstcov)_+] < \infty$, then for $\alpha\in(0,1)$,
  \begin{align}
    \label{eqn:dual}
    \worstsub(\model) \defeq \sup_{Q_{\worstcov} \in \mixdist} 
    \E_{\worstcov \sim Q_{\worstcov}} \left[ \model(\worstcov) \right]
    = \inf_{\eta \in \R} \left\{
    \frac{1}{\alpha} \E_P\hinge{\model(\worstcov) - \eta} + \eta
    \right\}
   = \frac{1}{\alpha} \int_0^\alpha P_{1-t}^{-1}(\model)\,\mathrm{d}t.
  \end{align}
 The infimum is attained at $\eta = P_{1-\alpha}^{-1}(\mu)$. Moreover, if $\model(\worstcov)$ has no probability mass at $P_{1-\alpha}^{-1}(\model)$, then $\worstsub(\model) = \E[\model(\worstcov)\,|\, \model(\worstcov) \ge P_{1-\alpha}^{-1}(\model)]$.
\end{lemma}
\noindent The dual~\eqref{eqn:dual} shows $\worstsub\opt$ is a
tail-average of $\condrisk(\worstcov)$, a popular risk measure known as the
conditional value-at-risk (CVaR) in portfolio
optimization~\citep{RockafellarUr00}. The dual optimum is attained at the
$(1-\alpha)$-quantile of the $\condrisk(\worstcov)$~{\cite[Theorem
  10]{RockafellarUr02}}, giving the worst-case subpopulation
\begin{equation}
  \label{eqn:thr}
  \pthr(\worstcov) = \frac{1}{\alpha} \indic{\condrisk(\worstcov) \ge \aq{\condrisk}}.
\end{equation}

\section{Estimation}
\label{section:estimation}

A key challenge in estimating $\worstsub\opt$
is that we can only observe losses $\loss(\theta(X_i); Y_i)$ and never observe
the conditional risk $\condrisk(\cdot)$~\eqref{eqn:cond-risk}. To estimate
$\condrisk(\cdot)$, we can solve an empirical approximation to the loss
minimization problem
\begin{equation}
  \label{eqn:first-pop}
  \minimize_{\model \in \modelclass} \quad \E\left[ \left( \loss(\theta(X); Y) - \model(\worstcov)
    \right)^2 \right]
\end{equation}
for some model class $\mathcal{H}$ (class of mappings $\mathcal{\worstcov} \to \R$). The
loss minimization formulation~\eqref{eqn:first-pop} allows the use of any machine learning
estimator, as well as standard tools for model selection (e.g. cross
validation). Denoting by $\bestmodel$ a minimizer of~\eqref{eqn:first-pop}, we
may have
$\bestmodel(\cdot) \neq \condrisk(\cdot) = \E[\loss(\theta(X); Y) \mid
\cdot]$ if the model class $\modelclass$ is not sufficiently rich. In the
following section, we provide guarantees that scale with the misspecification
error $\bestmodel - \condrisk$.

\paragraph{Plug-in estimator}
As $\condrisk(\cdot)$ must be estimated, this yields a \emph{semiparametric}
tail-risk estimation problem: $\condrisk(\cdot)$ can be a complex
nonparametric function, but ultimately we are interested in evaluating a
one-dimensional statistical functional $\worstsub(\condrisk)$.  A natural
plug-in approach is to split the data into auxiliary and main samples, where
we first estimate $\what{\model}$ using a sample average approximation of the
problem~\eqref{eqn:first-pop} on the auxiliary sample $S_1$.  On the main
sample $S_2$, we can estimate the worst-case subpopulation performance using
the dual
\begin{equation}
  \label{eqn:empirical-cvar}
  \what{\mathsf{W}}_{\alpha}(\model)
  \defeq 
  \inf_{\eta}
  \left\{ \frac{1}{\alpha |S_2|}  \sum_{i \in S_2}  
    \hinge{\model(\worstcov) - \eta} + \eta  \right\}.
\end{equation}
The final plug-in estimator is given by
$\what{\mathsf{W}}_{\alpha}(\what{\model})$.

\paragraph{Debiasing the plug-in estimator} The plug-in estimator is suboptimal since it does not take into account the
potential error incurred by using the approximation $\what{\model}$ in the
final step. To build intuition on this semiparametric error, we provide a
heuristic analysis of $\worstsub(\what{\model}) - \worstsub(\condrisk)$ using
a first-order Taylor expansion, ignoring the statistical error incurred in the
second stage for a moment. Abusing notation, the functional
$P \mapsto \worstsub(P) \defeq \worstsub(\E_{P}[\loss(\theta(X); Y)
\mid \worstcov])$ can be seen to be suitably differentiable so that there is a
continuous linear map $\dot{\worstsub}$ on the space of square integrable functions such that
\begin{align*}
  \frac{d}{dr} \left( \worstsub(P + r(\bar{P} -P)) -
  \worstsub(P) \right) = \dot{\worstsub}(\bar{P} - P).
\end{align*}
By the Riesz Representation Theorem, there is a function
$\nabla \worstsub(X, Y, Z; P)$ such that
\begin{equation*}
  \dot{\worstsub}(\bar{P} -P) = \int \nabla \worstsub(X, Y, Z; P)  d(\bar{P} - P),
\end{equation*}
where we assume $\nabla \worstsub$ has mean zero without loss of generality; the
random variable $\nabla \worstsub$ is often referred to as the pathwise derivative or
the efficient influence function of the functional $\worstsub$~\citep{Newey94}.

Under appropriate regularity conditions, we can apply the chain rule for
influence function calculus~\citep{Newey94, Kennedy22} and use Danskin's
theorem~\citep[Theorem 4.13]{BonnansSh00} to calculate functional
gradients. Noting that the dual optimum is attained at the
$(1-\alpha)$-quantile $\aq{\model_P}$ by first order conditions, we have
\begin{align*}
  \nabla \worstsub(X, Y, Z; P) & = \frac{1}{\alpha} \hinge{\model_P(\worstcov) - \aq{\model_P}}
  - \frac{1}{\alpha} \E_P[\hinge{\model_P(\worstcov) - \aq{\model_P}}] \\
  & \qquad + \pthr(\worstcov) (\loss(\theta(X); Y) - \model_P(\worstcov))
\end{align*}
where we use
\begin{align*}
  \model_P(\worstcov)  \defeq \E_P[\loss(\theta(X); Y) \mid \worstcov] ~~~\mbox{and}~~~
  \tau_P(\worstcov)  \defeq \alpha^{-1} \indic{\model_P(\worstcov) \ge \aq{\model_P}}.
\end{align*}
Taking a
first-order Taylor expansion around the learned parameter $\what{\model}$, we
arrive at
\begin{align}
  \label{eqn:taylor}
  \worstsub(\what{\model}) - \worstsub(\condrisk)
  = - \E_P[ \what{\tau}(\worstcov) (\loss(\theta(X); Y) - \what{\model}(\worstcov))]
  + \mathsf{Rem}_2
\end{align}
where $\hthr(\cdot)$ is an estimator of the worst-case
subpopulation~\eqref{eqn:thr} and $\mathsf{Rem}_2$ is a second-order remainder
term.

The Taylor expansion~\eqref{eqn:taylor} provides a natural approach to
correcting the first-order error of the plug-in estimator---a standard
approach called \emph{debiasing} in the semiparametric statistics
literature~\citep{Neyman59, Newey94, ChernozhukovChDeDuHaNeRo18}. Instead of
the plug-in~\eqref{eqn:empirical-cvar}, the \emph{debiased} estimator on the
main sample $S_2$ is given by
\begin{equation*}
  \inf_{\eta}
  \left\{ \frac{1}{\alpha |S_2|}  \sum_{i \in S_2}  
    \hinge{\what{\model}(\worstcov) - \eta} + \eta  \right\}
  + \frac{1}{|S_2|} \sum_{i \in S_2} \what{\tau}(\worstcov_i) (\loss(\theta(X_i); Y_i)
  - \what{\model}(\worstcov_i)).
\end{equation*}
The the debiased estimator automatically achieves a second-order error and as we
show in Section~\ref{section:asymptotics}, it achieves parametric rates of
convergence even when $\what{\model}$ converges more slowly.


\paragraph{Cross-fitting procedure} To utilize the entire sample, we take a
cross-fitting approach~\citep{ChernozhukovChDeDuHaNeRo18} where we partition
the data into $K$ folds. We assign a single fold as the main data and use the
rest as auxiliary data, switching the role of the main fold to get $K$
separate estimators; the final estimator is simply the average of the $K$
versions as outlined in Algorithm~\ref{alg:two-stage}. We take
$\hthr(z) \defeq \frac{1}{\alpha} \indic{\hmu(z) \ge \what{q}}$, where
$\what{q}$ is an estimator of $\aq{\hmu}$ based on the auxiliary data. The
quantile estimator $\what{q}$ can be computed using \emph{unsupervised}
observations $\worstcov$, which is typically cheap to collect.  To estimate
the threshold subpopulation size $\alpha\opt$, we simply take the cross-fitted
version of the plug-in estimator
\begin{equation}
  \label{eqn:certificate-emp}
  \what{\alpha}_{\indfold} \defeq \inf \{\alpha: \hworstsub(\what{\model}_\indfold) \le \thresh\}.
\end{equation}
Since $\alpha \mapsto \hworstsub(\what{\model}_\indfold)$ is decreasing, the
threshold can be efficiently found by a simple bisection search.

In Section~\ref{section:asymptotics}, we show that our cross-fitted augmented
estimator $\what{\omega}_{\alpha}$ has asymptotic variance
\begin{align}
  \label{eqn:var}
  \sigma^2_{\alpha}
  & \defeq \frac{1}{\alpha^2} \var\left(\hinge{\pmu(\worstcov) - \aq{\pmu}} \right)
    + {\color{blue}}\var\left( \pthr(\worstcov) (\loss(\theta(X); Y) - \pmu(\worstcov)) \right).
\end{align}
In particular, letting $z_{\delta}$ be the $(1 - \delta/2)$-quantile of a
standard normal distribution, our result gives confidence intervals with
asymptotically exact coverage
$\P(\worstsub\opt \in [\what{\omega}_{\alpha} \pm z_{\delta}
\what{\sigma}_{\alpha} / \sqrt{n}]) \to 1-\delta$.

\begin{algorithm}[hbtp]
  \caption{\label{alg:two-stage}
    Cross-fitting  procedure for estimating worst-case subpopulation performance~\eqref{eqn:cvar}}
  \begin{algorithmic}[1]
    \State \textsc{Input:} Subpopulation size $\alpha$, model class $\mathcal{H}$,
    $K$-fold partition $\cup_{k=1}^K I_k = [n]$ of
     $\{(X_i, Y_i, Z_i)\}_{i=1}^n$ s.t. $|I_k| = \frac{n}{K}$
   \State \textbf{\textsc{For}} $\indfold \in [K]$
   \State \hspace{10pt} \textbf{Estimate nuisance parameters}      \label{eqn:first} 
   Using the data $\{(X_i, Y_i, Z_i)\}_{i \in I_k^c}$, fit estimators 
   \State \hspace{10pt} 1.  Solve
   $\what{\model}_\indfold \in \argmin_{\model \in \modelclass}  \sum_{i \in \cfold}
    \left( \loss(\theta(X_i); Y_i) - \model(\worstcov_i) \right)^2$.
      \State \hspace{10pt} 2. $\hthr(z) \defeq \frac{1}{\alpha} \indic{\hmu(z) \ge \hq}$, where
      $\hq$ is an estimator of $\aq{\what{\model}_\indfold}$ on $\cfold$
    \State \hspace{10pt} \textbf{Compute augmented estimator} Using the data \label{eqn:second}
    $\{(X_i, Y_i, Z_i) \}_{i \in I_k}$, compute
    \begin{align*}
      \what{\omega}_{\alpha, k}
      & \defeq \inf_{\eta}
        \left\{ \frac{1}{\alpha} \E_{\worstcov \sim \empfold}
        \hinge{\hmu(\worstcov) - \eta} + \eta  \right\}
        + \E_{(X,Y,Z)\sim\empfold}\left[\hthr(\worstcov) (\loss(\theta(X); Y) - \what{\model}_{\indfold}(Z))\right] \\
      \what{\sigma}^2_{\alpha, k}
      & \defeq \frac{1}{\alpha^2} \var_{\worstcov \sim \empfold} \hinge{\hmu(\worstcov) - \hq}
        + \var_{(X,Y,Z)\sim\empfold} \left( \hthr(\worstcov) (\loss(\theta(X); Y) - \what{\model}_{\indfold}(\worstcov) ) \right)
    \end{align*}
    \State \textbf{\textsc{Return}} Estimator
    $\what{\omega}_{\alpha} = \frac{1}{K} \sum_{k \in [K]}
    \what{\omega}_{\alpha, k}$, and variance estimate
    $\what{\sigma}^2_{\alpha} = \frac{1}{K} \sum_{k \in [K]}
    \what{\sigma}^2_{\alpha, k}$
  \end{algorithmic}
\end{algorithm}

{
\paragraph{Empirical comparison between plug-in and debiased estimators}
To quantify the practical value of debiasing, we simulate the worst-case
subpopulation risk of a squared-error loss under a classical data-generating
process used in the causal inference literature. Following the example
constructed by~\citet{KangSc07}, we consider latent covariates
$\xi \sim \mbox{N}(0, I_{20})$ and outcomes
$Y = 210 + 27.4 \xi_1 + 13.7(\xi_2 + \xi_3 + \xi_4) + \varepsilon$ with
$\varepsilon \sim \mbox{N}(0,1)$.  The analyst observes nonlinear
transformations $X = g(\xi)$ of these covariates and fits the prediction rule
$\theta(X) = \theta^\top X$ using a fixed draw of
$\theta \sim \mbox{N}(0, 0.5^2 I_{20})$. We estimate the nuisance regression
$\mu(\worstcov) = \E[\loss(\theta(X); Y) \mid \worstcov]$ with an XGBoost
regressor. To remove overfitting bias, we compute out-of-fold predictions via
three-fold cross-fitting and compute both the plug-in and debiased estimators
on each held-out fold. The true worst-case subpopulation risk is approximated
using an independent sample of size $5 \times 10^4$. We vary the sample size
from $n = 10^2$ to $10^5$ and repeat each configuration $100$ times with a
target subpopulation mass of $\alpha = 0.2$, recording point estimates,
variance estimates, and nominal $90\%$ confidence intervals. The resulting
performance summaries are visualized in Figure~\ref{fig:mse-simulations} and
Figure~\ref{fig:bias-variance-simulations}.

\begin{figure*}[t]
  \centering
  \begin{subfigure}[t]{0.48\textwidth}
    \centering
    \includegraphics[width=\textwidth]{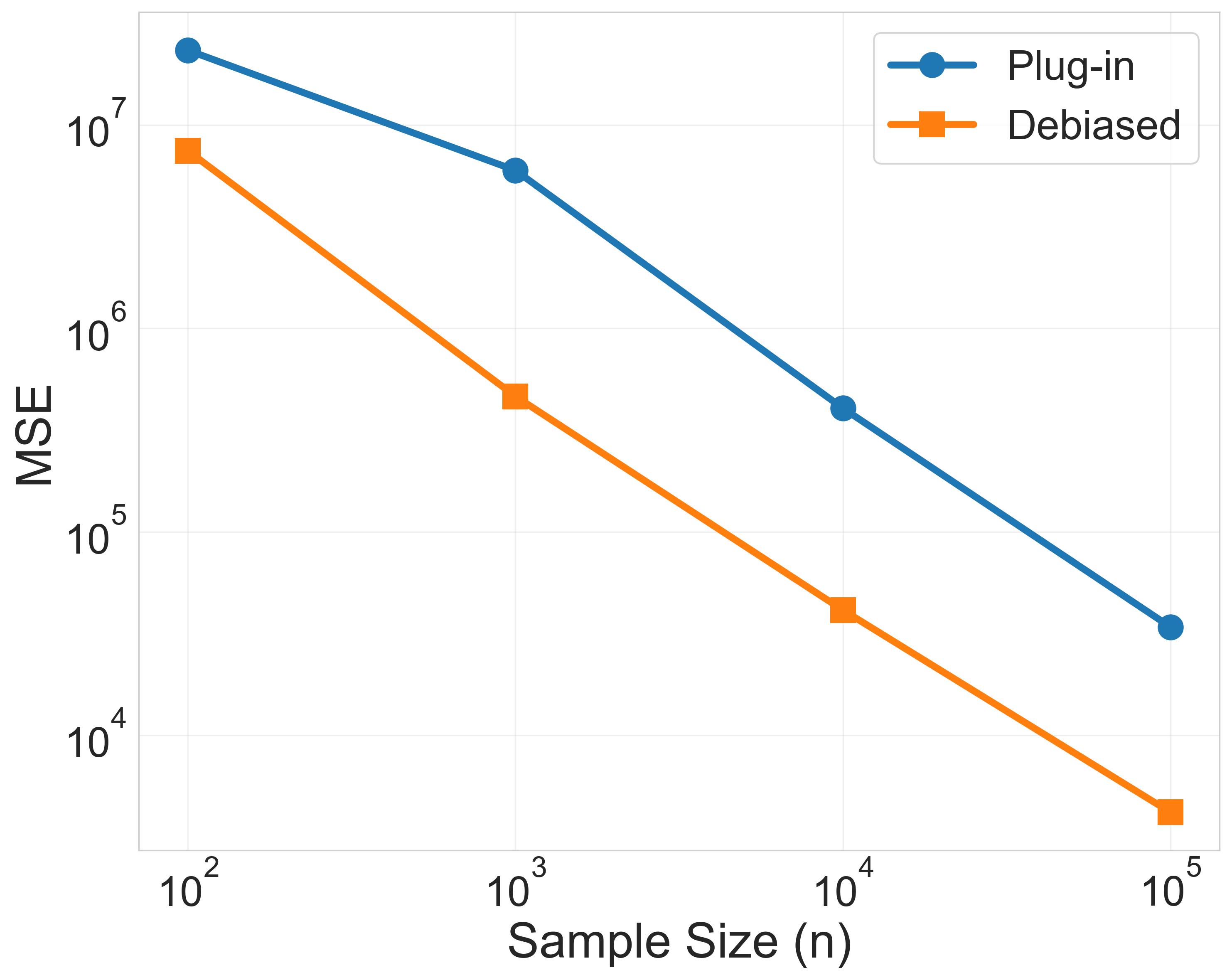}
    \caption{Mean squared error by sample size}
    \label{fig:mse-simulations-a}
  \end{subfigure}
  \hfill
  \begin{subfigure}[t]{0.48\textwidth}
    \centering
    \includegraphics[width=\textwidth]{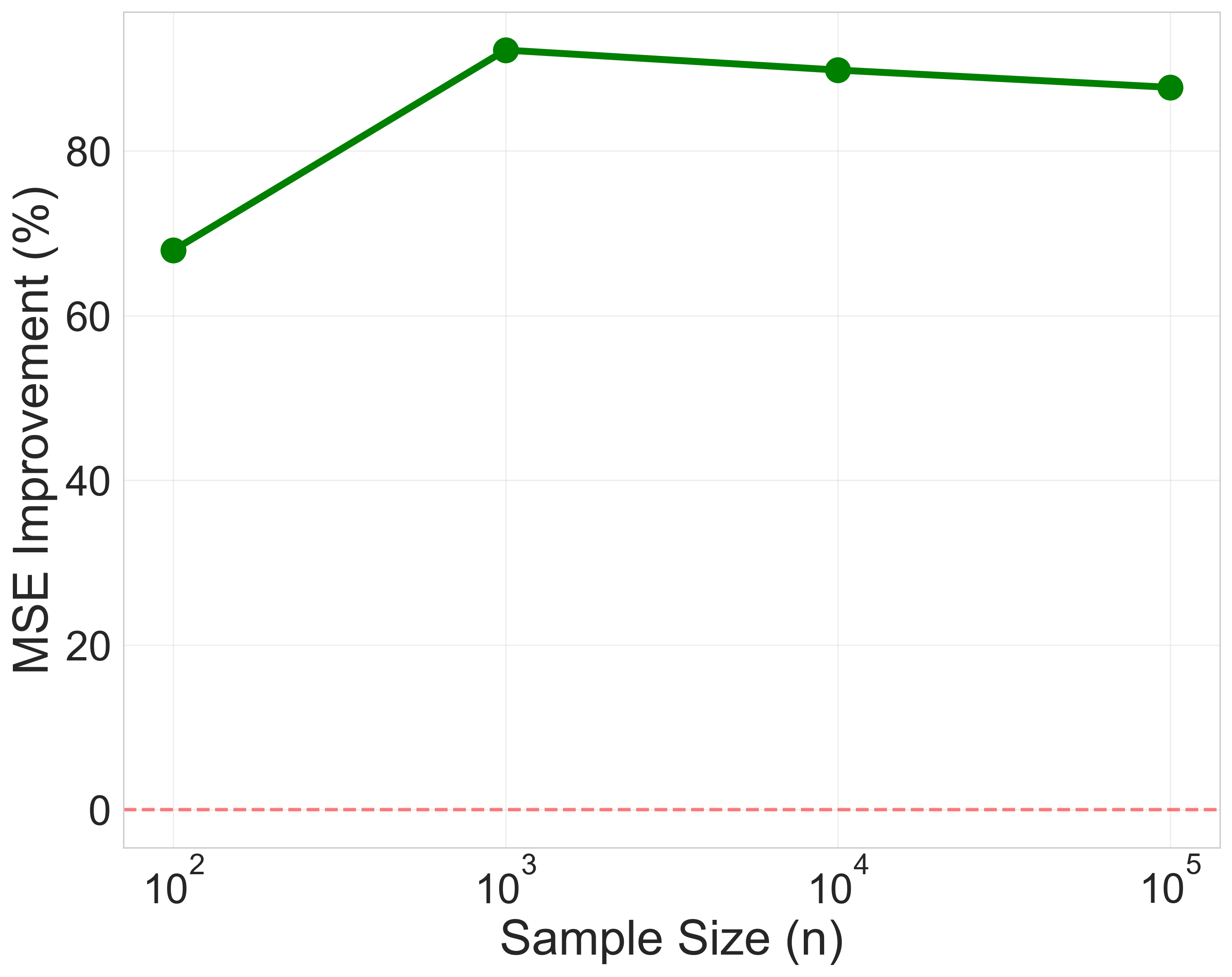}
    \caption{Relative MSE reduction over plug-in}
    \label{fig:mse-simulations-b}
  \end{subfigure}
  \caption{Debiasing estimator yields substantially lower MSE and rapidly
    gains relative efficiency as $n$ grows. Improvement is reported as
    $(\text{MSE}_\text{plug-in} -
    \text{MSE}_\text{debiased})/\text{MSE}_\text{plug-in}$.}
  \label{fig:mse-simulations}
\end{figure*}

In Figure~\ref{fig:mse-simulations}, we observe that the plug-in estimator's
MSE is roughly $3\times$ larger for $n = 10^2$ and remains nearly an order of
magnitude larger for $n = 10^4$.  We take a deeper look to understand the
source of improvement and observe that gains in MSE are driven almost entirely
by bias removal rather than additional regularization.
Figure~\ref{fig:bias-variance-simulations} shows that debiasing achieves more
than a two-fold bias reduction even in the smallest sample regime and over a
ten-fold reduction by $n = 10^4$.  The plug-in procedure suffers from
considerable finite-sample bias, which dominates its error profile even at
$n = 10^5$. In contrast, the debiased estimator markedly reduces bias while
keeping the variance stays within $10\%$ of the plug-in estimator across all
sample sizes, leading to dramatic gains in MSE.  These experiments confirm
that the orthogonalized correction is crucial for accurate worst-case
subpopulation risk estimation in realistic finite-sample settings.

\begin{figure}[t]
  \centering
  \begin{subfigure}[t]{0.48\textwidth}
    \centering
    \includegraphics[width=\textwidth]{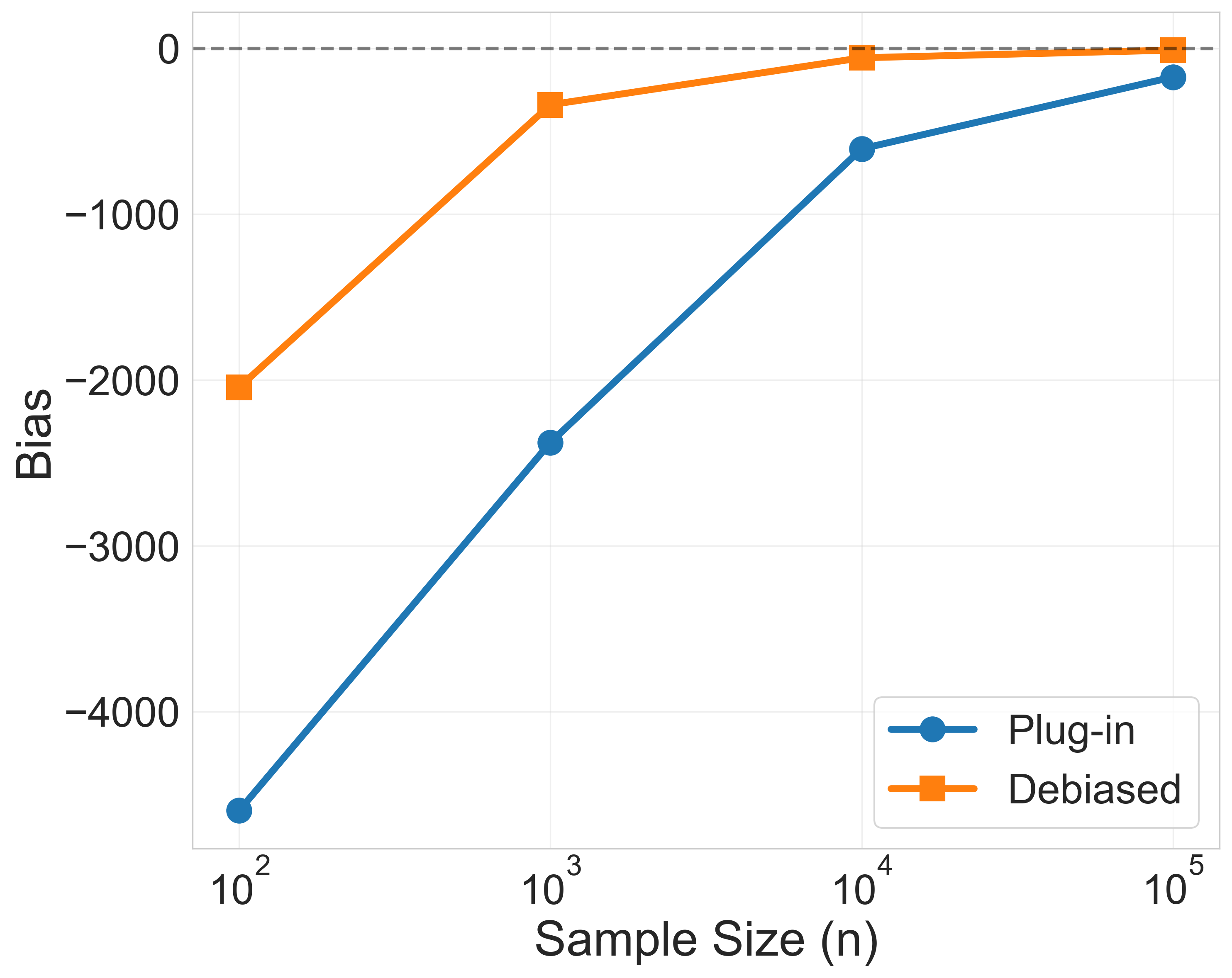}
    \caption{Bias of each estimator}
    \label{fig:bias-variance-simulations-a}
  \end{subfigure}
  \hfill
  \begin{subfigure}[t]{0.48\textwidth}
    \centering
    \includegraphics[width=\textwidth]{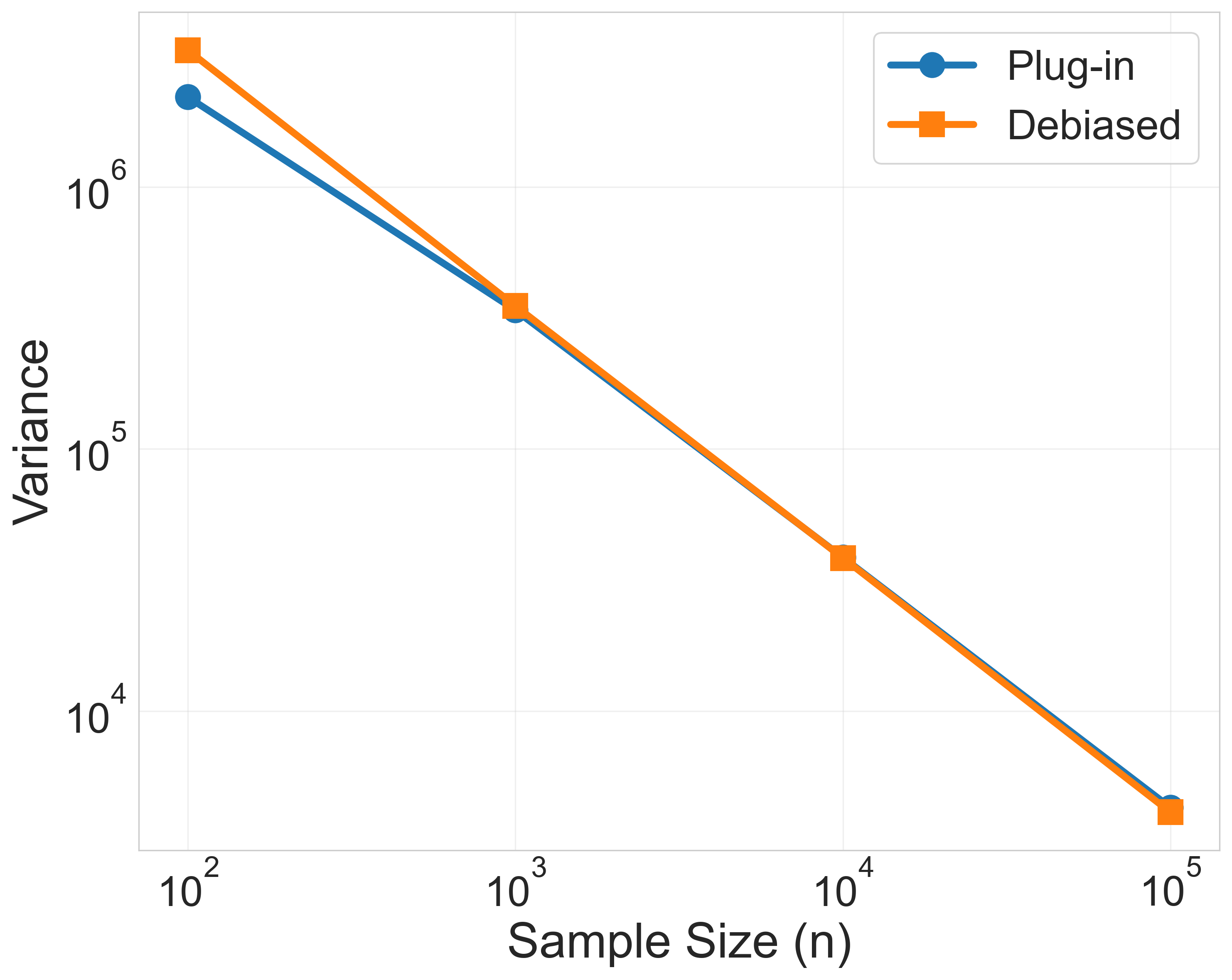}
    \caption{Variance of each estimator}
    \label{fig:bias-variance-simulations-b}
  \end{subfigure}
  \caption{Debiasing sharply reduces bias without inflating variance. Bias
    improvements exceed $2\times$ at $n=10^2$ and remain at least $6\times$
    through $n = 10^5$, while the variance stays comparable to the plug-in
    baseline.}
  \label{fig:bias-variance-simulations}
\end{figure}
}

\paragraph{Comparison with~\citet{SubbaswamyAdSa21}} 
So far, we considered an estimand that involves an optimization problem over
the dual variable $\eta$. We derived a debiased estimation approach accounting
for errors in estimating $\mu_P$, which gave us a formula involving
$(\mu_P, \tau_P)$. Later, we will show that our final estimator is indeed
debiased with respect to the nuisance parameters $(\mu_P, \tau_P)$.
Concurrent to an earlier conference version of this
work,~\citet{SubbaswamyAdSa21} study a similar setting and propose another
estimator different from ours.  In contrast to our approach, they treat
$(\mu_P, \eta_P)$ as nuisance parameters and define the estimand as
\begin{equation*}
  T(P; \mu, \eta) =  \frac{1}{\alpha} \E_P \hinge{\mu(\worstcov) - \eta} + \eta.
\end{equation*}
As is evident here, this approach requires debiasing with respect to
\emph{both} $\mu$ and $\eta$.

Given black-box estimates
$(\what{\mu}, \what{\eta})$,~\citet{SubbaswamyAdSa21} estimate
the following expression on a separate fold
\begin{equation*}
  \frac{1}{\alpha} \E_{\worstcov \sim P} \hinge{\what{\mu}(\worstcov) - \what{\eta}}
  + \what{\eta} + \frac{1}{\alpha} \E_{\worstcov\sim P}[\indic{\what{\mu}(\worstcov) \ge \what{\eta}} (\loss(\theta(X); Y) - \what{\mu}(\worstcov))].
\end{equation*}
They claim this approach is debiased and cite~\citet{JeongNa20}'s related
approach as the justification. However,~\citet{JeongNa20}---which we also draw
inspiration from---follow an exactly analogous approach as ours involving just
$(\mu_P, \tau_P)$ and do not consider $\eta$ as a separate nuisance parameter.
This difference results in an erroneous asymptotic result in~\citet[Theorem
1]{SubbaswamyAdSa21}; as far as we can tell, their proof of Neyman
orthogonality is incorrect and their proposed estimator does not appear
debiased. For example, they cite Danskin's theorem despite their functional
$T(P; \mu, \eta)$ not involving an optimization functional.


\section{Convergence guarantees}
\label{section:convergence}

To \emph{rigorously} verify the robustness of a model prior to deployment, we
present convergence guarantees for our estimator
(Algorithm~\ref{alg:two-stage}).  We begin by proving a central limit result
for our estimator, showing that debiasing allows us to achieve standard
$\sqrt{n}$-rates of convergence even when the fitted $\what{\mu}_k$ converge
at a slower rate.  Since asymptotic guarantees ignore the dimensionality of
$Z$, we then turn to finite sample concentration guarantees.
{
However, our
finite sample guarantees are limited in that they do not show the benefits of
our debiased estimator. Developing better mathematical machinery that allows
quantification of the benefits of debiasing remains a fruitful direction of
research. }
Finally, we provide convergence guarantees for our
estimator~\eqref{eqn:certificate-emp} for the certificate of
robustness~\eqref{eqn:certificate} in the appendix
(Section~\ref{section:certificate}).

\subsection{Asymptotics}
\label{section:asymptotics}

In contrast to the literature on debiased estimation, our estimand is
nonlinear in $P$, which requires a different proof approach than what is
standard (\eg,~\citep{ChernozhukovChDeDuHaNeRo18}).  Our asymptotic result
proof approach is inspired by~\citet{JeongNa20}, using standard tools in
empirical process theory.
\begin{assumption}
  \label{assumption:residuals}
  Bounded residuals
  {\small $\E[\loss(\theta(X); Y)^2] + \linfstatnorm{\E[ (\loss(\theta(X); Y) - \mu\opt(\worstcov))^2 \mid \worstcov]} < \infty$} 
\end{assumption}
\begin{assumption}
  \label{assumption:neyman} Let $\Lone{\hmu -\pmu} \cas 0$, and let there
  exist an envelope function $\env: \mathcal{\worstcov} \to \R$ satisfying
  $\E[\env(\worstcov)^2] < \infty$ and $\max(|\hmzero|, |\hmone|) \le \env$.
  There exists $\delta_n, \Delta_n \downarrow 0$, and $\hbound > 0$ such that
  with probability at least $1-\Delta_n$, for all $\indfold \in [K]$,
  $|\hthr| \le \hbound$, $\Linf{\hmu - \pmu} \le \delta_n n^{-1/3}$, and
  $|\hq - \quantile{1-\alpha}{\hmu}| \le \delta_n n^{-1/3}$.
\end{assumption}

To estimate the tail-average~\eqref{eqn:dual} (recall Lemma~\ref{lemma:dual}),
we need to estimate the quantile $\aq{\pmu}$. We assume that for functions
around $\pmu$, their positive density exists at the
$(1-\alpha)$-quantile~\cite[Chapter 3.7]{VanDerVaartWe96}.  Let $\mathcal{U}$ be a
set of (measurable) functions $\mu: \mathcal{\worstcov} \to \R$ such that
\begin{quote}
  $F_{r, \mu}$, the cumulative distribution of $(\pmu + r(\mu - \pmu))(\worstcov)$, is
  uniformly differentiable in $r \in [0, 1]$ at $\aq{\pmu + r(\mu - \pmu)}$,
  with a positive and uniformly bounded density. Formally, if we let
  $q_{r, \mu} \defeq \aq{\pmu + r(\mu - \pmu)}$, then for each $r \in [0, 1]$,
  there is a positive density $f_{r, \mu}(q_{r, \mu}) > 0$ such that
\begin{equation}
  \label{eqn:unif-diff}
  \lim_{t \to 0} \sup_{r \in [0, 1]}
  \left| \frac{1}{t} \left(F_{r, \mu}(q_{r, \mu} + t)
      - F_{r, \mu}(q_{r, \mu})\right)
    - f_{r, \mu}(q_{r, \mu}) \right| = 0.
\end{equation}
\end{quote}
We require this holds for our estimators $\mu = \hmu$ with high probability.
\begin{assumption}
  \label{assumption:regularity}
 $\exists \Delta_n' \downarrow 0$ s.t. with probability at least
  $1-\Delta_n'$, $\hmu \in \mathcal{U}$ for all $\indfold \in [K]$.
\end{assumption}

{
Let's consider a bounded open neighborhood $N$ containing the level sets
\begin{equation*}
  \{ z : \mu^\star(z) = q~~\mbox{for some}~q~\mbox{in a neighborhood of} ~\aq{\pmu} \}
\end{equation*}
Let $Z \in \R^d$ have a continuous density $p_Z(\cdot)$ satisfying
$0< c \le p_Z(\cdot) \le C < \infty$ on the set $N$.  If we consider
continuously differentiable models $\mu(\cdot)$ satisfying
$0< c' \le \norm{\nabla \mu(\cdot)} \le C' < \infty$ on the set $N$, we have
the uniform differentiability condition since the implicit function theorem
gives that the density of $\mu(\cdot)$ is given by
$p_{\mu}(t) = \int_{\{ z : \mu(z) = t \}} \frac{p_Z(z)}{ \| \nabla \mu(z) \| }
\, dS(z)$ where $dS$ denotes the $(d-1)$--dimensional surface measure on the
level set $\{ z : \mu(z)=t \}$.  For example, if $\mu_{\beta} = \beta^\top Z$
satisfies these conditions if $Z$ is Gaussian.
}

Our debiased estimator $\what{\omega}_{\alpha}$ enjoys central limit rates
with the influence function
\begin{align}
  \label{eqn:influence}
  \psi(X, Y, Z) \defeq
  \left[ \frac{1}{\alpha} \hinge{\pmu(\worstcov) - \aq{\pmu}} + \aq{\pmu} \right]
  - \worstsub\opt
  + \pthr(\worstcov) (\loss(\theta(X); Y) - \pmu(\worstcov)).
\end{align}
See Section~\ref{section:proof-clt} for the proof of the following central
limit result.
\begin{theorem}
  \label{theorem:clt}
  Under Assumptions~\ref{assumption:residuals}-\ref{assumption:regularity},
  $\sqrt{n} (\what{\omega}_{\alpha} - \worstsub\opt) \cd N(0, \var (\psi(X, Y, Z)))$.
\end{theorem}

\subsection{Concentration using the localized Rademacher complexity}
\label{section:finite-sample}

We give finite-sample convergence at the rate
$\scriptsize{O_p(\sqrt{\mathfrak{Comp}_n(\mathcal{H}) / n})}$, where
$\mathfrak{Comp}_n(\mathcal{H})$ is the localized Rademacher
complexity~\citep{BartlettBoMe05} of the model class $\mathcal{H}$ for estimating
the conditional risk $\condrisk(\worstcov)$.  We restrict attention to
nonnegative and uniformly bounded losses, as is conventional in the
literature.
\begin{assumption}
\label{assumption:bdd}
There is a $\lbound$ such that $\loss(\theta(X); Y) \in [0, \lbound]$, and
$\model(\worstcov) \in [0, \lbound]$ a.s. for all $\model \in \modelclass$.
\end{assumption}
\noindent Throughout this section, we do not stipulate well-specification, meaning that we
allow the conditional risk
$\condrisk(\worstcov) = \E[\loss(\theta(X); Y) \mid \worstcov]$ not to be in the model
class $\modelclass$.

To characterize the finite-sample convergence behavior of our estimator
$\what{\omega}_{\alpha}$, we begin by decomposing the error of the augmented
estimator $\what{\omega}_{\alpha,k}$ on each fold $I_k$ into two terms
relating to the two stages in Algorithm~\ref{alg:two-stage}. Recalling the
notation in Eq.~\eqref{eqn:dual} (so that
$\worstsub\opt = \worstsub(\condrisk)$), we have
\begin{equation*}
  \worstsub\opt - \what\omega_{\alpha,k}
  = \underbrace{T(P;\mu\opt,\tau\opt) - T(P;\what\model_k,\hthr)}_{(a): ~\mbox{\scriptsize first stage}}
  + \underbrace{T(P;\what\model_k,\hthr) - T(\what{P}_k;\what\model_k,\hthr)}_{(b): ~\mbox{\scriptsize second stage}},
\end{equation*}
because $\worstsub\opt = T(P;\mu\opt,\tau\opt)$ and $\what\omega_{\alpha,k} = T(P;\what\model_k,\hthr)$.
To bound term $(b)$, we prove concentration guarantees for estimators of the
dual~\eqref{eqn:dual} (see Proposition~\ref{prop:cvar-concentration} in
Appendix~\ref{section:proof-cvar-concentration}).  To bound term $(a)$, we use
a localized notion of the Rademacher complexity.

Formally, for $\xi_1, \ldots, \xi_n \in \Xi$ and i.i.d.\ random signs
$\varepsilon_i \in \{-1,1\}$ (independent of $\xi_i$), recall the standard
notion of (empirical) Rademacher complexity of
$\mathcal{G} \subseteq \{ g: \Xi \to \R\}$ 
\begin{equation*}
  \radcomp_n(\mathcal{G}) \defeq \E_{\varepsilon} \left[\sup_{g \in \mathcal{G}}
    \frac{1}{n} \sum_{i = 1}^n \varepsilon_i g(\xi_i) \right].
\end{equation*}
We say that a function $\psi : \R_+ \to \R_+$ is
\emph{sub-root}~\citep{BartlettBoMe05} if it is nonnegative, nondecreasing, and
$r \mapsto \psi(r) / \sqrt{r}$ is nonincreasing for $r > 0$. Any
(non-constant) sub-root function is continuous, and has a unique positive
fixed point. Let $\psi_n : \R_+ \to \R_+$ be a sub-root upper bound on the
localized Rademacher complexity 
  $\psi_n(r) \ge \E\left[\radcomp_n\left\{g \in \mathcal{G} : 
        \E[g^2] \le r \right\}\right]$.
(The localized Rademacher complexity itself is sub-root.) The fixed point of
$\psi_n$ characterizes generalization guarantees~\citep{BartlettBoMe05,
  Koltchinskii06a}. 


Let $\bestmodel$ be the best model in the model class $\mathcal{H}$
\begin{align*}
  \bestmodel \defeq \argmin_{\model \in \modelclass} \E[(\loss(\theta;X, Y) - \model(\worstcov))^2].
\end{align*}
Let $\psi_{|\cfold|}(r)$ be a subroot upper bound on the localized Rademacher
complexity around $\bestmodel$
\begin{align}
  \label{eqn:subroot}
   \psi_{|\cfold|}(r) \ge 2 \E\left[\radcomp_{|\cfold|}\left\{ \model \in \mathcal{H}:  \E[(\model(\worstcov) - \bestmodel(\worstcov))^2] \le
   r \lbound^2/4 \right\}\right].
\end{align}
We define $r_{|\cfold|}\opt$ as the fixed point of $\psi_{|\cfold|}(r)$.

As we show shortly, we bound the estimation error of our procedure using the
\emph{square root} of the excess risk in the first-stage
problem~\eqref{eqn:first-pop}
\begin{equation}
  \label{eqn:decomposition}
  \E\left[\left( \loss(\theta; X, Y) - \what{\model}_k(\worstcov)\right)^2 \mid \cfold\right]
  - \E\left[\left( \loss(\theta; X, Y) - \bestmodel(\worstcov)\right)^2 \right]
\end{equation}
By using a refined analysis offered by localized Rademacher complexities, we
are able to use a fast rate of convergence of
$\scriptsize{O_p(\mathfrak{Comp}_n(\mathcal{H})/n)}$ on the preceding excess risk.
In turn, this provides the following
$\scriptsize{O_p(\sqrt{\mathfrak{Comp}_n(\mathcal{H}) / n})}$ bound on the
estimation error as we prove in Appendix~\ref{section:proof-uniform}.  In the
bound, we have made explicit the approximation error term
$\norm{\bestmodel - \condrisk}_{L^2}$. As the model class $\mathcal{H}$ grows richer,
there is tension as the approximation error term will shrink, yet the
localized Rademacher complexity of $\mathcal{H}$ will grow.
\begin{theorem}
  \label{theorem:uniform}
  Let Assumption~\ref{assumption:bdd} hold. For some constant $C>0$,  
  for each fold $k\in[K]$, with probability at least $1-3\delta$,
  \begin{align*}
    \left|\worstsub\opt - \what{\omega}_{\alpha,k}\right|
    \le \frac{C \lbound}{\alpha} \left(  \sqrt{ {r}_{(1-K^{-1})n}\opt}
    + \sqrt{\frac{K\log(2/\delta)}{n}} \right) 
    + \frac{2}{\alpha} \norm{\bestmodel - \condrisk}_{L^2}.
  \end{align*} 
\end{theorem}
\noindent 
By controlling the fixed point
$r_{n}\opt$ of the localized Rademacher complexity, we are able to provide
convergence of our estimator~\eqref{eqn:second}.  For example, when
$\modelclass$ is a bounded VC-class~\citep{VanDerVaartWe96}, it is known that
its fixed point satisfy~\cite[Corollary 3.7]{BartlettBoMe05}
\begin{equation*}
  r_{n}\opt \asymp \log (n / \vcdim(\mathcal{H}))\cdot \vcdim(\mathcal{H}) / n,
\end{equation*}
where $\vcdim(\cdot)$ is the VC-dimension.

\subsection{Data-dependent dimension-free concentration}
\label{section:dim-free}

In some situations, it may be appropriate to define subpopulations
($\worstcov$) over features of an image, or natural language descriptions. For
such high-dimensional variables $\worstcov$ and complex model classes $\mathcal{H}$
such as deep networks, the complexity measure $\mathfrak{Comp}_n$ is often
prohibitively conservative and renders the resulting concentration guarantee
meaningless.  We provide an alternative concentration result that depends on
the size of model class $\modelclass$ only through the out-of-sample error in
the first-stage problem~\eqref{eqn:first-pop}.  This finite-sample,
data-dependent convergence result depends only on the out-of-sample
generalization error for estimating $\condrisk(\cdot)$. In particular, the
out-of-sample error can grow smaller as $\mathcal{H}$ gets richer, and as a result
of hyperparameter tuning and model selection, it is often very small for
overparameterized models such as deep networks. This allows us to construct
valid finite-sample upper confidence bounds for the worst-case subpopulation
performance~\eqref{eqn:cvar} even when $\worstcov$ is defined over
high-dimensional features and $\mathcal{H}$ represent deep networks.

For simplicity, denote
\begin{equation}
  \label{eqn:mse}
  \Delta_{S}(\model) \defeq \frac{1}{|S|} \sum_{i\in S} (\loss(\theta(X_i);Y_i) - \model(\worstcov_i))^2.
\end{equation}
for any function $\model:\mathcal{Z}\to\R$ on any data set $S$. We prove the
following result in Appendix~\ref{section:proof-dim-free}.
\begin{theorem}
\label{theorem:dim-free}
  Let Assumption~\ref{assumption:bdd} hold. For some constant $C>0$, 
   for each fold $k\in[K]$, with probability at least $1-3\delta$, 
  \begin{equation*}
    \left|\worstsub\opt - \what{\omega}_{\alpha,k}\right| \le \frac{2}{\alpha} 
  \Bigg(\sqrt{[\Delta_{I_k}(\what{\model}_\indfold) - \Delta_{I_k}(\bestmodel)]_+}
    + \norm{\bestmodel - \condrisk}_{L^2} 
    + C\lbound \left(\frac{2K\log(2/\delta)}{n}\right)^{1/4}\Bigg).
  \end{equation*}
  Moreover, if the model class $\modelclass$ is convex, then
  $\norm{\bestmodel-\condrisk}_{L^2}$ can be replaced with
  $\norm{\bestmodel-\condrisk}_{L^1}$.
\end{theorem}
\noindent Following convention in learning theory, we refer to our
data-dependent concentration guarantee \emph{dimension-free}. For
overparameterized model classes $\modelclass$ such as deep networks, the
localized Rademacher complexity in Theorem~\ref{theorem:uniform} becomes
prohibitively large~\citep{BartlettFoTe17, ZhangBeHaReVi17}. In contrast, the
current result can still provide meaningful finite-sample bounds: model
selection and hyperparameter tuning provides low out-of-sample performance in
practice, and the difference
$\Delta_{I_k}(\what{\model}_k) - \Delta_{I_k}(\bestmodel)$ can be often made
very small. Concretely, it is possible to calculate an upper bound on this
term as $\Delta_{I_k}(\bestmodel)$ is lower bounded by
$\min_{\model\in\modelclass}\Delta_{I_k}(\model)$.

\subsection{Extensions for heavy-tailed loss functions}
\label{sec:heavy-tail}
While it is standard to study uniformly bounded losses when considering
finite-sample convergence guarantees, it is nevertheless a rather restrictive
assumption. We now show concentration guarantees for broader classes of losses
such as sub-Gaussian or sub-exponential ones.  Our analysis builds on
\citet{Mendelson14}'s framework for heavier-tailed losses based on one-sided concentration inequalities.

Denote as $\mathcal{D}_{\condrisk}$ the $L_2(\P)$ unit ball centered at $\condrisk$, 
 the residuals $\zeta_i \defeq \loss(\theta(X_i); Y_i) - \condrisk(\worstcov_i)$ and $\varepsilon_i\in\{-1,1\}$ 
iid random signs, we define 
\begin{align*}
  \phi_{\cfold}(s) & \defeq \sup_{\model \in \modelclass \cap s\mathcal{D}_{\condrisk} }\left| \frac{1}{\sqrt{|\cfold|}} \sum_{i\in \cfold} \varepsilon_i \zeta_i (\model(Z_i) - \condrisk(Z_i)) 
  \right|, \\
  \alpha_{\cfold} (\gamma, \delta) & \defeq \inf \left\{ s > 0 : \P\left(\phi_{\cfold}(s) \le \gamma s^2\sqrt{|\cfold|}\right) \ge 1-\delta \right\}, \\
  \beta_{\cfold} (\gamma) & \defeq \inf \left\{ r > 0 : \E \sup_{\model \in \modelclass \cap s\mathcal{D}_{\condrisk} } \left| \frac{1}{\sqrt{|\cfold|}} \sum_{i\in \cfold} \varepsilon_i 
                            (\model(Z_i) - \condrisk(Z_i)) \right|\le \gamma r\sqrt{|\cfold|} \right\},\\
  Q_{\modelclass-\modelclass} (u) & \defeq \inf_{\model_1,\model_2 \in \modelclass} \P(|\model_1-\model_2| \ge u \norm{\model_1-\model_2}_{L^2}).
\end{align*}

\begin{assumption}
\label{assumption:well}
The conditional risk is in the model class $\mathcal{H}$:
$\condrisk(\worstcov) = \E[\loss(\theta(X); Y) \mid \worstcov] \in \mathcal{  H}$.
\end{assumption}

\begin{assumption}
\label{assumption:convex}
  The model class $\modelclass$ is closed and convex. 
\end{assumption}

\begin{assumption}
\label{assumption:subgaussian}
  The loss function $\ell$ is sub-Gaussian with parameter $\lbound^2$. 
\end{assumption}

\begin{theorem}[\citet{Mendelson14}, Theorem 3.1]
  Let Assumptions~\ref{assumption:well},~\ref{assumption:convex},~\ref{assumption:subgaussian} hold.
  Fix $\tau > 0$ for which $Q_{\modelclass-\modelclass}(2\tau) > 0$ 
  and set $\gamma < \tau^2 Q_{\modelclass-\modelclass}(2\tau)/16$. There is a numerical constant $C>0$ such 
  that for every $\delta\in(0,1)$ and $k\in[K]$, with probability at least $1-\delta-\exp(-(1-K^{-1})nQ_{\modelclass-\modelclass}(2\tau)^2/2)$,
  \begin{equation}
    \left|\worstsub(\condrisk) - \what{\omega}_{\alpha,k} \right| \le \frac{C}{\alpha} 
    \left( 
    \alpha_{\cfold}\left(\gamma, \frac{\delta}{4}\right) 
    +
    \beta_{\cfold}\left(\frac{\tau Q_{\modelclass-\modelclass}(2\tau)}{16} \right) 
    +
    \lbound\sqrt{ \frac{K\log(2/\delta)} {n} }
    \right)
  \end{equation}
\end{theorem}

For an example of where this result provides a tighter bound than that of Theorem~\ref{theorem:uniform}, 
we look at the persistence framework. 
Let $Z \in \mathbb{R}^m$ be a random vector with independent mean-zero, variance-1 random coordinates. 
Consider the linear model class $\modelclass=\{\langle t, \cdot\rangle: \E\norm{t}_1\le R\}$, and suppose 
the conditional risk is $\model\opt = \langle t\opt, \cdot \rangle + \varsigma$, where $\E\norm{t\opt}_1 
\le R$ and $\varsigma$ is an independent mean-zero random variable with variance at most $\lbound^2$.
\begin{lemma}[\citet{Mendelson14}, Theorem 4.6]
 Let Assumptions~\ref{assumption:bdd},~\ref{assumption:well},~\ref{assumption:convex} hold. 
 Fix $\lbound>1$. There exist constants $c_1$, $c_2$ and $c_3$ that depend only on $\lbound$ for 
 which with probability at least $1-2\exp(-c_3 |\cfold| v_2 \min\{B^{-2},R^{-1}\})$,
 \begin{equation}
   \left|\worstsub(\condrisk) - \hworstsub(\what{\model}_k)\right| 
   \le \frac{C}{\alpha} 
   \left( 
   \sqrt{ \max\{v_1,v_2\} } 
   +
   \lbound \sqrt{ \frac{\log(2/\delta)} {|I_k|} }
   \right),
 \end{equation}
 where we define
 \begin{equation}
   v_1  = \frac{R^2}{|\cfold|} \log\left(\frac{2c_1 m}{|\cfold}\right) \mathbf{1}\{|\cfold| \le c_1 m\}
 \end{equation}
 \begin{equation}
   v_2  = 
   \begin{cases}
   \frac{R\lbound}{\sqrt{|\cfold|}} \sqrt{\log\left(\frac{2 c_2 m \lbound}{\sqrt{|\cfold|}R}\right)} 
   & \text{ if } |\cfold| \le c_2 m^2 \lbound^2 / R^2,\\
   \frac{\lbound^2 m}{|\cfold|} & \text{otherwise}.
   \end{cases}
 \end{equation}
\end{lemma}


\section{Simulation experiments}
\label{section:experiments}

We begin by verifying the asymptotic convergence of our proposed two-stage
estimator through a simulation experiment on a classification task.

\begin{figure}[t]
	\centering
	\includegraphics[width = 0.4\textwidth]{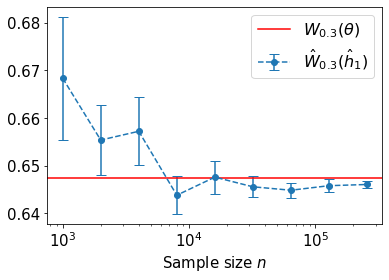}
	\caption{$\hworstsub(\what\model_1)$ and $\worstsub(\theta)$ from simulation experiments with $\alpha=0.3$}
	\label{fig:simulation}
\end{figure}


We illustrate the asymptotic convergence of our two-stage estimator
$\hworstsub(\what\model_1)$ of the worst-case subpopulation performance
$\worstsub(\theta)$. We conduct a binary classification experiment where we
randomly generate and fix two vectors $\theta, \theta_0^\star \in\mathbb{R}^d$
on the unit sphere. The data-generating distribution is given by
$X\overset{iid}{\sim} \mathsf{N}(\gamma, \Sigma)$ and
\begin{equation*}
Y \mid X = 
\begin{cases}
\text{sgn}(X^\top \theta_0^\star)  & X^1 \le z_{0.95} = 1.645 \\ 
-\text{sgn}(X^\top \theta_0^\star) & \text{otherwise}.
\end{cases}
\end{equation*}
In this data-generating distribution, there is a drastic difference between
subpopulations generated by $X^1 \le z_{0.95}$ and $X^1 > z_{0.95}$; typical
prediction models will perform poorly on the latter rare group.
The loss function is taken to be the hinge loss
$\loss(\theta; x, y) = [1-y\cdot\theta^\top x]_+$, where $y\in\{\pm1\}$. We
take the first covariate $X^1$ as our protected attribute $\worstcov$.  Let
$d=5$, $\Sigma = \mathsf{I}_5$, $\gamma = 0$.


We fix $\alpha = 0.3$.  To analyze the asymptotic convergence of our two-stage
estimator, for sample size ranging in 1,000 to 256,000 doubling each time, we
run 40 repeated experiments of the estimation procedure on simulated data.  We
split each sample evenly into $S_1$ and $S_2$ and using gradient boosted trees
in the package XGBoost~\cite{ChenGu16} to estimate the conditional risk. On a
log-scale, we report the mean estimate across random runs in
Figure~\ref{fig:simulation} alongside error bars. To compute the true
worst-case subpopulation performance $\worstsub(\condrisk)$ of the conditional
risk $\condrisk(X^1)$, we first run a Monte Carlo simulation for 150,000
copies of $X^1\sim \mathsf{N}(0,1)$. For each sampled $X^1$, we generate
100,000 copies of
$(X^2, X^3, X^4, X^5)\sim\mathsf{N}(\mathbf{0}, \mathsf{I}_4)$ independent of
$X^1$ and compute the mean loss among them to approximate the conditional risk
$\condrisk(X^1)$. Finally, we approximate $\worstsub(\condrisk)$ using the
empirical distribution of $\condrisk(\cdot)$, obtaining $6.47\times10^{-1}$.
We observe convergence toward the true value as sample size $n$ grows,
verifying the consistency of our two-stage estimator
$\hworstsub(\what{\model}_1)$.


{
\section{Case studies}
\label{section:case}

Now that we have verified the statistical validity of the proposed
methodology, we now provide case studies based on real datasets that shed
light on the applicability of the proposed framework. Along the way, we
also highlight the limitation of our worst-case subpopulation approach:
\begin{quote}
  \emph{It is impossible to guard against performance degradation on arbitrary
  out-of-distribution data. As such, our worst-case subpopulation approach
  provides inherently limited insights on subpopulation shifts of a
  certain size and is not meant to be taken as a panacea.}
\end{quote}
With this caveat in mind, we will use the following case studies to illustrate
how traditional model selection approaches that rely on average-case metrics
(e.g., accuracy, cross-entropy loss) can obscure significant performance
degradation on minority or tail subpopulations. In contrast, our framework can
serve as a useful diagnostic without requiring prior knowledge of specific
demographic attributes or access to out-of-distribution (OOD) data.

We begin by examining a precision medicine application (optimal Warfarin
dosage), a long-standing problem affecting millions of patients. We
demonstrates how worst-case subpopulation analysis can reveal critical
performance disparities masked by average-case metrics, highlighting the
challenge of achieving uniform robustness across patient subgroups.  Then, we
present two comprehensive case studies—ACS Income and satellite image
classification—to evaluate the performance of our metric on real-world test
sets. These case studies explore distribution shifts that do not necessarily
align with those used in our diagnostic analysis, allowing us to uncover the
types of insights our metric (worst-case subpopulation performance) can and
cannot provide in practical scenarios.

A key finding from these case studies is that out-of-distribution (OOD)
performance is governed by a trade-off between two competing factors: (1)
performance on in-support regions, where our metric remains predictive, and
(2) performance on out-of-support regions, where performance may degrade
arbitrarily.  By analyzing these real-world distribution shifts, we offer
guidance on when our metric can be reliably used for model selection and when
additional domain expertise or robustness-enhancing techniques may be
necessary.

For each case study, we compute our proposed metric $W_\alpha(\theta)$
(Algorithm~\ref{alg:two-stage}), across varying subpopulation sizes
$\alpha$. We validate this against held-out in-distribution subpopulation test
sets and out-of-distribution test sets. We validate the metric’s effectiveness
using both held-out in-distribution subpopulation test sets and
out-of-distribution test sets. Across case studies, we consistently observe
three key findings:
\begin{enumerate}
\item \textbf{Worst-case subpopulation metrics distinguish between models with
    similar average performance:} Our metric can identify models that do not
  maintain uniform performance across in-distribution subpopulations, in
  contrast to average-case metrics (e.g., accuracy, cross-entropy loss).

\item \textbf{In-support OOD shifts: Metric is predictive.}  When the OOD
  shifts occur within the support of the training distribution (e.g., new
  geographic regions with similar demographic structures in the ACS dataset),
  our in-distribution diagnostic can reliably predict OOD robustness. Models
  identified as robust by our metric consistently outperform alternatives in
  these settings.

\item \textbf{Out-of-support OOD shifts: Metric cannot guarantee robustness.}
  When shifts involve truly novel, out-of-support distributions---i.e., data
  regions not represented during training---even models deemed robust by our
  metric may suffer substantial performance degradation. This highlights a key
  limitation of our framework and underscores the importance of distinguishing
  between in-support and out-of-support shifts in OOD evaluation.
\end{enumerate}

These case studies enable us to investigate real-world distribution shifts
that include both in-support scenarios---where our diagnostic is effective—and
out-of-support scenarios---where robustness cannot be guaranteed. This dual
perspective sheds light on the strengths and limitations of worst-case
subpopulation analysis. While our framework offers actionable insights into
model vulnerabilities within the training distribution, it is not a formal
guarantee against all types of OOD shifts. Rather, it is intended as a
diagnostic tool to help practitioners build intuition about model behavior on
subpopulations observed during training.

}

\begin{figure}[t]
\centering
\includegraphics[width = 0.55\textwidth]{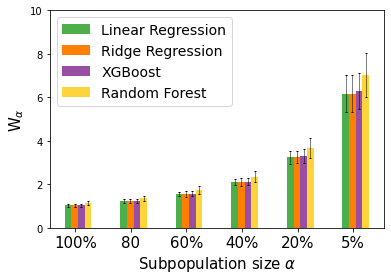}
\vspace{-5pt}
\caption{Worst-case subpopulation performance $\worstsub(\theta)$, where
  $\mathsf{W}_{1.0}(\theta) = \E[\loss(\theta(X); Y)]$. Results are averaged over 
  50 random seeds with error bars corresponding to 95\% confidence interval
  over the random runs.}
\label{fig:warfarin}
\end{figure}

\subsection{Warfarin Optimal Dosage}
Precision Medicine---an emerging approach that treats diseases at a personal
level incorporating individual variability in genes---has attracted great
attention in recent years. One important area precision medicine intends to
tackle is optimal dosage: given the patients' individual characteristics such
as demographic, genetic, and symptomatic information, is it possible to design
an automated algorithm to predict the optimal dosage for the patients?
Unfortunately, this task is often presented with much difficulty.  In the case
of Warfarin --- one of the most widely used anticoagulant agents --- its
optimal dosage can differ substantially across genetics, demographics, and
existing conditions of the patients by a factor as much as
10~\citep{IWPC09}. Traditionally, physicians often determine the dosage
through trial and error, but this large variation makes the appropriate dosage
hard to establish, and an incorrect dosage can lead to highly undesirable side
effects. It is therefore important to develop a more reliable method to help
determine the optimal dosage for the patients. Furthermore, to ensure fair
treatment to all patients, it is imperative that the model performs uniformly
well over all subpopulations.

We use the Warfarin optimal dosage prediction problem to illustrate how our
metric can be used as a robustness certificate of the model that informs model
selection. We consider the Pharmacogenetics and Pharmacogenomics Knowledge
Base dataset where the Warfarin optimal dosage is found through trial and
error by clinicians. The dataset consists of 4,788 patients (after excluding
missing data) with features representing \emph{demographics, genetic markers,
  medication history, pre-existing conditions, and reason for treatment}.  It
has been observed empirically in~\citet{IWPC09} that a linear model
outperforms a number of more complicated modeling approaches (including kernel
methods, neural networks, splines, boosting) for predicting the optimal
dosage, at least based on average prediction accuracy on the out-of-sample
test set.

Comparable average case performance does not guarantee similar worst case
performance; in dosage prediction problems fair treatment to all groups,
including the underrepresented groups, is essential~\citep{CharShMa18,
RajkomarHaHoCoCh18, GoodmanGoCu18, AMA18}. With far fewer model parameters
than other more expressive models, are linear models truly on par with
other models in ensuring uniform good performance overall all subgroups? 

To answer this question, we evaluate and compare the worst-case
subpopulation performance of different models over $\worstcov = X$.
We take the entire feature vector including all available demographic and
genetic information as core attributes defining the subpopulations. 
By taking such a core attribute vector we are being extra conservative, 
but this decision is motivated by the nature of the optimal dosage
prediction problem in that one shall make decisions based on 
worst case performance guarantees for all patients, irrespective of
their demographics, genetic markers, pre-existing conditions, etc.  

More specifically and following the approach in~\citet{IWPC09},
we take the root-dosage as our outcome $Y$,
and consider minimizing the squared loss function 
\[
\loss(\theta(X); Y) = (Y - \theta(X))^2.
\] 
We consider four popular models common used in practice:
\emph{Linear Regression, Ridge Regression, XGBoost, Random Forests}. 
Past literature has shown that Linear Regression model does not
underperform other more expressive models. If we can certify that
Linear Regression model is at least as robust as other models,
then we provide a certificate to the Linear Regression model
and one would naturally choose the linear models over others
thanks to their simplicity and interpretability.

Figure~\ref{fig:warfarin} plots our metric against different choice of
subpopulation size $\alpha$, for the four models considered.
We observe that the performance of linear model closely matches 
that of other more expressive models, and the trend holds over a range
of different subpopulation sizes, even for small $\alpha = 5\%$.
Our finding thus instills confidence in the linear regression model:
our diagnostic is able to certify its advantageous performance
even on tail subpopulations despite it is the simplest among the four models. 

At the same time, our diagnostic raises concerns about poor tail
subpopulation performance: all models suffer from significant
performance deterioration on small subpopulation sizes (e.g. $\alpha = 5\%$),  
and the prediction loss is as much as six times worse than the
average-case performance. This observation shows that achieving
uniformly good performance across subgroups is a challenging task
in the Warfarin example, and more attention is needed to address this
significant deterioration of performance on the worst-case subpopulation.

{
\subsection{ACS Income}

 We now present our first case study using data from the
  U.S. Census American Community Survey (ACS) \citep{DingHaMoSc21}. This
  application focuses on predicting key socioeconomic outcomes and evaluates
  the robustness of our diagnostic framework across diverse datasets and model
  architectures. We consider the ACS Income prediction task: given
  demographic, geographic, and employment features for each individual, the
  goal is to predict whether their annual income exceeds \$50K. The dataset
  includes all 50 states and Puerto Rico, spanning a wide range of demographic
  groups.

  In this study, we train models on data from the state of Alabama and treat
  the remaining 50 states as out-of-distribution (OOD) test domains. This
  setting enables us to assess whether our in-distribution worst-case
  subpopulation metric can guide model selection for genuinely new geographic
  contexts. We compare three widely used models: Logistic Regression, XGBoost,
  and Random Forest—trained using cross-entropy loss. Following our framework,
  we evaluate worst-case subpopulation performance over $\worstcov = X$ (all
  features) and analyze robustness across different subpopulation sizes
  $\alpha$.

  Figure~\ref{fig:income_acs} (left panel) reports our worst-case
  subpopulation metric $W_\alpha(\theta)$ computed on the Alabama state. While
  all three models---Logistic Regression, XGBoost, and Random Forest---achieve
  similar average accuracy, they exhibit notable differences in worst-case
  subpopulation performance, with XGBoost demonstrating slightly better
  robustness. This diagnostic is particularly valuable: by analyzing
  worst-case subpopulation performance on in-distribution validation data, we
  can effectively identify models that exhibit more uniform robustness across
  demographic groups present in the training distribution.

  \begin{figure}[h!]
  \centering
  \begin{minipage}[b]{0.49\textwidth}
    \centering \includegraphics[width=0.95\textwidth, height=5cm]{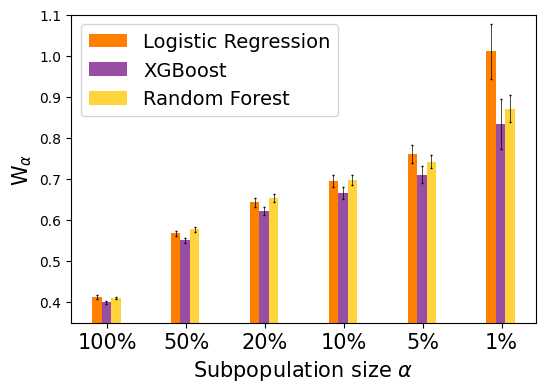}
  \end{minipage}
  \hfill
  \begin{minipage}[b]{0.49\textwidth}
    \centering \includegraphics[width=0.95\textwidth, height=5cm]{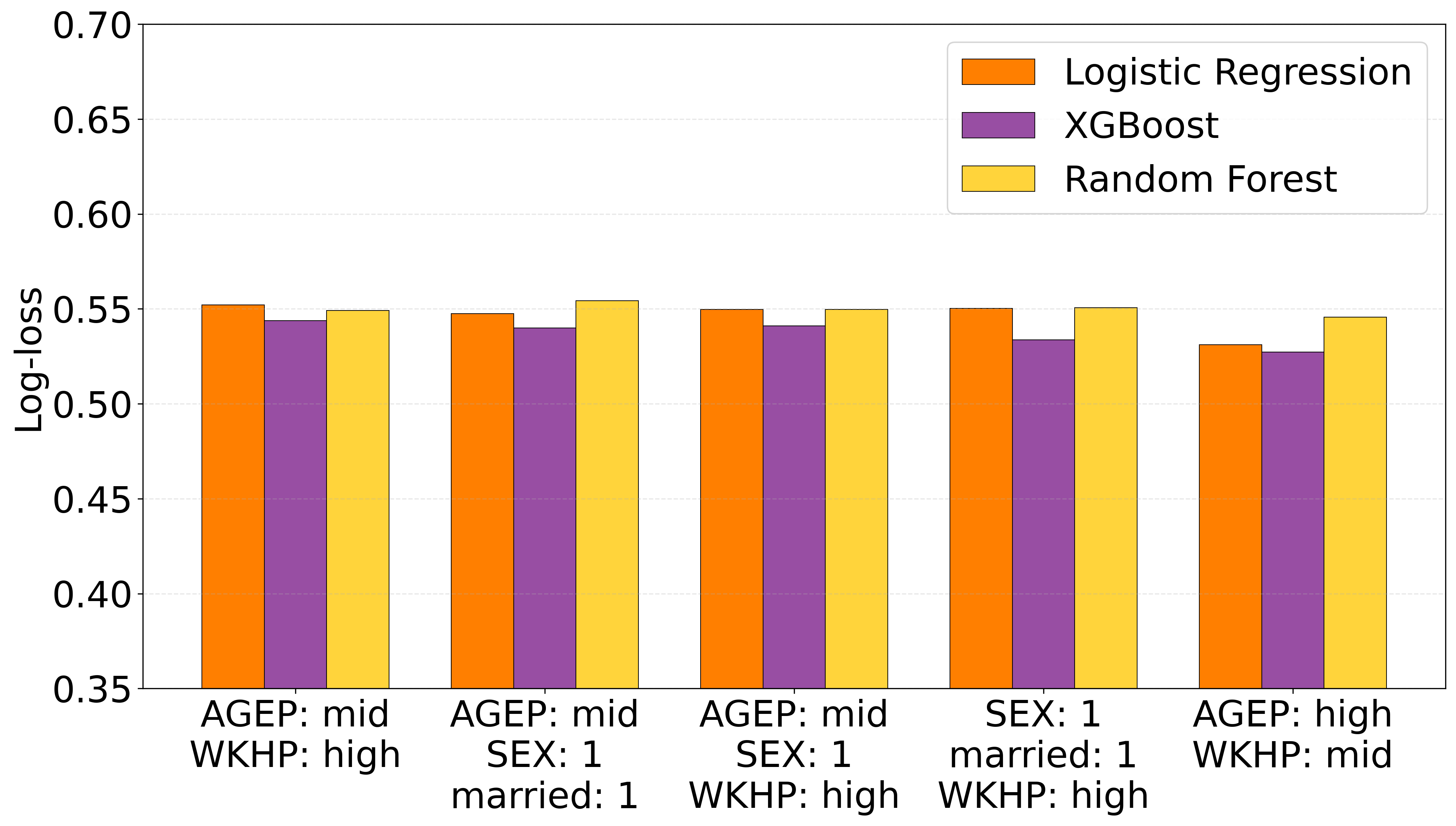}
  \end{minipage}
  \caption{\textbf{ACS Income:} \textbf{(left)} Worst-case subpopulation performance $W_\alpha(\theta)$ with $Z=X$, where $W_{1.0}(\theta) = \mathbb{E}[\ell(\theta(X); Y)]$. \textbf{(right)} Performance on the 5 worst subpopulations (ID) from 62 subpopulations constructed from intersections of top five predictive features (Age, Sex, WKHP, Married, Widowed).}
  \label{fig:income_acs}
\end{figure}

  \begin{figure}[h!]
  \centering
  \begin{minipage}[b]{0.325\textwidth}
  \centering \includegraphics[width=\textwidth]{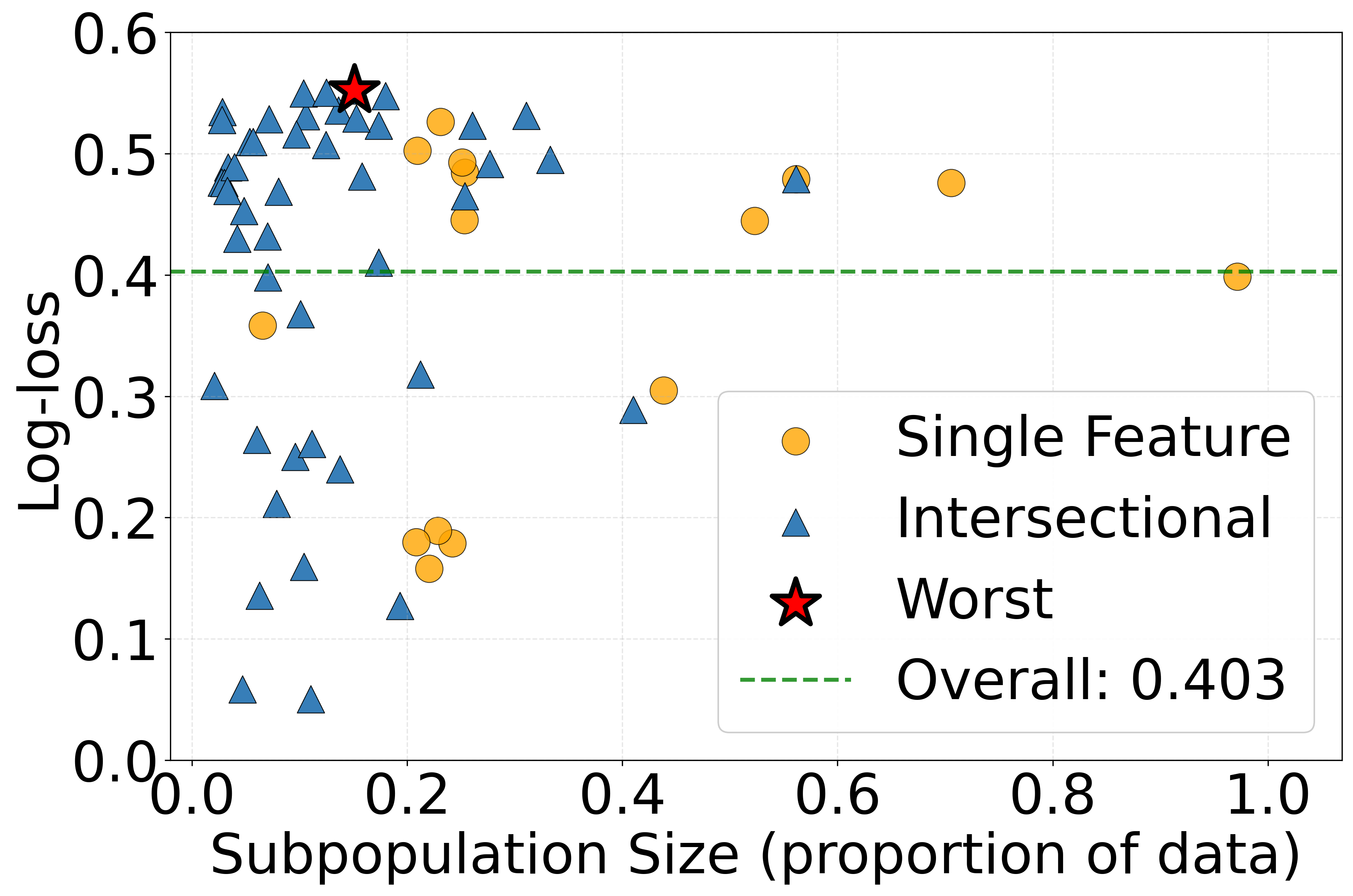}
  \centering{Logistic  Regression}
  \end{minipage}
  \hfill
  \begin{minipage}[b]{0.325\textwidth}
  \centering \includegraphics[width=\textwidth]{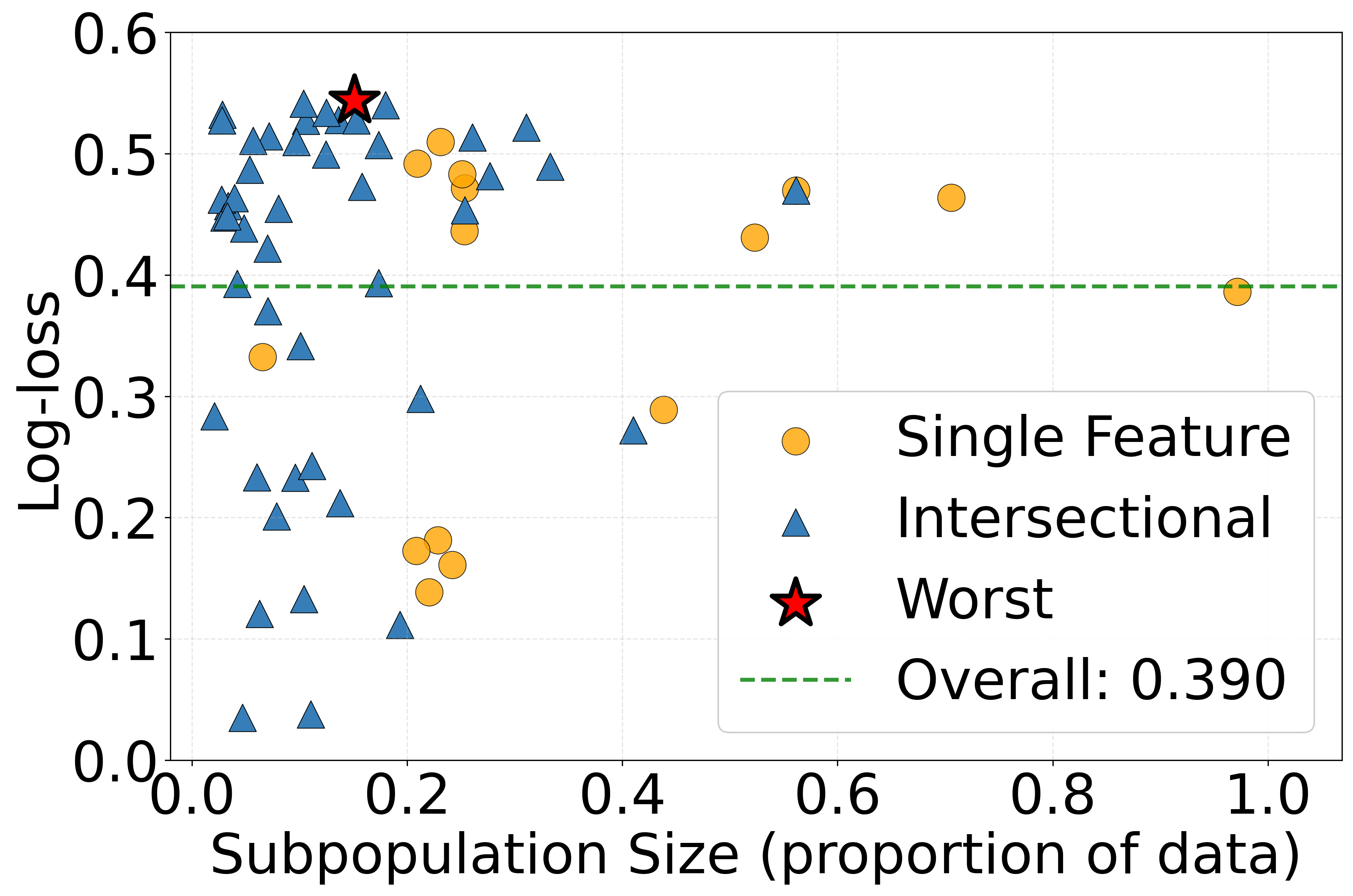}
  \centering{XGBoost}
  \end{minipage}
  \hfill
  \begin{minipage}[b]{0.325\textwidth}
  \centering \includegraphics[width=\textwidth]{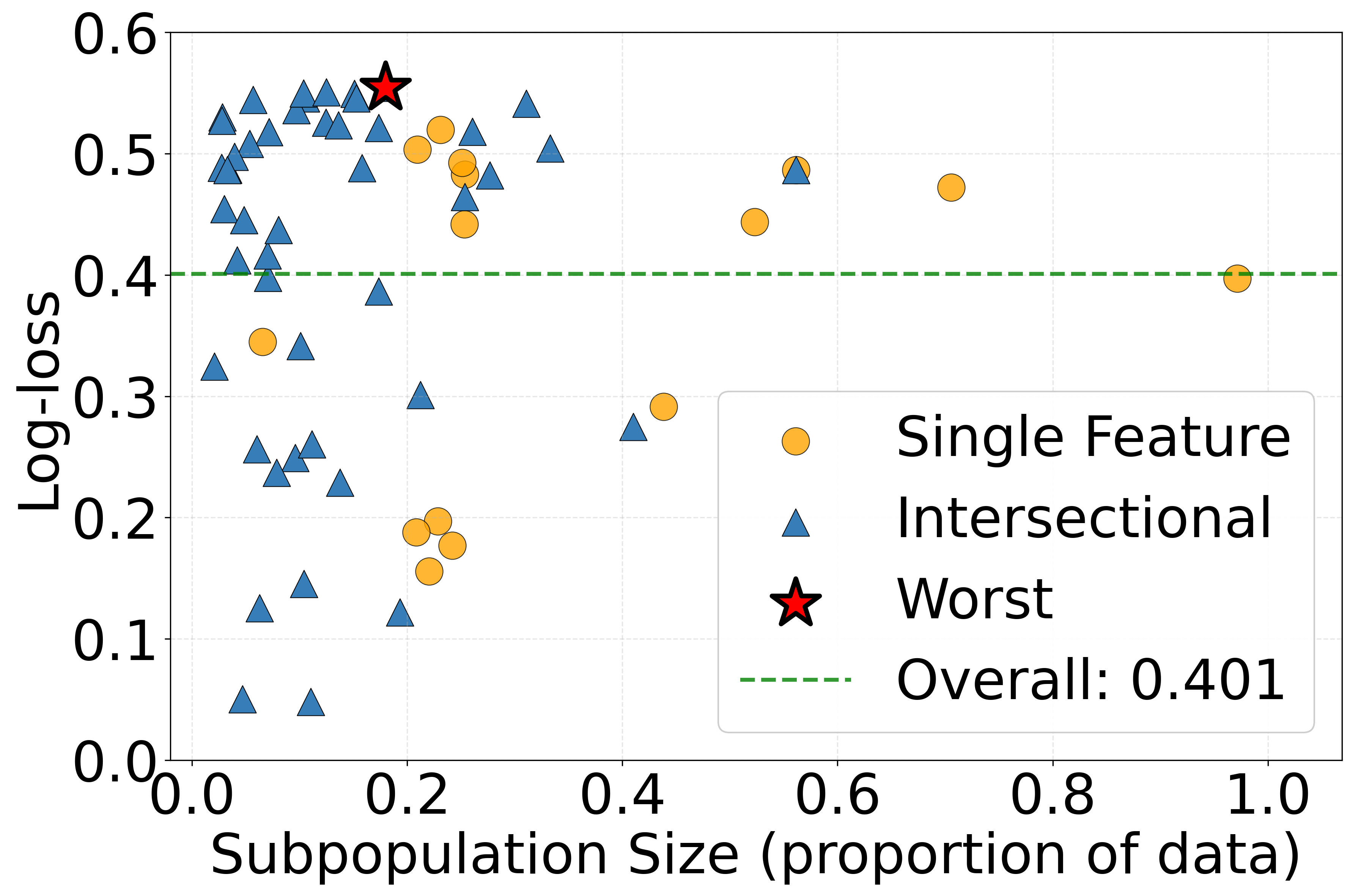}
  \centering{Random Forest}
  \end{minipage}
  \caption{\textbf{ACS Income:} Performance across all 62 in-distribution subpopulations, including single-feature subsets and intersections of top five predictive features (Age, Sex, WKHP, Married, Widowed), for Logistic Regression, XGBoost, and Random Forest models.}
  \label{fig:income_acs_id_subpopulation}
  \end{figure}

  To evaluate the reliability of our diagnostic, we test each model on
  multiple held-out in-distribution subpopulations and assess whether the
  metric $W_\alpha(\theta)$ provides a valid upper bound on performance across
  various subgroups. We construct 62 subpopulations by considering both
  individual demographic features and intersections of the top five predictive
  features (Age, Sex, WKHP, Married, Widowed). For Age, we partition
  individuals into three groups based on quartiles: younger adults aged
  $\leq$35 (\textbf{AGEP\_low}), middle-aged adults aged 36--62
  (\textbf{AGEP\_mid}), and older adults aged $\geq$63
  (\textbf{AGEP\_high}). Similarly, for weekly work hours (WKHP), we create
  three groups at the 25th and 75th percentiles: part-time workers with
  $\leq$32 hours/week (\textbf{WKHP\_low}), typical full-time workers with
  33--45 hours/week (\textbf{WKHP\_mid}), and workers with overtime/long hours
  $\geq$46 hours/week (\textbf{WKHP\_high}).

  As shown in Figure~\ref{fig:income_acs} (right) demonstrates that our metric
  consistently upper bounds the observed performance on the five
  worst-performing subpopulations (among the 62 subpopulations
  considered). Furthermore, Figure~\ref{fig:income_acs_id_subpopulation} shows
  that across all three model types---Logistic Regression, XGBoost, and Random
  Forest---$W_\alpha(\theta)$ provides a reliable upper bound on performance
  degradation across all 62 subpopulations, confirming the diagnostic's
  validity for identifying vulnerable groups.

  \begin{figure}[t]
  \centering \includegraphics[width=0.48\textwidth, height=5cm]{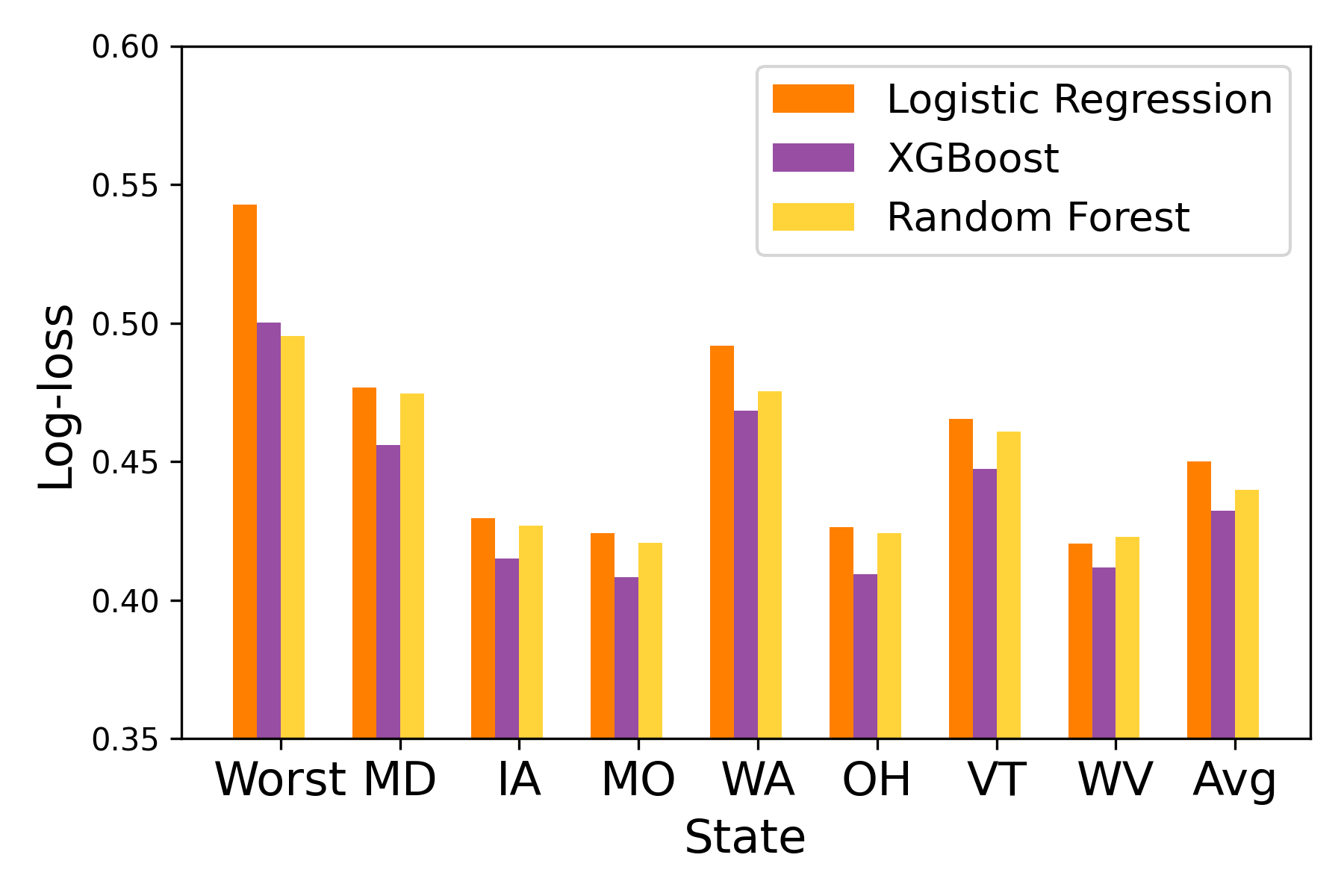}
  \caption{\textbf{ACS Income:}  Performance of the models on different states (OOD) including the worst state (among 50 states) and average across all states.}
  \label{fig:income_acs_ood}
  \end{figure}

\subsubsection{Out-of-distribution Performance}

Figure~\ref{fig:income_acs} (right panel) displays each model’s performance on
the 50 held-out states—representing genuine OOD scenarios—including both the
worst-performing state and the average across all states. Notably, our
worst-case subpopulation metric $W_\alpha$ provides informative predictions of
OOD performance in this ACS Income setting. XGBoost demonstrates superior
performance relative to the other models, and importantly, our metric
consistently upper bounds the observed performance across all the states.

However, in general, the metric cannot guarantee strong performance under
distribution shifts that fall outside the support of the training data. When
states exhibit demographic compositions or feature distributions that differ
substantially from those observed in Alabama, the gap between in-distribution
worst-case predictions and actual OOD outcomes can widen considerably.  Models
deemed robust within Alabama’s subpopulations will indeed perform well on
subpopulations $\tilde{P}_Z$ that lie within the support of the training
distribution $P_Z$but their performance on out-of-support shifts ${P}_Z'$
cannot be guaranteed. Performance on such OOD distributions is governed by two
competing factors:

\begin{enumerate}
\item \textbf{Out-of-support regions of ${P}_Z'$}: Regions containing feature
  combinations or demographic structures absent from training data, where all
  models may perform arbitrarily poorly.

\item \textbf{In-support regions of ${P}_Z'$}: Regions overlapping with the
  training distribution, where models identified as robust by our metric
  continue to outperform less robust alternatives.
\end{enumerate}

Overall OOD performance is determined by the balance between these two
components. In the ACS Income case study, most states share demographic
structures that fall largely within Alabama’s support, enabling our metric to
remain predictive; indeed, the diagnostic still upper-bounds performance
across all these distribution shifts. We next illustrate how shifts beyond the
support affect the reliability of our metric.

  \begin{figure}[t]
  \centering
  \begin{minipage}[b]{0.49\textwidth}
    \centering
    \includegraphics[width=0.95\textwidth, height=5cm]{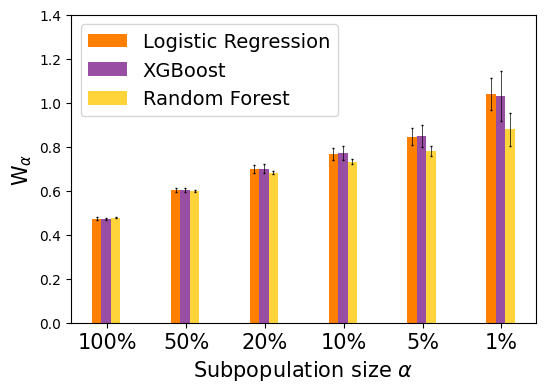}
  \end{minipage}
  \hfill
  \begin{minipage}[b]{0.49\textwidth}
    \centering
    \includegraphics[width=0.95\textwidth, height=5cm]{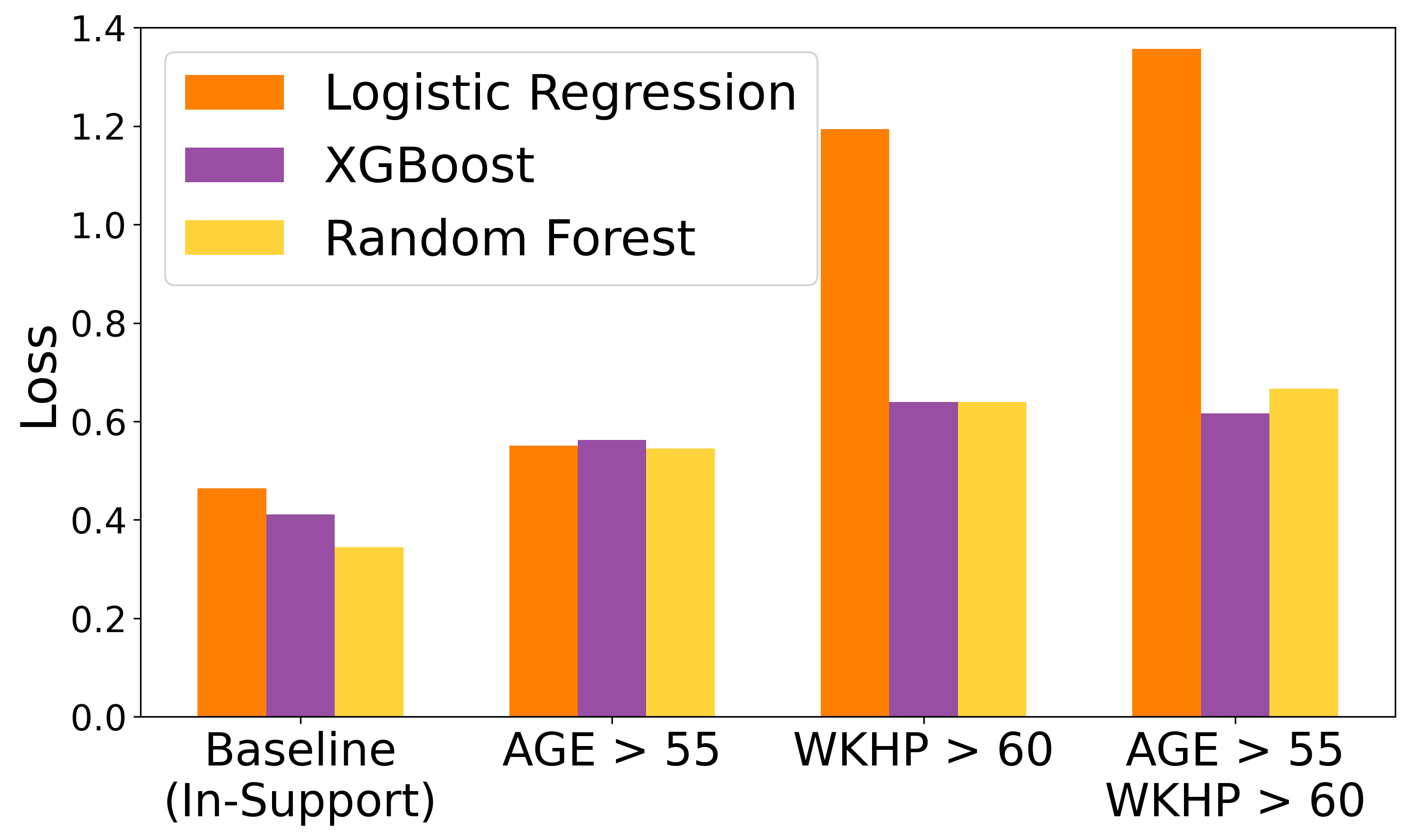}
  \end{minipage}
  \caption{\textbf{ACS Income (out-of-support shifts):} \textbf{(left)} Worst-case subpopulation performance $W_\alpha(\theta)$ with $Z=X$, where $W_{1.0}(\theta) = \mathbb{E}[\ell(\theta(X); Y)]$. \textbf{(right)} Performance on the out-of-support subpopulations. As shown in the logistic regression case, $W_\alpha$ does not guarantee an upper bound under out-of-support distribution shifts.}
  \label{fig:income_acs_out_of_support}
\end{figure}

\subsubsection{Out-of-support OOD shifts}

We now examine how \emph{out-of-support} distribution shifts affect the
validity of our robustness metric. We construct a variant of the ACS income
experiment in which the in-distribution (training) data is restricted to
individuals from the state of Alabama who satisfy the demographic filters
$ 30 \le \text{Age} \le 50, $ $35 \le \text{WKHP} \le 45.  $ Within this
in-distribution region, we evaluate the worst-case subpopulation performance
$W_\alpha(\theta)$ using $Z = X$. The left panel of
Figure~\ref{fig:income_acs_out_of_support} displays the resulting worst-case
subpopulation performance for the three models.

To assess robustness under genuinely out-of-support shifts, we evaluate all
models on a population lying entirely outside the support of the training
distribution, consisting of individuals with $ \text{Age} > 55, $
$ \text{WKHP} > 60.  $ The right panel of
Figure~\ref{fig:income_acs_out_of_support} reports performance in this
out-of-support region.

As the figure shows, our metric no longer provides an upper bound on the true
error for these out-of-support subpopulations for logistic regression. This
behavior is expected: because the shifted population lies entirely outside the
training data support, our sensitivity analysis framework cannot provide
performance guarantees there. In such cases, a model may perform arbitrarily
poorly, as observed for logistic regression. Other models (e.g., tree-based or
ensemble methods) continue to perform reasonably well even under these shifts,
but this behavior is incidental and not guaranteed by the metric.

This experiment highlights a central conceptual point: our sensitivity
analysis framework is a diagnostic tool for evaluating robustness to
subpopulation shifts within the support of the training distribution. It is
not a formal safeguard against performance degradation under unforeseen,
genuinely out-of-support distribution shifts.}

\subsection{Functional Map of the World (FMoW)}
\label{sec:fmow}

Training robust models has garnered significant attention in both the
operations research and machine learning communities. However, existing
approaches that directly enforce robustness often struggle to scale to modern
machine learning or deep learning settings, where models such as deep neural
networks contain hundreds of thousands—or even millions—of parameters. In
contrast, our framework focuses exclusively on evaluation, enabling us to
assess and certify the robustness of large-scale models that are otherwise
difficult to train robustly. In our second case study, we apply our method to
an image classification task and demonstrate its utility in evaluating the
robustness of state-of-the-art, large-scale models, including deep neural
networks.

{ 
The dataset used in this case study reflects real-world
  spatiotemporal distribution shifts, where models must generalize across
  different geographic regions and time periods. This setting allows us to
  validate our metric under realistic distribution shifts and assess its
  effectiveness in diagnosing robustness in deep learning models.  }

\subsubsection{Background}
We study a satellite image classification problem~\citep{ChristieFeWiMu18}. 
Satellite images can impact economic and environmental policies globally by
allowing large-scale measurements on poverty~\citep{AbelsonKuSu14},
population changes, deforestation, and economic 
growth~\citep{HanAhPaYaLeKiYaPaCh20}.  
It is therefore important to implement automated approaches that 
allow scientists and sociologists to provide continuous monitoring of
land usage and analyze data from remote regions at a relatively low cost
with models that perform reliably across time and space. 

We consider the Functional Map of the World (FMoW) dataset~\citep{ChristieFeWiMu18} 
comprising of satellite images, where the goal is to predict building / land
usage categories (62 classes). We take a recently published variant of this
dataset in~\citet{HuangLiVaWe17,KohSaEtAl20}, FMoW-WILDS,
that is designed specifically for evaluating model performance under
{\it temporal} and {\it spatial} distribution shifts. Due to the scale of the
dataset ($>$10K images), traditional robust training approaches do not scale;
we will show that our evaluation metric is fully scalable,
and our metric provides insights on the performances of SOTA
deep learning neural network models on future unseen data. 

We take the SOTA models reported in~\citet{KohSaEtAl20} of
FMoW as our benchmark. These models are deep neural networks
that benefit from \emph{transfer learning}: unlike traditional models
that are trained from scratch to solve the problem on-hand,
these benchmark models are built on \emph{pre-trained models} --- 
existing models whose parameters have been pre-trained on other,
usually massive datasets such as \emph{ImageNet} --- 
and then adapted to the present problem through parameter fine-tuning. 
It is observed that transfer models exhibit steller performance
across numerous datasets and matching, if not surpassing, that of
SOTA models trained from scratch for the specific problem. 
However, although these SOTA benchmark models on FMoW
achieve satisfactory out-of-sample ID accuracy on the past data,
all suffer from significant performance drop on the future OOD data,
suggesting that these benchmark models are not robust to temporal shifts.

Recently, \citet{RadfordKiEtAl21} published a self-supervising model, 
\emph{Contrastive Language-Image Pre-training (CLIP)},
that has been observed to exhibit many exciting robustness properties. 
\emph{CLIP} models are pre-trained on 400M image-text pairs using
natural language supervision and contrastive losses. The pre-training data
for \emph{CLIP} is 400 times bigger than \emph{ImageNet}, the standard
pre-training dataset used for FMoW benchmark models,
and~\citet{RadfordKiEtAl21} have observed that CLIP exhibits substantial
\emph{relative robustness gains} over other methods on natural distribution
shifts of \emph{ImageNet}. Previous work in~\citet{WortsmanIlLiKiHaFaNaSc21} 
has shown significant performance gains using \emph{CLIP}-based models on FMoW. 

Motivated by the challenges and opportunities, we take the perspective
of a practitioner, presented with different but comparable models and
with historical data, wishing to select the ``best" model that perform
well across all subgroups into the future. 
More concretely, we will illustrate the use of our procedure on models 
that achieve similar average accuracy and loss and are indistinguishable
in the view of traditional model diagnostic tools.  
We will demonstrate that our procedure is able to select the most robust
models that perform best on future data across different subpopulations.

\subsubsection{Problem Specifics}
The input $X_i$ is an RGB satellite image, each pixel represented by a vector in $\mathbb{R}^3$, 
and the label $Y_i \in \{0, 1, 2, \cdots, 61\}$ represents one of the 62
land use categories. We consider the following three non-overlapping subsets
of the data based on image taken times: 

\begin{itemize}
\item Training: select images in 2003 -- 2013 (76,863 images)
\item Validation: select images in 2003 -- 2013 (11,483 images)
\item Test:  images in 2016 -- 2018 (22,108 images)
\end{itemize}

We consider training and validation images (images from pre-2013) as
in-distribution (ID) data, and test images in taken between 2016 and 2018
as out-of-distribution (OOD) data. Each image also comes with meta information,
including the (longitude, latitude) location of which the photo was taken,
the continent/region information, and the weather information
(cloudiness on a scale from 1 to 10)~\citep{HuangLiVaWe17,KohSaEtAl20}. 
We define subpopulations based on meta information.

\begin{figure}[ht]
  \centering
     \begin{subfigure}[b]{0.49\textwidth}
         \centering
         \includegraphics[width=\textwidth, height=5cm]{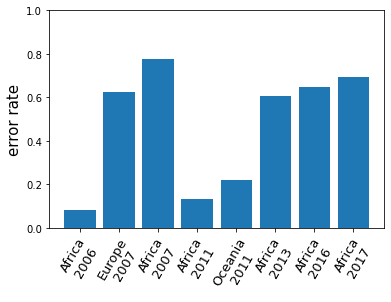}
     \end{subfigure}
     \begin{subfigure}[b]{0.49\textwidth}
         \centering
         \includegraphics[width=\textwidth, height=5cm]{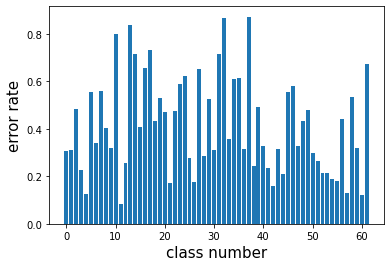}
        \end{subfigure}
        \vspace{-5pt}
     \caption{For \emph{DenseNet} ERM, spatiotemproal intersectionality (left) and
       performance by class (right)}
      \label{fig:fmow}
\end{figure}

Similarly to the Warfarin example, we observe that model performances remain
similar either temporally or spatially when each dimension is considered 
\emph{separately}, but there is substantial variability across intersections
of region and year. For a standard benchmark deep neural network model 
in~\citet{HuangLiVaWe17,KohSaEtAl20} that achieves near-SOTA performance,
we present these trends in Figure~\ref{fig:fmow}(a). Furthermore, in
Figure~\ref{fig:fmow}(b), we observe substantial variability in error rates
across different labels, indicating there is a varying level of difficulty in
classifying different classes. (We observed similar patterns for other models.)

Because of significant variation in difficulty in learning different classes, 
we consider the core attributes $\worstcov$ on which the subpopulation is
defined to be the full meta information vector concatenated with the true class label $y$. 
As we impose no assumption on $\worstcov$, we could also use the semantic
meaning of each class labels in place of categorical variables. We will go
back to this point in Section~\ref{sec:fmow-ext}.

We take three benchmark SOTA transfer learning models (all pretrained on
\emph{ImageNet}). Two of the three models come from~\citet{HuangLiVaWe17,KohSaEtAl20}.
Both models are \emph{DenseNet}-121 based models, but trained with
different objectives:
\begin{itemize}
\item \emph{DenseNet}-121 ERM: the model is trained on FMoW to
minimize the usual average training loss (mean squared loss);
\item \emph{DenseNet}-121 IRM: the model is trained on FMoW to
minimize the invariant risk (the loss adds an extra penalty term that 
penalizes feature distributions that have different optimal
linear classifiers for each domain), proposed in~\citet{ArjovskyBoGuLo20}.
\end{itemize} 
We consider a third model, the \emph{ImageNet}-pretrained Dual Path Network model 
(\emph{DPN}-68 model)~\citep{ChenLiXiJiYaFe17}.
This model has a different architecture from that of \emph{DenseNet} models.
All these models achieve an out-of-sample ID accuracy of $60\%$.

However, these SOTA benchmark models suffer from significant performance
drop on the OOD data: all models suffer from a performance drop of $~7\%$
in average accuracy, and this performance drop increases to a stunning $30\%$
for images coming from Africa (the region where all model perform the achieve
the lowest accuracy, which we will call the worst-case region). These
observations reveal that there is natural distribution shift from past data
to future unseen data. It also raises the flag that despite achieving great
ID performance, these SOTA benchmark models are not robust against these shifts.

On the other hand, \emph{CLIP} models have exhibited promising robustness
properties~\citep{RadfordKiEtAl21}. To adapt CLIP-based models to the
satellite image classification probelm, we adopt a weight-space ensembling
method (\emph{CLIP WiSE-FT}) in~\citet{WortsmanIlLiKiHaFaNaSc21}. This method
has been observed to exhibit large Pareto improvements in the sense that it
leads to a suite of models with improved performance with respect to both ID
and OOD accuracy. Motivated by the observed robustness gains, we consider the
\emph{CLIP WiSE-FT} model that achieves comparable performance on the FMoW ID
validation set to the benchmark \emph{ImageNet} pre-trained models.
Appendix~\ref{section:experiments-details} provides additional details on the
experimental settings and training specifications.

\paragraph{Goal: } 
Presented with both benchmark \emph{ImageNet}-based models
and \emph{CLIP WiSE-FT}, our goal as the practitioner is to choose
the one that generalizes best in the future unseen OOD test data
uniformly across all subpopulations.

\paragraph{Challenge: }
In the view of traditional model diagnostic tools, these models are
indistinguishable as they achieve comparable average ID accuracy and ID loss. 

Nevertheless, our proposed method is able to select models that
perform well ``in the future'' without requiring OOD data, and
at the same time our method raises awareness of difficulty in domain
generalization. We report the experiment results in the next section.

We compute estimators of $W_\alpha(\theta)$ (Algorithm~\ref{alg:two-stage}) on
the ID validation data using standard cross entropy loss. To validate that our
metric reliably captures in-distribution worst-case subpopulation performance,
we evaluate each model's actual performance across different spatiotemporal
subpopulations (defined by region, year, and class) and verify that our metric
provides a tight upper bound on the true worst-case performance.

\begin{figure}[htbp]
  \centering
     \begin{subfigure}[b]{0.48\textwidth}
         \centering
         \includegraphics[width=\textwidth]{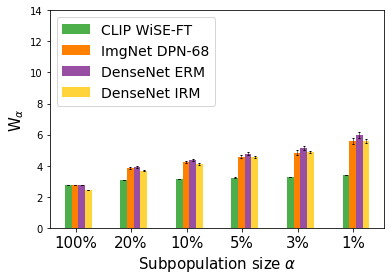}
     \end{subfigure}
     \begin{subfigure}[b]{0.48\textwidth}
         \centering
         \includegraphics[width=\textwidth]{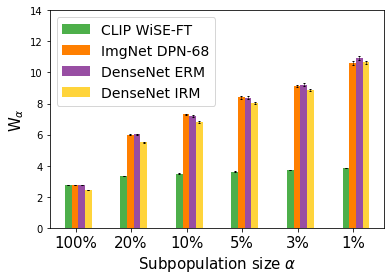}
        \end{subfigure}
        \caption{Left: $Z$ = (all metadata); Right: $Z$ = (all metadata, $Y$).
          Results are averaged over 50 random seeds with error bars
          corresponding to a 95$\%$ confidence interval over the random runs.}
        \label{fig:fmow-loss-all}
\end{figure}

In Figure~\ref{fig:fmow-loss-all}, we summarize the estimated worst-case
subpopulation performance $W_\alpha(\theta)$ across different subpopulation
sizes $\alpha$.  All models achieve comparable average ID accuracy of
$\sim 60\%$, with \emph{DenseNet} IRM having the best average-case cross
entropy loss. However, the metric $W_\alpha(\theta)$ reveals substantial
differences in worst-case subpopulation robustness: the \emph{ImageNet}
pre-trained models show significantly worse performance compared to \emph{CLIP
  WiSE-FT}, with this gap growing larger as the subpopulation size $\alpha$
becomes increasingly small.

This demonstrates a key value of our metric: it can identify models that
maintain more uniform robustness across in-distribution
subpopulations. Evaluations on worst-case subpopulations clearly show that
\emph{CLIP WiSE-FT} exhibits superior robustness against subpopulation shifts;
in contrast, average-case evaluations would incorrectly select \emph{DenseNet}
IRM. Our metric successfully distinguishes between models that appear
equivalent under traditional diagnostic tools but exhibit very different
performance on tail subpopulations.

We further observe a drastic performance deterioration on tail subpopulations
across all models.  The inclusion of label information in $\worstcov$
significantly deteriorates worst-case performance, demonstrating that our
metric reliably captures performance degradation even when accounting for
label distribution changes.

{
To further assess whether our metric reliably captures
  in-distribution subpopulation performance, we report detailed region-wise
  results in Table~\ref{tbl:fmow_id_regions}. In addition,
  Table~\ref{tbl:loss_2007_2012} provides more granular analyses broken down
  by year and region for representative years within the in-distribution
  period. These tables reveal that, despite the models exhibiting similar
  average in-distribution loss (approximately 2.8) and accuracy (around 60\%),
  their performance varies substantially across geographic regions and
  years. Moreover, models that rank highly according to our metric (e.g., CLIP
  WiSE-FT) display noticeably more uniform performance across both regions and
  years (see Table~\ref{tbl:loss_2007_2012}) compared to lower-ranked
  models. Crucially, our metric also upper bounds the loss observed for each
  model across all regions and years. }

\begin{table}[h]
\centering
\begin{tabular}{c|cc|cc|cc|cc}
   \midrule
& \multicolumn{2}{c|}
 {\textbf{CLIP WiSE-FT}} & \multicolumn{2}{c|}
 {\textbf{DenseNet ERM}} & \multicolumn{2}{c|}
 {\textbf{DenseNet IRM}} & \multicolumn{2}{c}
 {\textbf{DPN-68}} \\
   \cmidrule(lr){2-3} \cmidrule(lr){4-5} \cmidrule(lr){6-7} \cmidrule(lr){8-9}
  Region  & Acc & Loss & Acc & Loss & Acc & Loss & Acc & Loss \\
  \midrule
  Asia & 0.612 & \textbf{2.82} & 0.608 & 2.89 & 0.595 & 2.32 & 0.615 & 2.79 \\
  Europe & 0.594 & 2.81 & 0.590 & \textbf{2.90} & 0.564 & \textbf{2.65} & 0.589 & \textbf{2.93} \\
  Africa & 0.674 & 2.62 & 0.700 & 2.34 & 0.679 & 2.11 & 0.663 & 2.58 \\
  Americas & 0.627 & 2.71 & 0.638 & 2.56 & 0.614 & 2.29 & 0.635 & 2.50 \\
  Oceania & 0.721 & 2.60 & 0.729 & 2.08 & 0.729 & 1.66 & 0.749 & 1.97 \\
  \bottomrule
\end{tabular}
\caption{\label{tbl:fmow_id_regions} Region-wise performance on ID validation set.
Performance across regions are comparable across different models, validating that our worst-case
metric captures true subpopulation variations within the in-distribution data. Worst-case losses (per model) are highlighted in bold.}
\end{table}

\begin{table*}[h!]
\centering
\small
\begin{tabular}{l|rr|rr|rr|rr}
  \toprule
  & \multicolumn{2}{c|}{\textbf{CLIP WiSE-FT}} 
  & \multicolumn{2}{c|}{\textbf{DenseNet ERM}} 
  & \multicolumn{2}{c|}{\textbf{DenseNet IRM}} 
  & \multicolumn{2}{c}{\textbf{DPN-68}} \\
  \cmidrule(lr){2-3} \cmidrule(lr){4-5} \cmidrule(lr){6-7} \cmidrule(lr){8-9}
  Region & 2007 & 2012 & 2007 & 2012 & 2007 & 2012 & 2007 & 2012 \\
  \midrule
  Asia     & 2.83 & 2.71 & 2.98 & 2.43 & 2.80 & 1.95 & 2.95 & 2.29 \\
  Europe   & 3.03 & 2.75 & 4.16 & 2.79 & 3.91 & 2.38 & \textbf{5.02} & 2.76 \\
  Africa   & \textbf{3.31} & \textbf{2.88} 
           & \textbf{4.98} & \textbf{3.49} 
           & \textbf{6.28} & \textbf{2.93} 
           & 4.19 & \textbf{3.54} \\
  Americas & 2.75 & 2.69 & 2.60 & 2.41 & 2.96 & 2.20 & 2.83 & 2.38 \\
  Oceania  & 2.84 & 2.61 & 1.66 & 2.96 & 2.05 & 1.95 & 4.21 & 2.90 \\
  \bottomrule
\end{tabular}
\caption{\label{tbl:loss_2007_2012} Model performance (cross entropy loss) by region for years 2007 and 2012 under each model. Worst-case losses (per model and year) are highlighted in bold.}
\end{table*}

\subsubsection{Out-of-distribution Performance}


{
Table~\ref{tbl:fmow_ood_regions} reports model performance on
  out-of-distribution (OOD) data collected between 2016 and 2018. All models
  experience significant performance degradation under this temporal
  distribution shift, with the most pronounced drop—up to 20 percentage points
  in predictive accuracy—occurring for images collected in Africa. For a more
  detailed view, Table~\ref{tbl:loss_2016_2017} presents year- and region-wise
  performance metrics for representative years in the OOD test period (2016
  and 2017). Among the evaluated models, \emph{CLIP WiSE-FT} consistently
  demonstrates superior performance across all regions and years. Although all
  models exhibit considerable degradation—particularly in the Africa
  region—\emph{CLIP WiSE-FT} shows enhanced robustness, maintaining relatively
  stable performance even under substantial distribution shifts.

  Importantly, despite evaluating on OOD test sets where the in-support status
  of each image is not explicitly known, the observed performance trends align
  with the predictions of our diagnostic metric
  (Figure~\ref{fig:fmow-loss-all}). In particular, \emph{CLIP WiSE-FT}, which
  ranks highly under our metric, also performs best in OOD
  settings. Additionally, the worst-case losses observed during evaluation
  remain upper bounded by our metric, suggesting that much of the OOD data
  likely falls within the support of the training distribution.
}

\begin{table}[h]
\centering
\begin{tabular}{c|cc|cc|cc|cc}
  \cmidrule(lr){1-9}
  & \multicolumn{2}{c|}{\textbf{CLIP WiSE-FT}} & \multicolumn{2}{c|}{\textbf{DenseNet ERM}} & \multicolumn{2}{c|}{\textbf{DenseNet IRM}} & \multicolumn{2}{c}{\textbf{DPN-68}} \\
  \cmidrule(lr){2-3} \cmidrule(lr){4-5} \cmidrule(lr){6-7} \cmidrule(lr){8-9}
  Region & Acc & Loss & Acc & Loss & Acc & Loss & Acc & Loss \\
  \midrule
  Asia & 0.583 & 2.85 & 0.543 & 3.17 & 0.519 & 2.70 & 0.555 & 3.26 \\
  Europe & 0.580 & 2.80 & 0.554 & 3.26 & 0.533 & 2.78 & 0.553 & 3.28 \\
  Africa & 0.379 & \textbf{3.08} & 0.331 & \textbf{5.41} & 0.308 & \textbf{4.46} & 0.309 & \textbf{5.61} \\
  Americas & 0.575 & 2.79 & 0.560 & 3.29 & 0.538 & 2.74 & 0.553 & 3.31 \\
  Oceania & 0.661 & 2.68 & 0.574 & 3.29 & 0.556 & 2.49 & 0.566 & 2.93 \\
  \bottomrule
\end{tabular}
\caption{\label{tbl:fmow_ood_regions} Region-wise performance on OOD test set.
\emph{CLIP WiSE-FT} maintains superior robustness across all regions. Catastrophic
performance degradation is evident for all models on Africa (31--38\% accuracy),
whereas other regions maintain 54--66\% accuracy.  Worst case performance (by loss) is highlighted for each model.}
\end{table}

\begin{table*}[h!]
\centering
\small
\begin{tabular}{l|rr|rr|rr|rr}
  \toprule
  & \multicolumn{2}{c|}{\textbf{CLIP WiSE-FT}}
  & \multicolumn{2}{c|}{\textbf{DenseNet ERM}}
  & \multicolumn{2}{c|}{\textbf{DenseNet IRM}}
  & \multicolumn{2}{c}{\textbf{DPN-68}} \\
  \cmidrule(lr){2-3} \cmidrule(lr){4-5} \cmidrule(lr){6-7} \cmidrule(lr){8-9}
  Region & 2016 & 2017 & 2016 & 2017 & 2016 & 2017 & 2016 & 2017 \\
  \midrule
  Asia     & 2.88 & 2.80 & 3.39 & 2.77 & 2.89 & 2.35 & 3.43 & 2.95 \\
  Europe   & 2.77 & 2.93 & 3.08 & 4.12 & 2.62 & 3.56 & 3.13 & 3.97 \\
  Africa   & \textbf{3.08} & \textbf{3.07}
           & \textbf{5.30} & \textbf{5.50}
           & \textbf{4.27} & \textbf{4.61}
           & \textbf{4.90} & \textbf{6.15} \\
  Americas & 2.77 & 2.84 & 3.24 & 3.46 & 2.68 & 2.96 & 3.22 & 3.62 \\
  Oceania  & 2.64 & 2.97 & 3.17 & 4.33 & 2.39 & 3.36 & 2.83 & 3.89 \\
  \bottomrule
\end{tabular}
\caption{\label{tbl:loss_2016_2017} Cross entropy loss by region for years 2016 and 2017 under each model. Worst-case losses (per model and year) are highlighted in bold.}
\end{table*}

{

  These findings reinforce the central message of our work: the sensitivity
  analysis framework serves as a diagnostic tool for uncovering model
  vulnerabilities to subpopulation shifts within the support of the training
  distribution. Accordingly, the metric provides insight into a model’s
  robustness under distribution shifts that remain within the training data’s
  support. This includes many real-world scenarios and naturally accommodates
  complex intersectional structures, as it operates without requiring explicit
  demographic or region-based groupings—demonstrated in both the ACS and FMoW
  case studies.  However, as previously noted, our framework does not offer
  robustness guarantees for distribution shifts that lie outside the training
  distribution's support. The purpose of the case studies is not to position
  our diagnostic as a universal solution to distribution shift, but rather to
  offer a practical and grounded illustration of its capabilities and
  limitations. By examining realistic distribution shifts, we highlight the
  types of actionable insights that worst-case subpopulation analysis can
  yield.

  The practical takeaway is that practitioners should apply our diagnostic in
  tandem with domain expertise. In scenarios where out-of-distribution (OOD)
  data is expected to involve genuinely novel feature combinations or
  structural changes, additional robustness strategies—beyond worst-case
  subpopulation analysis—will likely be required. Understanding how geographic
  or demographic shifts deviate from the training distribution helps clarify
  both the strengths and boundaries of our approach.

\subsection{Flexibility in the choice of $\worstcov$}\label{sec:fmow-ext}
A key strength of our framework lies in the flexibility of the choice of
\worstcov \worstcov, which enables the modeler to define subpopulations at
varying levels of granularity.   We demonstrate this flexibility
  across both the ACS Income and FMoW datasets.

  \textbf{ACS Income:} We extend our earlier ACS Income experiments to examine
  how different choices of $Z$ (i.e., subpopulation-defining attributes)
  impact the worst-case subpopulation metric $W_\alpha$.
  Figure~\ref{fig:z_importance} shows $W_\alpha$ (for $\alpha=40\%$) computed
  under different subsets of demographic features. The results reveal that
  different selections of $Z$ lead to varying levels of worst-case
  subpopulation performance. Interestingly, the trend in $W_\alpha$ aligns
  closely with the predictive importance of the selected features. As shown in
  Figure~\ref{fig:fea_importance}, where we plot feature importance values
  derived from a random forest classifier, features such as Age (AGEP),
  Working hours per week (WKHP), and Sex (SEX) rank highest in predicting
  income $Y$. Notably, attributes with greater predictive importance
  correspond to lower worst-case subpopulation performance—indicating higher
  vulnerability to subgroup-specific errors. Moreover, increasing the
  intersectionality of features (i.e., considering multiple features jointly
  in $Z$) leads to further degradation in worst-case performance.

\begin{figure}[h!]
    \centering
    \includegraphics[width=0.8\linewidth]{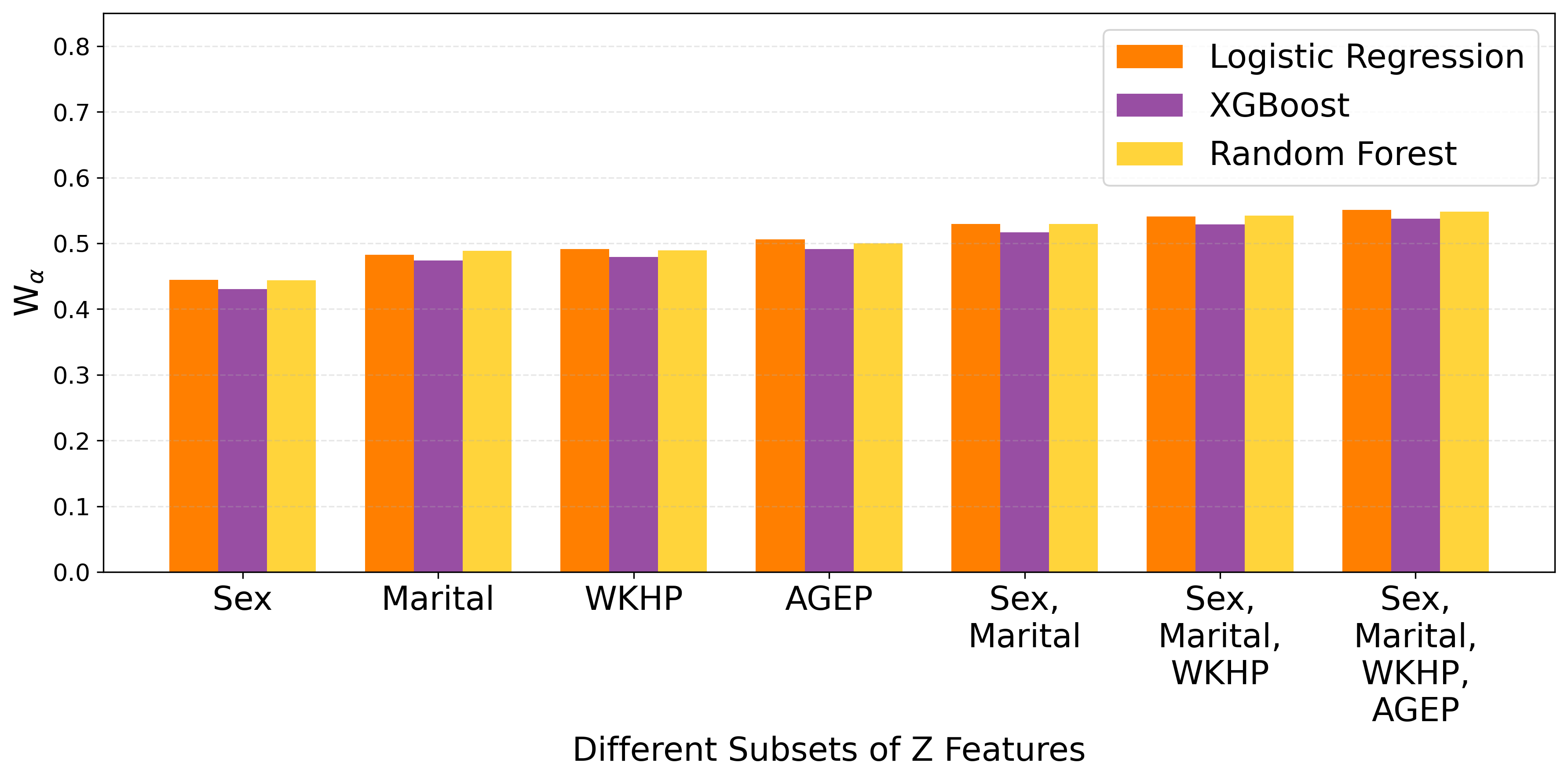}
    \caption{Worst-case subpopulation performance $W_\alpha(\theta)$ under different set of $Z's$ with $\alpha = 40\%$.  Here AGEP $:=$ Age,  WKHP $:=$  Working hours per week,  SEX $:=$ Sex, and  $\text{Marital} :=\{\text{married, widowed, divorced, separated, never}\}$}
    \label{fig:z_importance}
\end{figure}

\begin{figure}[h!]
    \centering
    \includegraphics[width=0.8\linewidth]{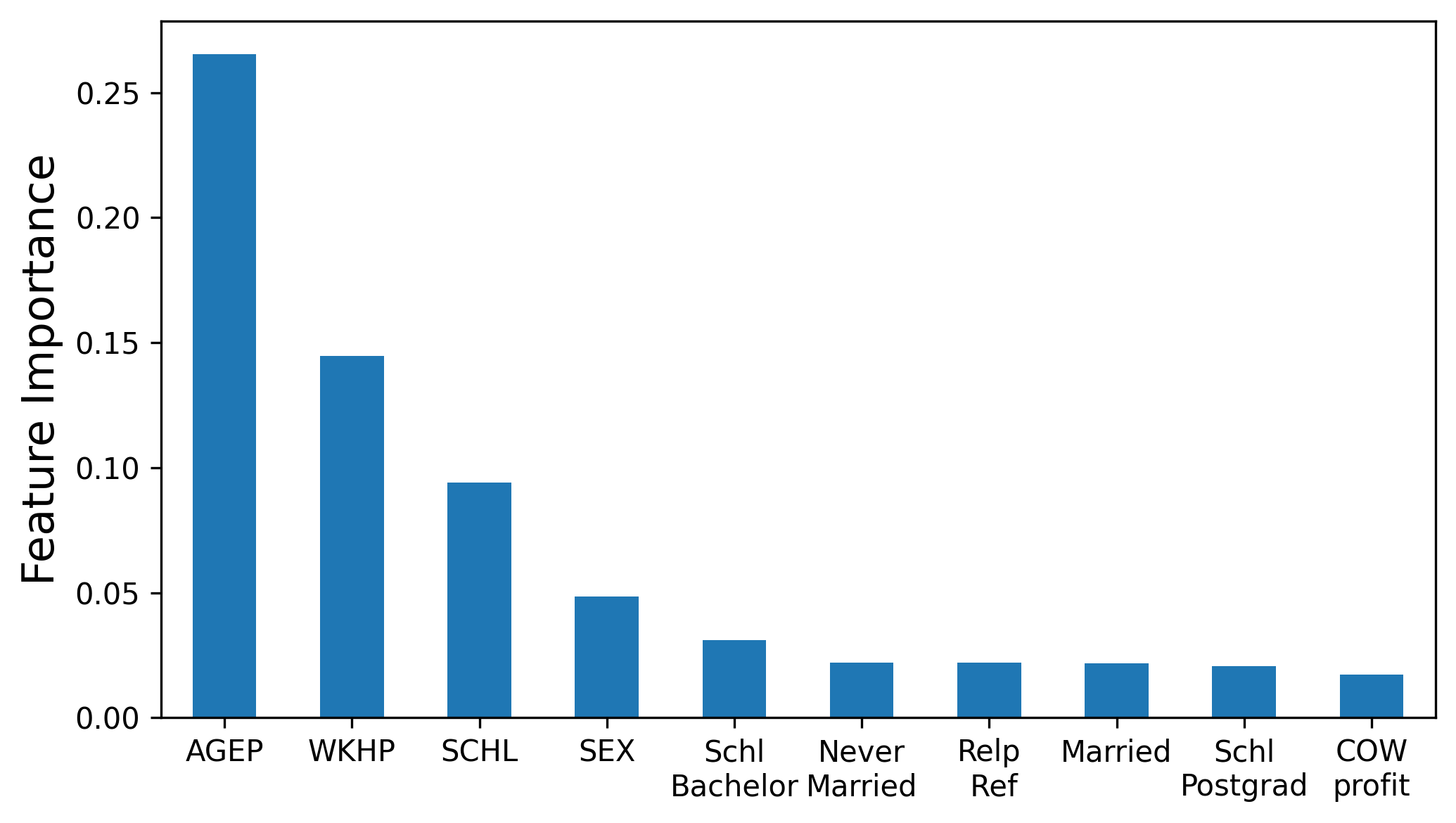}
    \caption{Top 10 features in predicting $Y$ from $X$ (ACS income). }
    \label{fig:fea_importance}
\end{figure}

 \textbf{FMoW:} As defining subpopulations over all metadata can be conservative, we present
additional results under various subsets  of meta-data $\{\text{Year, Region, Lat, Lon, Cloud Cover}\}$ in
Appendix~\ref{section:experiments-details}. 
 These experiments reveal that models show limited robustness when 
$Z$ includes spatial features such as Latitude and Longitude, or the label $Y$. This suggests that performance on subpopulations defined by geographic location (e.g., latitude/longitude) or by class label varies significantly, further emphasizing the need for flexible subgroup definitions in robustness evaluation.

\textbf{Text Embeddings:} Another advantage of our framework is that it allows for semantically informed subgroup definitions via text embeddings. Instead of treating labels as categorical variables, we can generate natural language descriptions of class labels by appending them to engineered prompts and encoding them using the CLIP text encoder \citep{RadfordKiEtAl21}. This yields semantically meaningful feature representations for the labels, which can then be used in defining subpopulations through 
\worstcov. 

The rationale behind using text embeddings is to capture semantic relationships between label values that one-hot encodings cannot represent. For example, suppose $Z = \{Y\}$ with $Y \in \{\text{Cat, Dog, Car, Airplane}\}$. A one-hot representation fails to encode any similarity between these classes. In contrast, text embeddings preserve semantic structure, capturing that “Cat” is closer to “Dog,” and “Car” is closer to “Airplane.” This enables more nuanced subgroup formation and improves the estimation of conditional expectations such as $\mathbb{E}[\ell(\theta(X), Y)|Z]$. 

In Appendix~\ref{section:modelflexibility}, we provide evaluation results demonstrating how substituting label 
$Y$ with its corresponding embedding vector when defining 
\worstcov enhances the granularity and interpretability of our robustness analysis.

}


\section{Connections to coherence and distributional robustness}
\label{section:extension}



The worst-case subpopulation performance~\eqref{eqn:cvar} only considers
subpopulations that comprise $\alpha$-fraction of the data. In this section,
we propose a generalized measure of model robustness that considers
subpopulation sizes on various scales. We show that our generalized notion of
worst-case subpopulation performance is closely related to coherent risk
measures and distributional robustness. Concretely, instead of a single
subpopulation size, we take the average over $\alpha \in (0, 1]$ using a
probability measure $\qa$
\begin{equation}
  \label{eqn:multi-scale}
  \int_{(0, 1]} \worstsub(\condrisk) d\qa(\alpha).
\end{equation}
The multi-scale average~\eqref{eqn:multi-scale} introduces substantial
modeling flexibility by incorporating prior beliefs on which subpopulation
sizes are of higher concern. It can consider arbitrarily small subpopulations
(see Proposition~\ref{prop:hcvar} to come), and by letting
$\qa = t \delta\{\alpha\} + (1-t) \delta \{1\}$ for $t \in [0,1]$, it interpolates between
average and the worst subpopulation performance $\worstsub(\condrisk)$.

Equipped with the multi-scale average~\eqref{eqn:multi-scale}, we define the
\emph{generalized worst-case subpopulation performance} over a (nonempty)
class $\qaset$ of probability measures on the half-open interval $(0, 1]$
\begin{equation}
  \label{eqn:gen}
  \mathsf{W}_{\qaset}(\model)
  \defeq \sup_{\qa \in \qaset} \int_{(0, 1]}
  \mathsf{W}_{\alpha}(\model) d\qa(\alpha).
\end{equation}
We show an equivalence between generalized worst-case subpopulation
performances~\eqref{eqn:gen} and \emph{coherent risk measures}, an axiomatic
definition of risk-aversion. By utilizing a well-known duality between
coherence and distributional robustness~\citep{Delbaen02, FollmerSc02,
  CheriditoDeKu04, RuszczynskiSh06}, we can further show that
$\mathsf{W}_{\qaset}(\model)$ is in fact flexible enough to represent any
worst-case performance over distribution shifts
\begin{equation}
  \label{eqn:dro}
  \model \mapsto \sup_{Q \in \mathcal{Q}} \E_Q[\model],
\end{equation}
where $\mathcal{Q}$ is a collection of probability measures 
dominated by $P$.  The worst-case subpopulation
performance~\eqref{eqn:cvar} is a particular distributionally robust
formulation with $\mathcal{Q}$ given by the set~\eqref{eqn:subpopulations}; our
equivalence results to come show the converse by utilizing the
generalization~\eqref{eqn:gen}.

Formally, let $(\Omega, \mathcal{F}, P)$ be the underlying probability space over
which all random variables are defined. We restrict attention to random
variables with finite moments: for $k \in [1, \infty)$, define
\begin{align*} \mathcal{L}^k \defeq \left\{\model: (\Omega, \mathcal{F}, P) \to \R
~~\mbox{s.t.}~~ \E_P[\model^k] < \infty \right\}.
\end{align*} A risk measure $\risk: \mathcal{L}^k \to \R \cup \{\infty\}$ maps a
random variable connoting a notion of loss, e.g., $\model(\worstcov) =
\E[\loss(\theta(X); Y) \mid \worstcov]$, to a single number representing the
modeler's disutility. We say that a risk measure is \emph{law-invariant} if
$\risk(\model) = \risk(\model')$ whenever $\model \eqd \model'$.  A
\emph{coherent} risk measure incorporates sensible notions of risk-aversion
through the following axioms.
\begin{definition}[{\citet[Definition 6.4]{ShapiroDeRu14}}] 
$\risk: \mathcal{L}^k \to \R \cup \{\infty\} $ is \textbf{coherent} if it satisfies
  \begin{enumerate}
  \item Convexity: $\risk(t\model+(1-t)\model') \le t
\risk(\model)+(1-t)\risk(\model)$ for all $t\in[0,1]$ and $\model,\model' \in
\mathcal{L}^k$
  \item Monotonicity: $\risk(\model)\le \risk(\model')$ if $\model, \model'
\in \mathcal{L}^k$ and $\model \le \model'$ $P$-almost surely
  \item Translation equivariance: $\risk(\model+c)=\risk(\model)+c$ for all
$c\in\R$ and $\model \in \mathcal{L}^k$
  \item Positive homogeneity: $\risk(c \model)= c\risk(\model)$ for all $c>0$
and $\model \in \mathcal{L}^k$
  \end{enumerate}
\end{definition}

While the above axioms have originally been proposed with economic and
financial applications in mind (e.g., see the tutorial~\citet{Rockafellar07}),
they can also be interpreted from the perspective of predictive
systems~\citep{WilliamsonMe19}. Convexity models diminishing marginal utility
(for loss/disutility); the modeler incurs higher marginal disutility when the
underlying prediction model $\theta(X)$ is already poor and incurring high
prediction errors. Monotonicity is a natural property to enforce. Translation
equivariance says that any addition of certain prediction error translates to
a proportional change in the modeler's disutility. Similarly, positive
homogeneity says a tenfold increase in prediction error leads to a
proportional increase in the modeler's distutility.

Standard duality results give a one-to-one correspondence between coherent
risk measures and particular distributionally robust formulations. For any
space $\mathcal{L}^k$, consider its dual space $\mathcal{L}^{k_*}$ where
$1/k + 1/k_* = 1$ and $k_*\in(1,\infty]$. Recall that for a risk measure
$\risk: \mathcal{L}^k \to\R \cup \{\infty\}$, we say $\risk(\cdot)$ is proper if
its domain is nonempty and define its \emph{Fenchel conjugate}
$\risk\opt: \mathcal{L}^{k_*}\to \R \cup \{\infty\}$ as
$\risk\opt(L) \defeq \sup_{\model \in \mathcal{L}^{k}} \left\{\E_{P}[L \model] -
  \risk(\model) \right\}$.  For a (sufficiently regular) convex function
$\risk: \mathcal{L}^k \to \R \cup \{\infty\}$, Fenchel-Moreau duality gives the
biconjugacy relation
\begin{equation}
  \label{eqn:biconjugacy}
  \risk(\model) = \sup_{L \in \mathcal{L}^{k_*}}
  \left\{ \E_P[L \model] - \risk\opt(L) \right\}.
\end{equation}
When $\risk(\cdot)$ is coherent, proper, and lower semi-continuous, its
\emph{dual set} (a.k.a. domain of $\risk\opt$) can be characterized as the
following
\begin{equation}
  \label{eqn:risk-dual}
  \dom \risk\opt =
  \left\{ L \in \mathcal{L}^{k_*} :
    L \ge 0, \E_P[L]=1, \text{ and } \E_P[L \model] \le \risk(\model)
    ~\forall \model\in \mathcal{L}^k
  \right\},
\end{equation}
where $\risk\opt(L) = 0$ whenever $L \in \dom \risk\opt$. The dual set is
weak* closed and is a set of probability density functions, so $\E_P[L\model]$
can be viewed as the expectation $\E_Q[\model]$ in under the probability
measure defined by $\frac{dQ}{dP} = L$. Collecting these observations, the
biconjugacy relation~\eqref{eqn:biconjugacy} gives the following result.
\begin{lemma}[{\citet[Theorem 6.42]{ShapiroDeRu14}}]
  \label{lemma:dro} The set of coherent, proper, lower semi-continuous, and
  law-invariant risk measures is identical to the set of mappings given by
  $\model \mapsto \sup_{Q\in\mathcal{Q}} \E_Q[\model]$ for some nonempty class of
  probability measures $\mathcal{Q}$ over $(\Omega, \mathcal{F}, P)$ satisfying
  $\frac{dQ}{dP} \in \mathcal{L}^{k_*}$ for all $Q \in \mathcal{Q}$.
\end{lemma}
\noindent In this equivalence, we do not require the class $\mathcal{Q}$ to be
convex since the dual set of $\model \mapsto \sup_{Q \in \mathcal{Q}} \E_Q[\model]$ is
the convex hull of $\mathcal{Q}$.

The main result of this section gives an equivalence between generalized
worst-case subpopulation performances~\eqref{eqn:gen} and distributionally
robust losses~\eqref{eqn:dro}, and in turn, coherent risk measures. Recall
that a probability space $(\Omega, \mathcal{F}, P)$ is \emph{nonatomic} if any
$S \in \mathcal{F}$ with $\P(S) > 0$ contains a subset $S' \in \mathcal{F}$ such that
$\P(S) > \P(S') > 0$.
\begin{theorem}
  \label{thm:wpa}
  A generalized worst-case subpopulation performance measure~\eqref{eqn:gen}
  is coherent, proper, lower semi-continuous, and law-invariant. If $P$ is nonatomic,
  then the converse also holds.
\end{theorem}
\noindent Nonatomicity is a mild assumption since most machine learning applications do not
depend on the underlying probability space: we can simply take the underlying
probability space to be the standard uniform space $\Omega = [0, 1]$ equipped
with the Borel sigma algebra and the Lebesgue/uniform measure.

Theorem~\ref{thm:wpa} is essentially a consequence of a well-known reformulation
of coherent risk measures known as the Kusuoka representation~\citep{Kusuoka01,
  Shapiro13k, ShapiroDeRu14}. Our proof is not novel, but we give it in
Appendix~\ref{sec:proof-wpa} for completeness; it is constructive so that
given a coherent risk measure $\risk(\cdot)$, we define the exact set of
probability measures $\qaset_{\risk}$ such that $\risk(\cdot)$ is equal to the
generalized worst-case subpopulation performance~\eqref{eqn:gen} defined with
$\qaset_{\risk}$.  Our construction makes concrete how the axioms of coherence
translate to multiple preferences over subpopulation sizes.

As an example, we consider the \emph{higher-order conditional
  value-at-risk}~\citep{Krokhmal07}, a more conservative risk measure than the
worst-case subpopulation performance~\eqref{eqn:cvar} we considered in prior
sections: for $\alpha \in (0, 1]$,
\begin{equation*}
  \risk_k(\model) \defeq \inf_{\eta\in\R}
  \left\{\frac{1}{\alpha} \left( \E_P \hinge{\model-\eta}^k \right)^{1/k} + \eta
    \right\}.
\end{equation*}
When $k = 1$, we recover the worst-case subpopulation
performance~\eqref{eqn:cvar}. The following result makes explicit the
equivalence relation given in Theorem~\ref{thm:wpa} for the risk measure
family $\risk_k(\cdot)$. 
\begin{proposition}
  \label{prop:hcvar}
  For $k \in (1, \infty)$,
  let $\qaset_{k} \defeq
  \left\{\lambda\in\Delta((0,1]): \int_0^1 \left(\int_u^1 a^{-1} d\lambda(a)\right)^{k_*} du
    \le \alpha^{-k_*} \right\}$. If $P$ is nonatomic, 
  $$ \risk_k(\model)
  = \sup_{Q \ll P} \left\{\E_Q[\model]: \E_P \left(\frac{dQ}{dP} \right)^{k_*} 
    \le \alpha^{-k_*} \right\}
  = \mathsf{W}_{\qaset_{k}}(\model).
  $$
\end{proposition}
\noindent In the above result, we can see that the set $\qaset_k$ allows
arbitrarily small subpopulations by using the $L^{k_*}(P)$-norm.  See
Appendix~\ref{sec:proof-hcvar} for its proof.

\section{Discussion}
\label{section:discussion}

To ensure models perform reliably under operation, we need to
\emph{rigorously} certify their performance under distribution shift prior to
deployment. We study the \emph{worst-case subpopulation performance} of a
model, a natural notion of model robustness that is easy to communicate with
users, regulators, and business leaders.  Our approach allows flexible
modeling of subpopulations over an arbitrary variable $\worstcov$ and
automatically accounts for complex intersectionality. We develop scalable
estimation procedures for the worst-case subpopulation
performance~\eqref{eqn:cvar} and the certificate of
robustness~\eqref{eqn:certificate} of a model. Our convergence guarantees
apply even when we use high-dimensional inputs (e.g. natural language) to
define $\worstcov$. Our diagnostic may further inform data collection and
model improvement by suggesting data collection efforts and model fixes on
regions of $\mc{Z}$ with high  conditional
risk~\eqref{eqn:cond-risk}.

The worst-case performance~\eqref{eqn:cvar} over mixture components as
subpopulations~\eqref{eqn:subpopulations} provides a strong guarantee over
arbitrary subpopulations, but it may be overly conservative in cases when there is
a natural geometry in $\worstcov \in \mc{Z}$. Incorporating such
problem-specific structures in defining a tailored notion of subpopulation is a
promising research direction towards operationalizing the concepts put forth
in this work. As an example, \citet{SrivastavaHaLi20} recently studied similar
notions of worst-case performance defined over human annotations.

{
Our finite sample concentration guarantees are limited in that they cannot
show the benefits of debiasing. Thus, the only theoretical results in this
work that can quantify the benefits of debiasing is our asymptotic result.
This is especially restrictive since debiasing is a technique to remove bias
that arises in \emph{finite samples} due to estimation of nuisance parameters.
Developing advanced statistical learning theory that allow quantification of
this behavior remains an important open problem.
}

We focus on the narrow question of evaluating model robustness under
distribution shift; our evaluation perspective is thus inherently
limited. Data collection systems inherit socioeconomic inequities, and
reinforce existing political power structures. This affects \emph{all} aspects
of the ML development pipeline, and our diagnostic is no panacea.  A notable
limitation of our approach is that we do not explicitly consider the power
differential that often exists between those who deploy the prediction system
and those for whom it gets used on.  Systems must be deployed with considered
analysis of its adverse impacts, and we advocate for a holistic approach
towards addressing its varied implications.

%




\bibliographystyle{abbrvnat}

\ifdefined\useorstyle
\setlength{\bibsep}{.0em}
\else
\setlength{\bibsep}{.7em}
\fi

\bibliography{./bib}

\ifdefined\useorstyle



\newpage
\begin{APPENDICES}

\section{Proof of Theorem~\ref{theorem:clt}}
\label{section:proof-clt}

Let $D \defeq (X, Y, Z)$,  $\cfoldinf$ be the set of indices \emph{not} in $I_k$ (as $n \to \infty$)
 and define $T$ to be the debiased functional
\begin{equation}
  \label{eqn:functional}
  T(P; \model, \tau) \defeq \inf_{\eta} \left\{ \frac{1}{\alpha} \E_P \hinge{\model(\worstcov) - \eta} + \eta\right\}
  + \E_{P} [\tau(\worstcov) (\loss(\theta(X); Y) - \model(\worstcov))].
\end{equation}
The cross-fitted estimator is
$\what{\omega}_{\alpha} = \frac{1}{K} \sum_{\indfold = 1}^{K} T(\what{P}_k;
\hall)$, where $\what{P}_k$ is the empirical distribution on the $k$-th
fold. Our goal is to show
\begin{align*}
  & \sqrt{|I_k|} \left( T\left(\what{P}_k; \hall\right)
  - T(P; \pall) \right) \\
  & = \sqrt{|I_k|} \left( T\left(\what{P}_k; \hall\right) - T\left( P; \hall\right)
      \right) 
    + \sqrt{|I_k|} \left( T\left(P; \hall\right) - T(P; \pall) \right) \\
    & =\frac{1}{\sqrt{|I_k|}} \sum_{i \in I_k} \psi(D_i)
  + o_p(1).
\end{align*}

We begin by establishing
\begin{align}
  \label{eqn:conv-classical}
  \sqrt{|I_k|} \left( T\left(\what{P}_k; \hall\right) - T\left( P; \hall\right)
  \right) = \frac{1}{\sqrt{|I_k|}} \sum_{i \in \fold} \psi(D_i) + o_p(1).
\end{align}

{
We begin by showing that the feasibility region in the dual formulation of the
 can be restricted to a compact set.  Let $S_{\alpha}$ be
an interval around $\aq{\pmu}$
\begin{equation*}
  S_{\alpha} \defeq [\aq{\pmu} \pm 1].
\end{equation*}
\begin{proposition}
  \label{proposition:cvar-domain}
  Under the conditions of Theorem~\ref{theorem:clt},
  \begin{align}
    \label{eqn:cpt-domain-ok}
    \inf_{\eta \in S_{\alpha}} \left\{ \frac{1}{\alpha}
    \E_{\worstcov \sim Q}\hinge{\hmu(\worstcov) - \eta} + \eta \right\}
    & = \inf_{\eta} \left\{ \frac{1}{\alpha}
      \E_{\worstcov \sim Q} \hinge{\hmu(\worstcov) - \eta} + \eta \right\}
      ~~\mbox{eventually} 
  \end{align}
  almost surely for $Q = \what{P}_k, P$.
\end{proposition}
\noindent See Appendix~\ref{section:proof-cvar-domain} for a proof of
Proposition~\ref{proposition:cvar-domain}.

This almost sure equivalence allows us to replace $T$ with its counterpart
where the dual solution set is restricted to a compact region
\begin{align}
  \label{eqn:cpt-functional}
  T_{S_{\alpha}} (Q; \mu, \tau)
  \defeq \inf_{\eta \in S_{\alpha}} \left\{ \frac{1}{\alpha}
  \E_{D \sim Q}\hinge{\mu(\worstcov) - \eta} + \eta
  \right\} + \E_{D \sim Q}[\tau(\worstcov) (\loss(\theta(X); Y) - \model(\worstcov))].
\end{align}
Below, we will use the notation
\begin{equation*}
  \lambda_{\rm opt}(H) \defeq \inf_{\eta \in S_{\alpha}} H(\eta)
\end{equation*}
and rewrite the above functional  as
\begin{align*}
  T_{S_{\alpha}}(\what{P}_k; \hall) & = \lambda_{\rm opt}(\what{P}_k) + \E_{D \sim
                                      \what{P_k}}[\tau(\worstcov) (\loss(\theta(X); Y) - \model(\worstcov))], \\
  T_{S_{\alpha}}(P_n; \hall) & = \lambda_{\rm opt}(P_n)
                               + \E_{D \sim P}[\tau(\worstcov) (\loss(\theta(X); Y) - \model(\worstcov))],
\end{align*}
Proposition~\ref{proposition:cvar-domain} ensures that convergence of
$T_{S_{\alpha}}$ implies the convergence~\eqref{eqn:conv-classical}.  We will
show convergence conditional on the event where $\hmu \in \mathcal{U}$
\begin{align}
  \label{eqn:good-event}
  \event_{n, k} \defeq
  \left\{ \hmu \in \mathcal{U}, \mbox{  and conditions of Assumption~\ref{assumption:neyman} holds for $k$}
  \right\}.
\end{align}
This implies the unconditional result~\eqref{eqn:conv-classical} since
$\P(\event_{n, k}) \to 1$ by
Assumptions~\ref{assumption:neyman},~\ref{assumption:regularity}.

We will use the functional delta method to the map
$Q \mapsto T_{S_{\alpha}}(Q; \hall)$.  A prerequisite for this is to use
standard empirical process theory to show the empirical measure $\what{P}_k$
satisfies the uniform CLT over random variables
\begin{subequations}
  \label{eqn:functions}
  \begin{align*}
    & f_{n, \eta}(D)
     \defeq 
       \frac{1}{\alpha} \hinge{\hmu(\worstcov) - \eta} + \eta 
      ~\mbox{for}~ \eta \in S_{\alpha},\\
    &  \hthr(\worstcov)(\loss(\theta(X); Y) - \hmu(\worstcov)).
  \end{align*}
\end{subequations}
To simplify notation, we  define
\begin{equation*}
f_{n, \aq{\pmu} + 2}(D)
\defeq \hthr(\worstcov)(\loss(\theta(X); Y) - \hmu(\worstcov))
\end{equation*}
so that the stochastic process $f_{n, \eta}$ represents both types of random
variables with $\eta \in \Lambda \defeq S_{\alpha} \cup \{\aq{\pmu} + 2\}$.
Formally, let $\ell^{\infty}(\Lambda)$ be the usual space of uniformly bounded
functions on $\Lambda$ endowed with the sup norm. We treat measures as bounded
functionals $\what{P}_k: \eta \mapsto \E_{D \sim \what{P}_k} f_{n, \eta}(D)$
and $P_n: \eta \mapsto \E_{D \sim P} f_{n, \eta}(D)$.
\begin{proposition}[{\citet[Prop. 4]{JeongNa20}}]
  \label{prop:donsker}
  Conditional on
  $\event_{n,k}$,
  \begin{align*}
    \sqrt{n} \left(\E_{D \sim \what{P}_k} f_{n, \eta}(D)
    - \E_{D \sim P} f_{n, \eta}(D)\right)
    \cd \mathbb{G} ~~\mbox{in}~~\ell^{\infty}\left(\Lambda\right),
  \end{align*}
  where $\mathbb{G}$ is a Gaussian process on
  $\Lambda = S_{\alpha} \cup \{\aq{\pmu} + 2\}$ with covariance
  $\Sigma(\eta, \eta')$
  \begin{align*}
    & \frac{1}{\alpha^2} \E\left[ \hinge{\pmu(\worstcov) - \eta}
      \hinge{\pmu(\worstcov) - \eta'} \right]
      + \frac{\eta'}{\alpha} \E\left[ \hinge{\pmu(\worstcov) - \eta}  \right]
      + \frac{\eta}{\alpha} \E\left[ \hinge{\pmu(\worstcov) - \eta'}  \right]
      ~ \mbox{if}~\eta, \eta' \in S_{\alpha} \\
    & \frac{1}{\alpha^2}
      \E\left[ \hinge{\pmu(\worstcov) - \eta} (\loss(\theta(X); Y) - \pmu(\worstcov)) \right]
      ~\mbox{if}~\eta \in S_{\alpha}, \eta' = \aq{\pmu} + 2 \\
    & \E\left[ \pthr(\worstcov)^2(\loss(\theta(X); Y) - \pmu(\worstcov))^2 \right] ~\mbox{if}~\eta = \eta' = \aq{\pmu} + 2.
  \end{align*}
\end{proposition}
\noindent Without loss of generality, we use the almost surely equivalent
version of the Gaussian process $\mathbb{G}$ that have continuous sample
paths.


To apply the functional delta method, it remains to show that the functional
of interest is appropriately smooth. While classical results~\cite[Theorem
6.5.3]{ShapiroDeRu21} give Gateaux differentiability, we actually need a
stronger notion of uniform Hadamard differentiability.  
First, we review notation for the functional delta method. Let
$\lambda: \mathbb{D}_{\lambda} \subset \mathbb{D} \to \R$ be a functional on a
metrizable topological vector space $\mathbb{D}$ and denote its (arbitrary)
subset by $\mathbb{D}_{\lambda}$. We use $r_n$ to denote a sequence of
constants $r_n \to \infty$, and treat $P_n, P: \eta \mapsto \E_P f_{n, \eta}$
as elements of $\mathbb{D}_{\lambda} \subset \mathbb{D}$ such that
$P_n \to P$.
\begin{lemma}[{\citep[Delta method, Theorem 3.9.5]{VanDerVaartWe96}}]
  \label{lemma:delta}
  Let $\mathbb{D}_0 \subseteq \mathbb{D}$ and let $\Omega_n$ be sample spaces
  defined for each $n$.  For every converging sequence $H_n \in \mathbb{D}$
  such that $P_n + r_n^{-1} H_n \in \mathbb{D}_{\lambda}$ ~for all $n$, and
  $H_n \to H \in \mathbb{D}_0 \subset \mathbb{D}$, let there be a map
  $d\lambda_P(\cdot)$ on $\mathbb{D}_0$ such that
  \begin{align*}
    r_n(\lambda(P_n + r_n^{-1} H_n) - \lambda(P_n)) \to d\lambda_P(H).
  \end{align*}
  Let $\xi_n: \Omega_n \to \mathbb{D}_{\lambda}$ be maps with
  $\sqrt{n} (\xi_n - P_n) \cd \xi$ in $\mathbb{D}$, where $\xi$ is separable
  and takes values in $\mathbb{D}_0$. If $d\lambda_P(\cdot)$ can be extended
  to the whole of $\mathbb{D}$ as a linear, continuous map, then
  \begin{align*}
    r_n(\lambda(\xi_n) - \lambda(P_n)) - d\lambda_P(r_n(\xi_n - P)) \cp 0.
  \end{align*}
\end{lemma}

We want to apply this canonical delta method to the functional
\begin{equation*}
\lambda = T_{S_{\alpha}} ~~\mbox{with}~~ \mathbb{D} = \ell^{\infty}(\Lambda),
r_n = \sqrt{|\fold|},
\xi_n = \what{P}_k: \eta \mapsto \E_{\what{P}_k} f_{n, \eta}.
\end{equation*}
In what follows, remember that the domain of interest
$\mathbb{D}_{\lambda_{\rm opt}}$ is defined by the functions
\begin{align*}
  \eta \mapsto
  \begin{cases}
    \frac{1}{\alpha} \E_{Q}\hinge{\mu(\worstcov) - \eta} + \eta
    & ~~\mbox{if}~\eta \in S_{\alpha} \\
    \E_{Q}[\tau(\worstcov) (\loss(\theta(X); Y) - \model(\worstcov))] 
    & ~~\mbox{if}~\eta = \aq{\pmu} + 2
  \end{cases}
\end{align*}
such that $Q$ is a probability on $\mc{D}$, $\E[\mu^2(X)] < \infty$,
$e(\cdot) \in [c, 1-c]$, $|h| \le M_h$, and $P$ is an element of this set with
$\mu = \pmu$. 

The following lemma confirms the hypothesis of Lemma~\ref{lemma:delta}---it is
essentially known (Danskin's theorem) but we give a full proof in
Appendix~\ref{section:proof-danskin} for completeness. Recall that the Gaussian
process $\mathbb{G}$ has continuous sample paths lying in
$\mathbb{D}_0 \defeq \left\{ H \in \ell^{\infty}(\Lambda): \eta \mapsto
  H(\eta)~\mbox{is continuous} \right\}$.
\begin{lemma}
  \label{lemma:danskin}
  Assume that the hypothesis of Theorem~\ref{theorem:clt} holds.
  On the event $\event_{n, k}$,
  $\lambda_{\rm opt}: \mathbb{D}_{\lambda_{\rm opt}} \subset
  \ell^{\infty}(\Lambda) \to \R$ satisfies the following: for every converging
  sequence $H_n \in \ell^{\infty}(\Lambda)$ s.t.
  $P_n + |I_k|^{-1/2} H_n \in \mathbb{D}_{\lambda_{\rm opt}}$ for all $n$, and
  $H_n \to H \in \mathbb{D}_0 \defeq \left\{ H \in \ell^{\infty}(\Lambda):
    \eta \mapsto H(\eta)~\mbox{is continuous} \right\}$,
  \begin{align}
    \sqrt{|I_k|} (\lambda_{\rm opt}(P_n + |I_k|^{-1/2} H_n) - \lambda_{\rm opt}(P_n))
    \to H(\aq{\pmu}) \eqdef d\lambda_{{\rm opt},P}(H).
    \label{eqn:cpt-conv-classical}
  \end{align}
\end{lemma}
\noindent Conclude that conditional on $\event_{n, k}$, the
convergence~\eqref{eqn:cpt-conv-classical} holds. As argued above, this shows
our final claim~\eqref{eqn:conv-classical}.

}

Finally, we show the term
$\sqrt{|I_k|} \left( T\left(P; \hall\right) - T(P; \pall) \right)$ vanishes.
Let $\rem: [0, 1] \to \R$
\begin{align}
  \label{eqn:remainder}
  \rem(r) \defeq
  T\left(P; (1-r) (\pall) + r (\hall)\right) - T(P; \pall),
\end{align}
so that $\rem(0) = 0$, and $\rem(1)$ is equal to
$T\left(P; \hall\right) - T(P; \pall)$. On the event $\event_{n, k}$,  $\rem(r)$ is differentiable under
Assumptions~\ref{assumption:residuals},~\ref{assumption:neyman}
\begin{align}
  \rem'(r) =  \E_{\worstcov \sim P}
  \left[ (\hmu - \pmu)(\worstcov)
  \left(\frac{1}{\alpha} \indic{\rmu(\worstcov) \ge \aq{\rmu}}
  - \pthr(\worstcov) \right) \right]                                                                    \end{align}
where $(\rall) \defeq (\pall) + r((\hall) - (\pall))$.
 (It is easy to
check this using Danskin's theorem.)

The mean value theorem then gives
$\rem(1) = \rem(0) + \rem'(r) \cdot (1-0) = \rem'(r)$ for some $r \in [0,
1]$. Debiasing nominally guarantees $\rem'(0) = 0$, but going further we will now show
$\sup_{r \in [0, 1]} |\rem'(r)| = o_p(n^{-1/2})$.
Since $\hthr \in [-M, M]$ on $\event_{n, k}$, elementary calculations and
repeated applications of Holder's inequality yield
\begin{align}
  \sup_{r \in [0, 1]} |\rem'(r)| & \le \Linf{\hmu - \pmu}   \sup_{r \in [0, 1]} \Lone{\frac{1}{\alpha} \indic{\rmu(\worstcov) \ge \aq{\rmu}} - \pthr(\worstcov)} 
               \label{eqn:deriv-bound}
\end{align}
where $C$ is a positive constant that only depends on $c$, and $M$. The last
term in the bound is bounded by $\delta_n n^{-1/2}$ by the definition of
$\event_{n, k}$.

Our uniform differentiability assumption~\eqref{eqn:unif-diff} for $F_{\rmu}$
guarantees the following notions of smoothness.  We omit its derivations as
the calculations are elementary but tedious~\citep{JeongNa20}.
\begin{lemma}
  \label{lemma:thr-conv}
  On the event $\event_{n, k}$, we have $\sup_{r \in [0, 1]}
    \left| \aq{\rmu} - \aq{\pmu} \right|
    = O\left(\Linf{\hmu - \pmu}\right) =  O(\delta_n n^{-1/3})$ and
\begin{align*}
  & \sup_{r \in [0, 1]} \Lone{\frac{1}{\alpha} \indic{\rmu(\worstcov) \ge \aq{\rmu}} - \pthr(\worstcov)} \\
  & \lesssim n^{1/6} \left( \Lone{\hmu - \pmu}  + | \what{q}_k - \aq{\hmu}| 
    +\sup_{r \in [0, 1]} \left| \aq{\rmu} - \aq{\pmu} \right| \right) \\
  & \qquad + n^{-1/6} \delta_n + \Linf{\hmu - \pmu}.
\end{align*}
\end{lemma}
\noindent We conclude that the bound~\eqref{eqn:deriv-bound} is
$O(\delta_n n^{-1/2})$ on the event $\event_{n, k}$, meaning
$\sup_{r \in [0, 1]} |\rem'(r)| = o(n^{-1/2})$ on the event $\event_{n, k}$.
Since $\P(\event_{n, k}) \to 1$ from
Assumptions~\ref{assumption:neyman},~\ref{assumption:regularity}, we have the final result.

{
\subsection{Proof of Proposition~\ref{proposition:cvar-domain}}
\label{section:proof-cvar-domain}

We show that the optima
\begin{align*}
  \argmin_{\eta} \left\{ \frac{1}{\alpha}
    \E_{\worstcov \sim \what{P}_k}\hinge{\hmu(\worstcov) - \eta} + \eta \right\},
   ~~~~\argmin_{\eta} \left\{ \frac{1}{\alpha}
  \E_{\worstcov \sim P}\hinge{\hmu(\worstcov) - \eta} + \eta \right\}
\end{align*}
converge to their population limit.  First, we use following elementary result
to characterize the limiting quantity.
\begin{lemma}[~\citet{RockafellarUr00}]
  \label{lemma:cvar-soln}
  If a random variable $\xi$ has a positive density at the
  $(1-\alpha)$-quantile $\aq{\xi} \defeq \inf\{t: F_{\xi}(t) \ge 1-\alpha\}$,
  then
  \begin{equation*}
    \argmin_{\eta} \left\{ \frac{1}{\alpha} \E\hinge{\xi - \eta} + \eta
      \right\} = \left\{ P^{-1}_{1-\alpha}(\xi)\right\}.
  \end{equation*}
\end{lemma}
\noindent Applying Lemma~\ref{lemma:cvar-soln} to $\xi = \pmu(\worstcov)$, we have
\begin{align*}
 \{ \aq{\pmu} \} = \argmin_{\eta} \left\{ \frac{1}{\alpha}
  \E \hinge{\pmu(\worstcov) - \eta} + \eta \right\}.
\end{align*}

To show convergence, 
define
$g(\eta) \defeq \frac{1}{\alpha} \E\hinge{\pmu(\worstcov) - \eta} + \eta$ and
\begin{align*}
  \what{g}_{1, n, k}(\eta) \defeq
  \frac{1}{\alpha}
  \E_{\worstcov \sim \what{P}_k}\hinge{\hmu(\worstcov) - \eta} + \eta,~~~
    \what{g}_{2, n, k}(\eta) \defeq
  \frac{1}{\alpha}
  \E_{\worstcov \sim P}\hinge{\hmu(\worstcov) - \eta} + \eta.
\end{align*}
Our argument relies on epi-convergence theory.
\begin{lemma}[{\citet[Theorems 7.17, 7.31]{RockafellarWe98}}]
  \label{lemma:epi}
  Let $g_n, g: \R \to \R$ be proper, closed, convex, and coercive functions,
  and let $\argmin_{\eta} g(\eta) = \{\eta\opt\}$ be unique. If $g_n \to g$
  pointwise, then
  $\sup_{\eta \in \argmin_{\eta'} g_n(\eta')} |\eta - \eta\opt| \to 0$.
\end{lemma}

To verify the hypothesis of Lemma~\ref{lemma:epi}, note
$\what{g}_{1, n, k}, \what{g}_{2, n, k}, g$ are all proper, continuous,
convex, and coercive, and $g$ has a unique optimum from
Lemma~\ref{lemma:cvar-soln}. Assumption~\ref{assumption:neyman} implies that
$\what{g}_{2, n, k} \to g$ pointwise.  To show $\what{g}_{1, n, k} \cas g$
pointwise, note that
\begin{align}
  |\what{g}_{1, n, k}(\eta) - g(\eta)|
  \le & \frac{1}{\alpha} \left| \E_{\worstcov \sim \what{P}_k} \hinge{\hmu(\worstcov) - \eta}
  - \E_{\worstcov \sim P} \hinge{\hmu(\worstcov) - \eta} \right| \nonumber \\
  & + \frac{1}{\alpha} \left| \E_{\worstcov \sim P} \hinge{\hmu(\worstcov) - \eta}
  - \E_{\worstcov \sim P} \hinge{\pmu(\worstcov) - \eta} \right|.   \label{eqn:g-conv}
\end{align}
Assumption~\ref{assumption:neyman} implies the second term vanishes pointwise.
The first term vanishes due to SLLN for triangular arrays.
\begin{lemma}[{\citet[Theorem 2]{HuMoTa89}}]
  \label{lemma:triangular-slln}
  Let $\{\xi_{ni}\}_{i=1}^n$ be a triangular array where
  $\worstcov_{n1}, \worstcov_{n2}, \ldots$ are independent random variables for any fixed
  $n$. If there exists a real-valued random variable $\xi$ such that
  $|\xi_{ni}|\le\xi$ and $\E[\xi^2] <\infty$, then
  $\frac{1}{n} \sum_{i=1}^n (\xi_{ni} - \E[\xi_{ni}]) \cas 0$.
\end{lemma}
\noindent If we condition on $\{D_i\}_{i \in \cfoldinf}$, we can apply
Lemma~\ref{lemma:triangular-slln} since each element in
$\{\hinge{\hmu(\worstcov_i) - \eta}\}_{i \in \fold}$ is mutually independent.
For any $\eta \in \R$, the first term in the bound~\eqref{eqn:g-conv} thus
vanishes a.s. conditional on $\{D_i\}_{i \in \cfoldinf}$.  By dominated
convergence, it follows that this term vanishes a.s. unconditionally.

\subsection{Proof of Lemma~\ref{lemma:danskin}}
\label{section:proof-danskin}
The following proof is due to~\citet{Romisch05}.  We use
$\eta \mapsto Q(\eta)$ to refer to members of $\mathbb{D}_{\lambda}$ and let
$S(F, \epsilon)$ be $\epsilon$-approximate minima of $F$
\begin{align*}
  S(F, \epsilon)
  =\left\{ \eta: F(\eta) \le \inf_{\eta \in S_{\alpha}}  F(\eta) + \epsilon \right\}.
\end{align*}
($S(P, 0) = \{\aq{\pmu}\}$ by Lemma~\ref{lemma:cvar-soln}.)

Let us first show the upper bound
$\limsup_{n \to \infty} \sqrt{|I_k|} \left(\lambda_{\rm opt}(P_n +
  |I_k|^{-1/2} H_n) - \lambda_{\rm opt}(P_n)\right) \le d\lambda_{{\rm opt},
  P}(H)$. Notice that
\begin{align*}
  \sqrt{|I_k|} \left(
  \lambda_{\rm opt}(P_n + |I_k|^{-1/2} H_n) - \lambda_{\rm opt}(P_n) \right)
  & \le |I_k|^{1/2} \left( (P_n + |I_k|^{-1/2} H_n)(\eta_n) - P_n(\eta_n)
    + |I_k|^{-1}\right) \\
  & \le H(\eta_n) + \norm{H_n - H} + |I_k|^{-1/2}
\end{align*}
where $\eta_n \in S(P_n, |I_k|^{-1})$.  Since
$\Linf{\hmu - \pmu} \le \delta_n$ on the event $\event_{n, k}$,
$\eta_n \in S(P, |I_k|^{-1} + \alpha^{-1} \delta_n)$.  Then,
$\lim \eta_n = \aq{\pmu}$ since for any convergent subsequence $\eta_{n_m}$,
its limit must be contained in the singleton $S(P, 0)$:
Lipschitzness of $\eta \mapsto P(\eta)$ implies
$\eta\opt \in S(P, (\alpha^{-1}+1) |\eta_{n_m} - \eta\opt| + |I_k|^{-1} +
\alpha^{-1} \delta_{n_m})$, we further implies
$\lim_{n \to \infty} H(\eta_n) = H(\aq{\pmu}) = d\lambda_{{\rm opt}, P}(H)$ by
continuity of $H \in \mathbb{D}_0$.

Now, we proceed to the lower bound
$\liminf_{n \to \infty} \sqrt{|I_k|} \left(\lambda_{\rm opt}(P_n +
  |I_k|^{-1/2} H_n) - \lambda_{\rm opt}(P_n)\right) \ge d\lambda_{{\rm opt},
  P}(H)$. Begin by noting that
\begin{align*}
  & \lambda_{\rm opt}(P_n + |I_k|^{-1/2} H_n) - \lambda_{\rm opt}(P_n) \\
  & \ge (P_n + |I_k|^{-1/2} H_n)(\eta_n) - |I_k|^{-1}
    - P_n(\eta_n) \\
  & \ge |I_k|^{-1/2} H(\eta_n) + |I_k|^{-1/2}\norm{H_n - H} - |I_k|^{-1}
\end{align*}
for $\eta_n \in S(P_n + |I_k|^{-1/2} H_n, |I_k|^{-1})$.
By elementary algebra, we have
\begin{align*}
  S(P_n + |I_k|^{-1/2} H_n, |I_k|^{-1})
  & \subseteq S(P_n, |I_k|^{-1/2}  \norm{H_n} + |I_k|^{-1}) \\
  & \subseteq S(P, |I_k|^{-1/2} \norm{H_n} + |I_k|^{-1} +
  \alpha^{-1}\delta_n)
\end{align*}
on the event $\event_{n, k}$ so we can again conclude
$\lim \eta_n = \aq{\pmu}$ and continuity of $H$ gives the desired inequality.


}


\section{Proof of finite-sample concentration results}
\label{section:proof-finite-sample}

Our results are based on a general concentration guarantee for estimating the
dual reformulation~\eqref{eqn:dual} for any given $\model(\worstcov)$.  We
give this result in Appendix~\ref{section:proof-cvar-concentration}, and build
on it in subsequent proofs of key results. In the following, we use $\lesssim$
to denote inequality up to a numerical constant that may change line by line.

\subsection{Concentration bounds for worst-case subpopulation performance}
\label{section:proof-cvar-concentration}

Since $\loss(\what{y}; y) \ge 0$ for losses used in most machine
learning problems, we assume that $\modelclass$ consists of nonnegative
functions. To show exponential concentration guarantees, we consider
sub-Gaussian conditional risk models $\model(\worstcov)$. Note the concentration 
results here are more general than needed for the purpose of proving the main results, 
because any random variable bounded in $[0,\lbound]$ is inherently sub-Gaussian with 
paramater $\lbound^2/4$.

\begin{definition}
  \label{assumption:tails}
  A function $\model: \mathcal{Z} \to \R$ with $\E|\model(\worstcov)|<\infty$ is sub-Gaussian with parameter $\sigma^2$ if
  \begin{align*}
    \E\left[\exp\left( \lambda (\model(\worstcov) - \E[\model(\worstcov)]) \right)\right]
    \le \exp\left( \frac{\sigma^2\lambda^2}{2}\right)
    ~~\mbox{for all}~\lambda \in \R.
  \end{align*}
\end{definition}
\noindent The sub-Gaussian assumption can be relaxed to sub-exponential
random variables, with minor and standard modifications to subsequent
results. We omit these results for brevity.

Define a dual plug-in estimator for the worst-case subpopulation performance of $\model(\worstcov)$ on $I_k$
\begin{equation}
  \label{eqn:emp-cvar}
  \hworstsub(\model) = \inf_{\eta}
  \left\{
    \frac{1}{\alpha | I_k|} \sum_{i \in I_k} \hinge{\model(Z_i) - \eta} + \eta
  \right\}.
\end{equation}
The following result shows that for any sub-Gaussian $\model$ that is bounded
from below, the plug-in estimator~\eqref{eqn:second} converges at the rate
$O_p(|I_k|^{-1/2})$.

\begin{proposition}
  \label{prop:cvar-concentration}
  There is a universal constant $C>0$ such that for all $h\ge0$ that is sub-Gaussian 
  with parameter $\sigma^2$, 
  \begin{equation*}
    |\hworstsub(\model) - \worstsub(\model)|
    \le \frac{C \sigma}{\alpha} \sqrt{\frac{\log(2/\delta)}{|I_k|}} 
    \text{ with probability at least } 1-\delta.
  \end{equation*}
\end{proposition}
\noindent We prove the proposition in the rest of the subsection. By a judicious
application of the empirical process theory, our bounds---which apply to 
nonnegative random variables---are simpler than existing
concentration guarantees for conditional value-at-risk~\citep{Brown07,
  PrashanthJaKo19}.

Our starting point is the following claim, which bounds
$|\hworstsub(\model) - \worstsub(\model)|$ in terms of the suprema of
empirical process on $\{z \mapsto \hinge{\model(z) - \eta}: \eta \ge 0\}$.
\begin{claim}
  \label{claim:restricted-eta}
  \begin{align}
    \label{eqn:restricted-eta}
    \left|\hworstsub(\model) - \worstsub(\model)\right|
    \le \frac{1}{\alpha} \sup_{\eta \ge 0}
    \left| \frac{1}{|I_k|} \sum_{i \in I_k} \hinge{\model(\worstcov_i) - \eta}
    - \E\hinge{\model(\worstcov) - \eta} \right|
  \end{align}
\end{claim}
\noindent The crux of this claim is that $\eta$ does not range over $\R$, but
rather has a lower bound; the value $0$ can be replaced with any almost sure
lower bound on $\model(\worstcov)$. Deferring the proof of
Claim~\ref{claim:restricted-eta} to the end of the subsection, we proceed by
bounding the suprema of the empirical process in the preceding display.

We begin by introducing requisite concepts in empirical process theory, which
we use in the rest of the proof; we refer readers to~\citet{VanDerVaartWe96}
for a comprehensive treatment.  Recall the definition of Orlicz norms, which
allows controlling the tail behavior of random variables.
\begin{definition}[Orlicz norms]
  \label{def:orlicz}
  Let $\psi: \R \to \R$ be a non-decreasing, convex function with
  $\psi(0) = 0$.  For any random variable $W$, its Orlicz norm
  $\norm{W}_{\psi}$ is
  \begin{align*}
    \norm{W}_{\psi} \defeq \inf\left\{t > 0: \E\left[ \psi\left(\frac{|W|}{t}\right)\right] \le 1
    \right\}.
  \end{align*}
\end{definition}
\begin{remark}
  From Markov's inequality, we have
  \begin{equation*}
    \P(|W| > t) \le \P\left( \psi\left(\frac{|W|}{\norm{W}_{\psi}}\right)
      \ge \psi\left(\frac{t}{\norm{W}_{\psi}} \right) \right)
    \le \psi\left(\frac{t}{\norm{W}_{\psi}} \right)^{-1}.
  \end{equation*}
  For $\psi_p(s) = e^{s^p} - 1$, a similar argument yields
  \begin{align}
    \label{eqn:orlicz-tail-ineq}
    \P( |W| > t)  \le 2 \exp\left( -t^{p} /\norm{W}_{\psi_p}^p \right).
  \end{align}
\end{remark}

A sub-Gaussian random variable $\model(\worstcov)$ with parameter $\sigma^2$
has bounded Orlicz norm $\norm{\model(\worstcov)}_{\psi_2} \le 2 \sigma$ (see,
for example,~\citet[Section 2.4]{Wainwright19} and~\citet[Lemma
2.2.1]{VanDerVaartWe96}).

\begin{remark}
  The converse also holds: for $W$ such that
  $\P(|W| > t) ]\le c_1 \exp(-c_2 t^p)$ for all $t$, and constants
  $c_1, c_2 > 0$ and $p \ge 1$, Fubini gives
  \begin{align*}
    \E \exp\left( \frac{|W|^p}{t^p} \right) - 1 
    = \E\left[ \int_0^{|W|^p} t^{-1/p} \exp(t^{-1/p} s) ds \right]
    = \int_0^\infty \P(|W|^p > s) t^{-1/p} \exp\left(t^{-1/p} s\right) ds.
  \end{align*}
  Using the tail probability bound, the preceding display is bounded by
  \begin{equation*}
    c_1 \int_0^\infty  \exp(-c_2 s)t^{-1/p} \exp(t^{-1/p} s) ds
    = \frac{c_1t^{-1/p}}{c_2 - t^{-1/p}}.
  \end{equation*}
  So the Orlicz norm $\norm{W}_{\psi_p}$ is bounded by
  $\left( \frac{1+c_1}{c_2} \right)^{1/p}$.
\end{remark}

In the following, we let $W$ be the right hand side of the
bound~\eqref{eqn:restricted-eta}, and control its Orlicz norm
$\norm{W}_{\psi_2}$ using Dudley's entropy integral~\citep{VanDerVaartWe96}.
We use the standard notion of the covering number. For a vector space
$\mathcal{V}$, let $V \subset \mathcal{V}$ be a collection of vectors. Letting
$\norm{\cdot}$ be a norm on $\mathcal{V}$, a collection
$\{v_1, \ldots, v_\covnum\} \subset \mathcal{V}$ is an \emph{$\epsilon$-cover} of
$\mathcal{V}$ if for each $v \in \mathcal{V}$, there is a $v_i$ satisfying
$\norm{v - v_i} \le \epsilon$. The \emph{covering number} of $V$ with respect
to $\norm{\cdot}$ is
\begin{equation*}
  \covnum(\epsilon, V, \norm{\cdot}) \defeq
  \inf\left\{\covnum \in \N : ~ \mbox{there~is~an~}
    \epsilon \mbox{-cover~of~} V
    ~ \mbox{with~respect~to~} \norm{\cdot} \right\}.
\end{equation*}
For a collection $\fclass$ of functions $f : \mathcal{\worstcov} \to \R$, let $F$
be its envelope function such that $|f(z)| \le F(z)$ for all
$ z \in \mathcal{\worstcov}$. The following result controls the suprema of
empirical processes using the (uniform) metric entropy. The result is based on
involved chaining arguments~\cite[Section 2.14]{VanDerVaartWe96}.
\begin{lemma}[{\citet[Theorem 2.14.1 and 2.14.5]{VanDerVaartWe96}}]
  \label{lemma:dudley}
  \begin{align*}
    & \sqrt{|I_k|} \norm{\sup_{f \in \fclass} \left| \frac{1}{|I_k|} \sum_{i \in I_k} f(\worstcov_i)
    - \E f(\worstcov)\right|}_{\psi_2} \\
    & \lesssim \norm{F}_{\psi_2} + \norm{F}_{L^2(P)} \sup_{Q} \int_0^1
      \sqrt{1 + \log \covnum(\epsilon \norm{F}_{L^2(Q)}, \fclass, L^2(Q))} d\epsilon,
  \end{align*}
  where the supremum is over all discrete probability measures $Q$ such that
  $\norm{F}_{L^2(Q)} > 0$.
\end{lemma}

Evidently, $F(z) = \hinge{\model(z)} = \model(z)$ is an envelope
function for the following class of functions
\begin{align*}
  \fclass = \{ z \mapsto \hinge{\model(z) - \eta}: \eta \ge 0\}.
\end{align*}
Using the tail probability bound~\eqref{eqn:orlicz-tail-ineq}, we conclude
\begin{align*}
  & \sup_{\eta \ge 0}
  \left| \frac{1}{|I_k|} \sum_{i \in I_k} \hinge{\model(\worstcov_i) - \eta}
  - \E\hinge{\model(\worstcov) - \eta} \right| \\
  & \lesssim \sqrt{\frac{\log\left(2/\delta\right)}{|I_k|}}
  \left( \norm{F}_{\psi_2} + \norm{F}_2
   \sup_{Q} \int_0^1
      \sqrt{1 + \log \covnum(\epsilon \norm{F}_{L^2(Q)}, \fclass, L^2(Q))} d\epsilon
  \right),
\end{align*}
with probability at least $1-\delta$.

Since we have $\norm{F}_{L^2(P)} \le \norm{F}_{\psi_2} \lesssim \sigma$, it
now suffices to show that the above uniform metric entropy is bounded by a universal
constant.  We use the standard notion of VC-dimension~\cite[Chapter 2.6, page
135]{VanDerVaartWe96}.
\begin{lemma}[{\citet[Theorem 2.6.7]{VanDerVaartWe96}}]
  \label{lemma:vc-bounds-cov}
  Let $\vcdim(\fclass)$ be the VC-dimension of the collection of subsets
  $\{(z, t): t< f(x)\}$ for $f \in \fclass$. For any probability measure $Q$
  such that $\norm{F}_{L^2(Q)} > 0$ and $0 < \epsilon  < 1$, we have
  \begin{align*}
    \covnum(\epsilon \norm{F}_{L^2(Q)}, \fclass, L^2(Q))
    \lesssim \vcdim(\fclass) (16e)^{\vcdim(\fclass)}
    \left( \frac{1}{\epsilon}\right)^{2(\vcdim(\fclass) - 1)}.
  \end{align*}
\end{lemma}
\noindent Translations of a monotone function on $\R$ has VC-dimension $2$.
\begin{lemma}[{\citet[Theorem 2.6.16]{VanDerVaartWe96}}]
  \label{lemma:vc-monotone}
  The class of functions $\fclass' = \{ z \mapsto \hinge{\model(z) - \eta}: \eta \in \R\}$
  has VC-dimension $\vcdim(\fclass') = 2$.
\end{lemma}
\noindent From Lemmas~\ref{lemma:vc-bounds-cov} and~\ref{lemma:vc-monotone},
we conclude that for the function class
$\fclass = \{ z \mapsto \hinge{\model(z) - \eta}: \eta \ge 0\}$, the uniform
metric entropy
\begin{align*}
  \sup_{Q} \int_0^1
  \sqrt{1 + \log \covnum(\epsilon \norm{F}_{L^2(Q)}, \fclass, L^2(Q))} d\epsilon
\end{align*}
is bounded by a universal constant. This gives our desired result.

\paragraph{Proof of Claim~\ref{claim:restricted-eta}}

To show the bound~\eqref{eqn:restricted-eta}, we use the dual reformulation
for both $\worstsub(\model)$
and its empirical approximation $\hworstsub(\model)$ on $I_k$.
For any probability measure $P$, recall two different definitions of the
quantile of $\model(\worstcov)$
\begin{align*}
  P^{-1}_{1-\alpha}(\model(\worstcov)) & \defeq \inf \{t: \P_{\worstcov}(\model(\worstcov) \le t)
  \ge 1-\alpha\} \\
  P^{-1}_{1-\alpha, +}(\model(\worstcov)) & \defeq \inf \{t: \P_{\worstcov}(\model(\worstcov) \le t)
  > 1-\alpha\}.
\end{align*}
We call $P^{-1}_{1-\alpha, +}(\model(\worstcov))$ the upper
$(1-\alpha)$-quantile. The two values characterize the optimal solution set
of the dual problem~\eqref{eqn:dual}; they are identical when
$\model(\worstcov)$ has a positive density at $\aq{\model(\worstcov)}$. 
\begin{lemma}[{\citet[Theorem 10]{RockafellarUr02}}]
  \label{lemma:cvar-dual-optima}
  For any probability measure $P$ such that $\model(\worstcov) \ge 0$
  $P$-a.s. and $\E_P[\model(\worstcov)_+] < \infty$, we have
  \begin{equation*}
    [  P^{-1}_{1-\alpha}(\model(\worstcov)),   P^{-1}_{1-\alpha, +}(\model(\worstcov))] =  \argmin_{\eta \in \R} \left\{
    \frac{1}{\alpha} \E_P\hinge{\model(\worstcov) - \eta} + \eta
      \right\}.
  \end{equation*}
\end{lemma}

Since $P$ was an arbitrary measure in Lemmas~\ref{lemma:dual}
and~\ref{lemma:cvar-dual-optima}, identical results follow for the empirical
distribution on $I_k$. Hence, we have
\begin{align*}
   \left|\hworstsub(\model) - \worstsub(\model)\right| 
  & = \left| \inf_{\eta \in \R} \left\{
    \frac{1}{\alpha |I_k|} \sum_{i \in I_k} \hinge{\model(\worstcov_i) - \eta} + \eta
    \right\}
    - \inf_{\eta \in \R} \left\{
    \frac{1}{\alpha} \E\hinge{\model(\worstcov) - \eta} + \eta
    \right\} \right| \\
    & = \left| \inf_{\eta \ge 0} \left\{
    \frac{1}{\alpha |I_k|} \sum_{i \in I_k} \hinge{\model(\worstcov_i) - \eta} + \eta
    \right\}
    - \inf_{\eta \ge 0} \left\{
    \frac{1}{\alpha} \E\hinge{\model(\worstcov) - \eta} + \eta
      \right\} \right|
\end{align*}
where we used Lemma~\ref{lemma:cvar-dual-optima} to restrict the feasible
region in the last equality. The preceding display is then bounded by
\begin{align*}
   \sup_{\eta \ge 0} \left| 
    \frac{1}{\alpha |I_k|} \sum_{i \in I_k} \hinge{\model(\worstcov_i) - \eta} + \eta
    - 
    \frac{1}{\alpha} \E\hinge{\model(\worstcov) - \eta} - \eta \right|.
\end{align*}

\subsection{Proof of Theorem~\ref{theorem:uniform}}
\label{section:proof-uniform}

We abuse notation and use $C$ for a numerical constant that may change line to
line. From the decomposition~\eqref{eqn:decomposition}, it suffices to bound
term $(a)$ and term $(b)$ separately. 

Term $(b)$ can be bounded with the help of Proposition~\ref{prop:cvar-concentration} 
because $\what{\model}_\indfold(\cdot)$ is trained on a sample $I_k^c$
independent from $I_k$ used to estimate the worst-case subpopulation
performance (Eq.~\eqref{eqn:emp-cvar}).  More precisely, recalling 
that any bounded random variable random variable taking values in $[0, \lbound]$ is
sub-Gaussian with parameter $\lbound^2/4$, Proposition~\ref{prop:cvar-concentration} implies
\begin{align*}
  \left| \hworstsub(\what{\model}_\indfold) - \worstsub(\what{\model}_\indfold)
  \right| \le \frac{C \lbound}{\alpha} \sqrt{\frac{\log(2/\delta)}{|I_k|}}~~\mbox{with probability at least}~1-\delta.
\end{align*}
For the debiasing term, Hoeffding's inequality implies with probability 
at least $1-\delta$ conditional on $\cfold$,
\begin{align*}
  & \left| \E_{\what{P}_k} [\hthr(Z)(\loss(\theta(X);Y) - \what\model_k(Z)) \mid \cfold]
  - \E_P [\hthr(Z)(\loss(\theta(X);Y) - \what\model_k(Z)) \mid \cfold] \right| \\
  & \lesssim \frac{\lbound}{\alpha} \sqrt{\frac{\log(2/\delta)}{|I_k|}}, 
\end{align*}
because $\hthr\in\{0,1/\alpha\}$ and $\loss(\cdot;\cdot), \model(\cdot)\in[0,\lbound]$. 
Then the same bound holds with total probability at least $1-\delta$. Hence, with probability 
at least $1-2\delta$, term $(b)$ is bounded by 
\[
  (b) \equiv T(P;\what\model_k,\hthr) - T(\what{P}_k;\what\model_k,\hthr) 
  \lesssim \frac{\lbound}{\alpha} \sqrt{\frac{\log(2/\delta)}{|I_k|}}.
\]

To bound term $(a)$ in the decomposition~\eqref{eqn:decomposition}, we first note
\begin{align*}
   \left|\worstsub(\what{\model}_\indfold) - \worstsub(\condrisk)\right|
  &  \le \frac{1}{\alpha} \sup_{\eta} \left| \E\left[ \hinge{\what{\model}_\indfold(\worstcov) - \eta} 
  \mid I_k^c \right] - \E\hinge{\condrisk(\worstcov) - \eta} \right| \\
  &  \le \frac{1}{\alpha}  \E \left[ \left|\what{\model}_\indfold(\worstcov) - \condrisk(\worstcov)\right| \mid I_k^c \right],
\end{align*}
where the first inequality follows from the dual~\eqref{eqn:dual}, and the second inequality 
follows from the non-expansiveness of the function $\hinge{\cdot}$.  Similarly for the debiasing term,
\begin{align*} 
\left| \E [\hthr(Z)(\loss(\theta(X);Y) - \what\model_k(Z)) \mid \cfold]\right| 
& = \left| \E [ \E [\hthr(Z)(\loss(\theta(X);Y) - \what\model_k(Z)) \mid Z, \cfold] \mid \cfold]\right| \\
& = \left| \E [\hthr(Z)(\model\opt(Z) - \what\model_k(Z)) \mid \cfold]\right| \\
& \le \frac{1}{\alpha}  \E \left[ \left|\what{\model}_\indfold(\worstcov) - \condrisk(\worstcov)\right| \mid I_k^c \right],
\end{align*}
where the first equality follows from the law of total probability, the second by definition of $\condrisk$, 
and the inequality because $\hthr(\cdot)\in\{0, 1/\alpha\}$. Hence, 
\begin{align*}
  |(a)| &\le |\worstsub(\condrisk) - \worstsub(\what{\model}_\indfold)| + |\E [\hthr(Z)(\loss(\theta(X);Y) - \what\model_k(Z)) \mid \cfold]| \\
  &\le \frac{2}{\alpha}  \E \left[ \left|\what{\model}_\indfold(\worstcov) - \condrisk(\worstcov)\right| \mid I_k^c \right], \\
  &  \le \frac{2}{\alpha} \sqrt{\E \left[ \left(\what{\model}_\indfold(\worstcov) - \condrisk(\worstcov)\right)^2\mid I_k^c \right] }
  = \frac{2}{\alpha} \sqrt{\error(\modelclass, I_k^c)},
\end{align*}
where the first inequality follows from the definition of $(a)$, the second inequality 
follows from the bounds above, the last inequality uses Holder inequality, and we define 
the generalization error for the first-stage estimation problem~\eqref{eqn:first-pop} 
based on $I_k^c$, 
\begin{align*}
  \error(\modelclass, I_k^c) 
  \defeq & \E\left[\left(\condrisk(\worstcov) - \what{\model}_\indfold(\worstcov)\right)^2 \mid I_k^c\right]\\
  = & \E \left[ (\loss(\theta(X); Y) - \what{\model}_\indfold(\worstcov))^2 \mid I_k^c \right]
    - \E (\loss(\theta(X); Y) - \condrisk(\worstcov))^2 \\
  = & \E \left[ (\loss(\theta(X); Y) - \what{\model}_\indfold(\worstcov))^2 \mid I_k^c \right]
    - \E (\loss(\theta(X); Y) - \bestmodel(\worstcov))^2 
    + \E \left(\condrisk(\worstcov) - \bestmodel(\worstcov)\right)^2 
\end{align*}

We use the following concentration result based on the localized Rademacher
complexity~\citep{BartlettBoMe05}.
\begin{lemma}[{\citet[Corollary 5.3]{BartlettBoMe05}}]
  \label{lemma:loc-rad}
  Let Assumption~\ref{assumption:bdd} hold.
  Then, with probability at least $1-\delta$,
  \begin{align*}
    \E \left[ (\loss(\theta(X); Y) - \what{\model}_\indfold(\worstcov))^2 \mid I_k^c \right]
    - \E (\loss(\theta(X); Y) - \bestmodel(\worstcov))^2 
    \le
      C \lbound^2 \left(r_{|I_k^c|}\opt
      +  \frac{\log(1/\delta)}{|I_k^c|}\right).
  \end{align*}
\end{lemma}
\noindent Using $\sqrt{a + b + c} \le \sqrt{a} + \sqrt{b} +\sqrt{c}$ 
for $a, b, c \ge 0$, we have the desired result.

\subsection{Proof of Theorem~\ref{theorem:dim-free}}
\label{section:proof-dim-free}
Instead of the decomposition~\eqref{eqn:decomposition} we use for
Theorem~\ref{theorem:uniform}, we use an alterantive form
\begin{align}
  \label{eqn:decomposition-dim-free}
  \what{\omega}_{\alpha,k} - \worstsub(\condrisk)
  &= \underbrace{\hworstsub(\what{\model}_\indfold) - \hworstsub(\condrisk)}_{(a): ~\mbox{\scriptsize first stage}}
  + \underbrace{\hworstsub(\condrisk) - \worstsub(\condrisk)}_{(b): ~\mbox{\scriptsize second stage}} \nonumber \\
  &\quad\quad + \underbrace{\E_{\what{P}_k}[\hthr(Z)(\loss(\theta(X);Y)-\what\model_k(Z)) 
  \mid \cfold]}_{(c): ~\mbox{\scriptsize debiasing term}}
\end{align}
Term $(b)$ can be bounded using Proposition~\ref{prop:cvar-concentration}
as before. Without assuming $\condrisk \in \modelclass$, recall that any bounded random variable taking
values in $[0, \lbound]$ is sub-Gaussian with parameter $\lbound^2/4$, so Proposition~\ref{prop:cvar-concentration} yields
\begin{align*}
  |(b)| = \left| \hworstsub(\condrisk) - \worstsub(\condrisk)\right|
  \le C \frac{\lbound}{\alpha} \sqrt{\frac{\log(2/\delta)}{|I_k|}}~~\mbox{with probability at least}~1-\delta.
\end{align*}


We first notice the following bound on term $(a)$.
\begin{align}
  (a) = \left|\hworstsub(\what{\model}_\indfold) - \hworstsub(\condrisk)\right|
 &  \le \frac{1}{\alpha} \sup_{\eta}   \left|
 \E_{\what{P}_k} \left[ \hinge{\what{\model}_\indfold(\worstcov_i) - \eta}
   - \hinge{\condrisk(\worstcov_i) - \eta} \mid \cfold \right] \right|
 \nonumber \\
 &  \le  
   \frac{1}{\alpha} \E_{\what{P}_k}\left[\left| \what{\model}_\indfold(\worstcov) 
   - \condrisk(\worstcov)\right|\mid \cfold\right]
   \equiv \frac{1}{\alpha} \norm{\what{\model}_\indfold(\worstcov) 
   - \condrisk(\worstcov)}_{L^1(\what{P}_k|\cfold)}.
   \label{eqn:emp-two-norm}
\end{align}

Next we will bound the debiasing term $(c)$ with the same quantity. We start by observing Hoeffding's 
inequality implies 
\begin{align*}
  & \left| \E_{\what{P}_k}[\hthr(Z)(\loss(\theta(X);Y)-\what\model_k(Z)) \mid \cfold]
  - \E [\hthr(Z)(\loss(\theta(X);Y)-\what\model_k(Z)) \mid \cfold] \right| \\
  & \le \frac{\lbound}{\alpha} \sqrt{\frac{\log(2/\delta)}{|I_k|}}
\end{align*}
with probability at least $1-\delta$ because $\hthr\in\{0,1/\alpha\}$ and $\loss,\what\model_k \in 
[0,\lbound]$ almost surely. Then notice by definition of $\condrisk \equiv \E[\loss(\theta(X);Y) 
\mid Z]$ we know $\E [\hthr(Z)(\loss(\theta(X);Y)-\what\model_k(Z)) \mid \cfold] = \E [\hthr(Z)
(\condrisk(Z)-\what\model_k(Z)) \mid \cfold]$ by conditioning on $Z$. Then we again invoke Hoeffding's 
inequality to argue with probability at least $1-\delta$,
\begin{align*}
  \left| \E [\hthr(Z) (\condrisk(Z)-\what\model_k(Z)) \mid \cfold]
  - \E_{\what{P}_k} [\hthr(Z) (\condrisk(Z)-\what\model_k(Z)) \mid \cfold] \right|
  \le \frac{\lbound}{\alpha} \sqrt{\frac{\log(2/\delta)}{|I_k|}}.
\end{align*}
Lastly we notice 
\begin{align*}
  \left|\E_{\what{P}_k} [\hthr(Z) (\condrisk(Z)-\what\model_k(Z)) \mid \cfold]\right| 
  & \le \frac{1}{\alpha} \E_{\what{P}_k} [|\condrisk(Z)-\what\model_k(Z)| \mid \cfold] \\
  & \equiv \frac{1}{\alpha} \norm{\condrisk(Z) - \what\model_k(Z)}_{L^1(\what{P}_k|\cfold)}
\end{align*}
because $\hthr\in\{0,1/\alpha\}$. Hence, we conclude with probability at least $1-2\delta$,
\[ 
  |(c)| \le \frac{1}{\alpha} \norm{\condrisk(Z) - \what\model_k(Z)}_{L^1(\what{P}_k|\cfold)} + \frac{2\lbound}{\alpha} 
  \sqrt{\frac{\log(2/\delta)}{|I_k|}}.
\]
Thus we have shown with probability at least $1-3\delta$,
\[
  |\what{\omega}_{\alpha,k} - \worstsub\opt| \le \frac{2}{\alpha}\norm{\condrisk(Z) - 
  \what\model_k(Z)}_{L^1(\what{P}_k|\cfold)} + \frac{C\lbound}{\alpha} 
  \sqrt{\frac{\log(2/\delta)}{|I_k|}}.
\]
We now present two approaches to bounding the empirical $L^1$-norm of of the error $\condrisk(Z) - \what\model_k(Z)$ 
for whether the model class $\modelclass$ is convex. Before we move on, notice the following identity that is useful 
for both cases, which can be interpreted geometrically as the cosine theorem in the $L^2(\what{P}_k)$ space. For any 
two functions $\model, \tilde\model: \mathcal{Z} \to [0,\lbound]$, 
\begin{equation}
\label{eqn:cosine-identity}
  \norm{\tilde{\model}(\worstcov) - \model(\worstcov)}_{L^2(\what{P}_k)} 
  = \Delta_{I_k}(\tilde{\model}) -  \Delta_{I_k}(\model)
  + 2\E_{\what{P}_k} \left[(\loss(\theta(X);Y) - \model(Z))
     \left(\tilde{\model}(\worstcov) - \model(\worstcov)\right)\right].
\end{equation}

\subsubsection{Continuing proof of Theorem~\ref{theorem:dim-free} with a non-convex model class}
First note H\"older's inequality implies \[\norm{\condrisk(Z) - 
\what\model_k(Z)}_{L^1(\what{P}_k|\cfold)} \le \norm{\condrisk(Z) - 
\what\model_k(Z)}_{L^2(\what{P}_k|\cfold)},\] so it suffices to bound the $L^2$-norm. In order to 
use the identity~\eqref{eqn:cosine-identity} with $\model = \condrisk$ and $\tilde\model = \what{\model}_k$, 
we notice by definition of $\condrisk \equiv \E[\loss(\theta(X);Y) \mid Z]$ that
$\E[(\loss(\theta(X);Y) - \condrisk(Z)) (\what{\model}_k(\worstcov) - \model(\worstcov))] = 0$ 
for all $\tilde\model$. Since $(\loss(\theta(X);Y) - \condrisk(Z)) (\what{\model}_k(\worstcov) 
- \condrisk(\worstcov))$ is almost surely bounded in $[-\lbound^2,\lbound^2]$ and i.i.d., Hoeffding
inequality~\cite[Ch. 2]{Wainwright19} yields
\begin{align*}
  & \left| \E_{\what{P}_k} [ (\loss(\theta(X);Y) - \condrisk(Z)) (\what{\model}_k(\worstcov)
  - \condrisk(\worstcov)) ]\right| \\
  & \le \lbound^2 \sqrt{\frac{2\log(2/\delta)}{|I_k|}}
  ~~\mbox{with probability at least}~1-\delta.
\end{align*}
Hence, the identity~\eqref{eqn:cosine-identity} implies with probability at least $1-\delta$, 
\begin{align*}
  \norm{\condrisk(Z) - \what\model_k(Z)}_{L^2(\what{P}_k|\cfold)}^2 \le 
  \left[ \Delta_{I_k}(\what{\model}_\indfold) -  \Delta_{I_k}(\bestmodel) \right]
    + \left[ \Delta_{I_k}(\bestmodel) -  \Delta_{I_k}(\condrisk) \right]
    + \lbound^2 \sqrt{\frac{2\log(2/\delta)}{|I_k|}}
\end{align*}

Similarly, Hoeffding inequality implies with probability at least $1-\delta$,
\begin{align*}
  \Delta_{I_k}(\bestmodel) - \Delta_{I_k}(\condrisk)
  & \le \E(\loss(\theta(X); Y) - \bestmodel(\worstcov))^2 - \E (\loss(\theta(X); Y) - \condrisk(\worstcov))^2
    + \lbound^2\sqrt{\frac{2\log(1/\delta)}{|I_k|}} \\
  & = \norm{\bestmodel - \condrisk}_{L^2}^2 + \lbound^2\sqrt{\frac{2\log(1/\delta)}{|I_k|}},
\end{align*}
where the equality follows by the definition of the conditional risk $\condrisk(\worstcov) \equiv 
\E[\loss(\theta(X);Y)\mid Z]$.
Hence, with probability at least $1-2\delta$, 
\begin{align*}
  \norm{\condrisk(Z) - \what\model_k(Z)}_{L^2(\what{P}_k|\cfold)}
  &\le \sqrt{
    \Delta_{I_k}(\what{\model}_\indfold) - \Delta_{I_k}(\bestmodel) 
    + \norm{\bestmodel - \condrisk}_{L^2}^2 
    + 2\lbound^2 \sqrt{\frac{2\log(2/\delta)}{|I_k|}} }\\
  &\le \sqrt{[\Delta_{I_k}(\what{\model}_\indfold) - \Delta_{I_k}(\bestmodel)]_+}
    + \norm{\bestmodel - \condrisk}_{L^2} 
    + \sqrt{2}\lbound \left(\frac{2\log(2/\delta)}{|I_k|}\right)^{1/4}.
\end{align*}

Therefore, we conclude that with probability at least $1-5\delta$, \[
  \left|\worstsub\opt - \what{\omega}_{\alpha,k}\right| \le \frac{2}{\alpha} 
  \left(\sqrt{[\Delta_{I_k}(\what{\model}_\indfold) - \Delta_{I_k}(\bestmodel)]_+}
    + \norm{\bestmodel - \condrisk}_{L^2} 
    + C\lbound \left(\frac{2\log(2/\delta)}{|I_k|}\right)^{1/4}\right).
\]

\subsubsection{Continuing proof of Theorem~\ref{theorem:dim-free} with a convex model class} 

First notice with probability at least $1-\delta$, 
\begin{align*}
  \norm{\condrisk(Z) - \what\model_k(Z)}_{L^1(\what{P}_k|\cfold)} 
  & \le \norm{\bestmodel(Z) - \what\model_k(Z)}_{L^1(\what{P}_k|\cfold)} + 
  \norm{\condrisk(Z) - \bestmodel(Z)}_{L^1(\what{P}_k)} \\
  & \le \norm{\bestmodel(Z) - \what\model_k(Z)}_{L^2(\what{P}_k|\cfold)} + 
  \norm{\condrisk(Z) - \bestmodel(Z)}_{L^1(\what{P}_k)} \\
  & \le \norm{\bestmodel(Z) - \what\model_k(Z)}_{L^2(\what{P}_k|\cfold)} + 
  \norm{\condrisk(Z) - \bestmodel(Z)}_{L^1(P)} + \frac{\lbound}{2} \sqrt{\frac{\log(1/\delta)}{|I_k|}},
\end{align*}
where the first inequality follows by the triangle inequality, the second by
H\"older's inequality, and the last by Hoeffding inequality because
$\condrisk,\what\model_k\in[0,\lbound]$. The
identity~\eqref{eqn:cosine-identity} implies
\begin{align*}
  & \norm{\bestmodel(Z) - \what\model_k(Z)}_{L^2(\what{P}_k|\cfold)}^2 \\
    & =
  \Delta_{I_k}(\what{\model}_\indfold) - \Delta_{I_k}(\bestmodel) +
  2\E_{\what{P}_k}[(\loss(\theta(X_i);Y_i) - \bestmodel(\worstcov_i))
  (\what{\model}_\indfold(\worstcov_i) - \bestmodel(\worstcov_i))].
\end{align*}

Since we assume the model class $\modelclass$ is convex and $\what{\model}_\indfold\in\modelclass$,
 the first-order condition of $\bestmodel \in \arg\min_{\model\in\modelclass} 
 \E(\loss(\theta(X);Y) - \model(\worstcov))^2$ gives
$$\E[(\loss(\theta(X);Y) - \bestmodel(\worstcov)) (\what{\model}_\indfold(\worstcov) - \bestmodel(\worstcov)) \mid \worstcov, I_k^c] \le 0,$$
so Hoeffding inequality implies with probability at least $1-\delta$,
\begin{align}
  \E_{\what{P}_k}[(\loss(\theta(X);Y) - \bestmodel(\worstcov))
  (\what{\model}_\indfold(\worstcov) - \bestmodel(\worstcov))]
  \le \lbound^2 \sqrt{\frac{2\log(1/\delta)}{|I_k|}}.
\end{align}

Hence, with probability at least $1-2\delta$, 
\begin{align*}
  & \norm{\condrisk(Z) - \what\model_k(Z)}_{L^1(\what{P}_k|\cfold)} \\
  &\le \norm{\bestmodel - \condrisk}_{L^1} + \sqrt{
    \Delta_{I_k}(\what{\model}_\indfold) - \Delta_{I_k}(\bestmodel) 
    + 2\lbound^2 \sqrt{\frac{2\log(1/\delta)}{|I_k|}} }
    + \frac{\lbound}{2} \sqrt{\frac{\log(1/\delta)}{|I_k|}} \\
  &\le \sqrt{[\Delta_{I_k}(\what{\model}_\indfold) - \Delta_{I_k}(\bestmodel)]_+}
    + \norm{\bestmodel - \condrisk}_{L^1} 
    + C \lbound \left(\frac{2\log(2/\delta)}{|I_k|}\right)^{1/4}.
\end{align*}

Therefore, we conclude that with probability at least $1-5\delta$, \[
  \left|\worstsub\opt - \what{\omega}_{\alpha,k}\right| \le \frac{2}{\alpha} 
  \left(\sqrt{[\Delta_{I_k}(\what{\model}_\indfold) - \Delta_{I_k}(\bestmodel)]_+}
    + \norm{\bestmodel - \condrisk}_{L^1} 
    + C\lbound \left(\frac{2\log(2/\delta)}{|I_k|}\right)^{1/4}\right).
\]


\section{Additional experiment details}
\label{section:experiments-details}

In this section, we present additional experiments for the Functional Map of
the World (FMoW) dataset. Due to the ever-changing nature of aerial images and
the uneven availability of data from different regions, it is imperative that
ML models maintain good performance under temporal (learn from the past and
generalize to future) and spatial distribution shifts (learn from one region
and generalize to another). Without having access to the out-of-distribution
samples, our diagnostic raises awareness on brittleness of model performance
against subpopulation shifts.

\subsection{Dataset Description}

The original Functional Map of the World (FMoW) dataset
by~\citep{ChristieFeWiMu18} consists of over 1 million images from over 200
countries. We use a variant, FMoW-WILDS, proposed by~\citet{KohSaEtAl20},
which temporally groups observations to simulate distribution shift across
time. Each data point includes an RGB satellite image $x$, and a corresponding
label $y$ on the land / building use of the image (there are 62 different
classes). FMoW-WILDS splits data into non-overlapping time periods: we train
and validate models $\theta(\cdot)$ on data collected from years 2002-2013,
and simulate distribution shift by looking at data collected during
2013-2018. Data collected during 2002-2013 (``in-distribution'') is split into
training ($n = $76,863), validation ($n = $19,915), and test
($n=$11,327). Data collected during 2013-2018 (``out-of-distribution'') is
split into two sets: one consisting of observations from years 2013-2016
($n = $19,915), and another consisting of observations from years 2016-2018
($n=$22,108). All data splits contain images from a diverse array of
geographic regions. We evaluate the worst-case subpopulation performance on
in-distribution validation data, and study model performance under
distribution shift on data after 2016.

\subsection{Models Evaluated}

We consider \emph{DenseNet} models as reported by~\citet{KohSaEtAl20},
including the vanilla empirical risk minimization (ERM) model and models
trained with robustness interventions (IRM~\citep{ArjovskyBoGuLo20} method;
\citet{KohSaEtAl20} notes that ERM's performance closely match or outperform
``robust'' counterparts even under distribution shift. We also evaluate
\emph{ImageNet} pre-trained \emph{DPN}-68 model from~\citet{MillerTaRaSaKoShLiCaSc21}. As separate experiments, we also consider \emph{ResNet}-18 and 
\emph{VGG}-11 from~\citet{MillerTaRaSaKoShLiCaSc21}, and the results are reported in~\ref{section:addexp}.  

CLIP (Contrastive Language-Image Pre-training) is a newly proposed model
pre-trained on 400M image-text pairs, and has been shown to exhibit strong
zero-shot performance on out-of-distribution
samples~\citep{RadfordKiEtAl21}. Although not specifically designed
for classification tasks, CLIP can be used for classification by predicting
the class whose encoded text is the closest to the encoded image. We consider the weight-space ensembled \emph{CLIP WiSE} models proposed in ~\citep{WortsmanIlLiKiHaFaNaSc21} as it is observed that these models exhibit robust behavior on FMoW. \emph{CLIP WiSE} models are constructed by linearly combining the model weights of 
\emph{CLIP ViT-B16 Zeroshot model} and \emph{CLIP ViT-B16 FMoW end-to-end finetuned} model. 

To illustrate the usage of our method, we choose the \emph{CLIP WiSE} model
that has similar ID validation accuracy as the \emph{DenseNet} Models. This
turns out to be putting $60\%$ weight on \emph{CLIP ViT-B16 Zeroshot model}
and $40\%$ weight on \emph{CLIP ViT-B16 FMoW end-to-end
  finetuned}. \emph{DenseNet} Models have average ID validation loss
$2.4 - 2.8$, but \emph{CLIP WiSE} has average ID validation loss $1.6$. To
ensure fair comparison, we calibrate the temperature parameter such that the
average loss of \emph{CLIP WiSE} matches the worst average loss of the models
considered. We deliberately make \emph{CLIP WiSE} no better than any
\emph{DenseNet} Models, in the hope that out metric will recover its
robustness property.

\begin{table}[t]
\centering
\begin{tabular}{c|c}

\# & \textbf{Text Prompt} \\
\hline
1 & ``CLASSNAME"\\
2 & ``a picture of a CLASSNAME."\\
3 & ``a photo of a CLASSNAME."\\
4 & ``an image of an CLASSNAME"\\
5 & ``an image of a CLASSNAME in asia."\\
6 & ``an image of a CLASSNAME in africa."\\
7 & ``an image of a CLASSNAME in the americas."\\
8 & ``an image of a CLASSNAME in europe."\\
9 & ``an image of a CLASSNAME in oceania."\\
10 & ``satellite photo of a CLASSNAME" \\
11 & ``satellite photo of an CLASSNAME" \\ 
12 & ``satellite photo of a CLASSNAME in asia."\\
13 & ``satellite photo of a CLASSNAME in africa."\\
14 & ``satellite photo of a CLASSNAME in the americas."\\
15 & ``satellite photo of a CLASSNAME in europe."\\
16 & ``satellite photo of a CLASSNAME in oceania."\\
17 & ``an image of a CLASSNAME"\\

\end{tabular}
\caption{Text prompts for CLIP text encoders}
\label{table:prompt}
\end{table}

\subsection{Flexibility of our metric}
\label{section:modelflexibility}
We implement Algorithm~\ref{alg:two-stage} by partitioning the ID validation
data into two; we estimate $\model\opt(\worstcov)$ using XGBoost on one
sample, and estimate $\worstsub(\cdot)$ at varying subpopulation size $\alpha$
on the other. By switching the role of each split, our final estimator
averages two versions of $\hworstsub(\what{\model})$.

\subsubsection{A less conservative Z}
In Section~\ref{section:experiments}, we report results when $\worstcov$ is
defined over all metadata consisting of (longitude, latitude, cloud cover,
region, year), as well as the label $Y$. Defining subpopulations over such a
wide range of variables may be overly conservative in some scenarios, and to
illustrate the flexibility of our approach, we now showcase a more tailored
definition of subpopulations. Since FMoW-WILDS is specifically designed for
spatiotemporal shifts, a natural choice of $\worstcov$ is to condition on
(region, year). Motivated by our observation that some classes are harder
to predict than others (Figure~\ref{fig:fmow}(b)), we also consider
$\worstcov$= (region, year, label $Y$).  We plot our findings in
Figure~\ref{fig:year_region}. If we simply define $\worstcov$= (year,
region), the corresponding worst-case subpopulation performance is less
pessimistic. However, when we add labels to $Z$, we again see a
drastic decrease in the worst-case subpopulation performance, and that 
\textit{CLIP WiSE-FT} outperforms all other models by a significant
amount. This is consistent with our motivation in defining subpopulations over
labels; our procedure automatically takes into account the interplay between
class labels and spatiotemporal information.

\begin{figure}[htbp]
  \centering
       \vspace{-10pt}
     \begin{subfigure}[b]{0.48\textwidth}
         \centering
         \includegraphics[width=\textwidth]{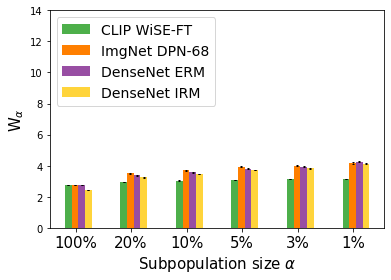}
     \end{subfigure}
     \begin{subfigure}[b]{0.48\textwidth}
         \centering
         \includegraphics[width=\textwidth]{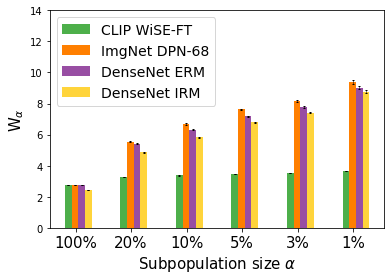}
        \end{subfigure}
     \caption{In the left panel $\worstcov$ = (year, region); in the right panel $\worstcov$ = (year, region, label $Y$). Here we take $\worstcov$ to contain only spatial and temporal information, a less conservative counterpart to the experiment reported in the main text. We again see that introduction of labels in $\worstcov$ drastically increase our metric, showing varying difficulty in learning different labels. }
      \label{fig:year_region}
\end{figure}

\subsubsection{Using semantics of the labels}
Alternatively, we may wish to define subpopulations over rich natural language
descriptions on the input $X$. To illustrate the flexibility of our procedure
in such scenarios, we consider subpopulations defined over the semantic
meaning of the class names: CLIP-encoded class names using the 17 prompts
reported in Table~\ref{table:prompt}. For comparison, we report the
(estimated) worst-case subpopulation performance~\eqref{eqn:cvar} when we take
$\worstcov$ = (all metadata, encoded labels) and (all metadata, label Y, encoded labels) in
Figure~\ref{fig:semantics}. {
Additional experiments using other combinations of features for 
Z—including latitude, longitude, cloud cover, region, and year—are shown in Figures~\ref{fig:fmow_1}–\ref{fig:fmow_3}}. We observe that in this case, the semantics of the
class names do not contribute to further deterioration in robustness, and the
relative ordering across models remains unchanged.

\begin{figure}[htbp]
  \centering
       \vspace{-10pt}
     \begin{subfigure}[b]{0.48\textwidth}
         \centering
         \includegraphics[width=\textwidth]{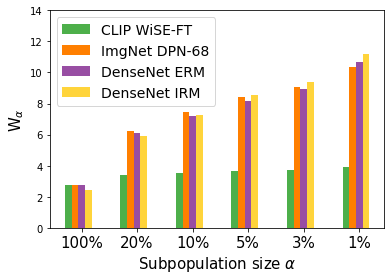}
     \end{subfigure}
     \begin{subfigure}[b]{0.48\textwidth}
         \centering
         \includegraphics[width=\textwidth]{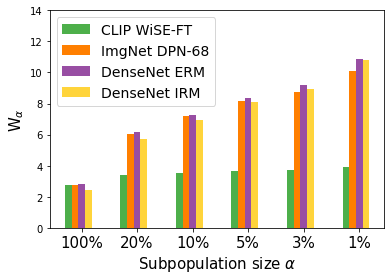}
        \end{subfigure}
     \caption{In the left panel, $\worstcov$ = (all meta, encoded labels); in the right panel, $\worstcov$ = (all meta, label Y, encoded labels). We see that in this case no significant difference is introduced when semantics of the class names are included.}
      \label{fig:semantics}
\end{figure}

\begin{figure}[h!]
\centering
\begin{minipage}[b]{0.49\textwidth}
\centering \includegraphics[width=\textwidth, height=5.5cm]{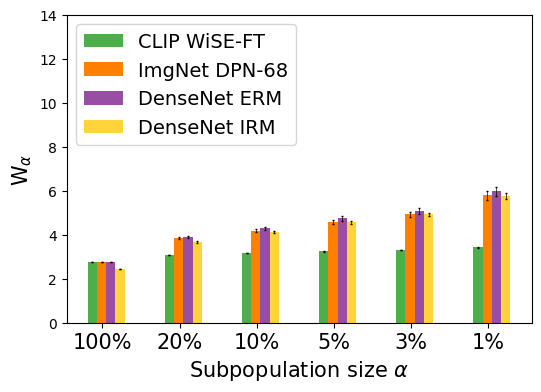}
   \centering{$Z=\{\text{Lat, Lon, Cloud Cover}\}$}
\end{minipage}
\hfill
\begin{minipage}[b]{0.49\textwidth}
\centering \includegraphics[width=\textwidth, height=5.5cm]{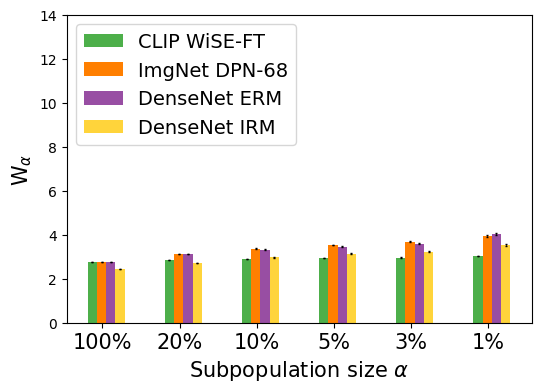}
  \centering{$Z=\{\text{Region, Cloud Cover}\}$}
\end{minipage}
\caption{Worst-case subpopulation performance $W_\alpha(\theta)$ under different Z's and $\alpha's$.}
\label{fig:fmow_1}
\end{figure}

\begin{figure}[h!]
\centering
\begin{minipage}[b]{0.49\textwidth}
\centering \includegraphics[width=\textwidth, height=5.5cm]{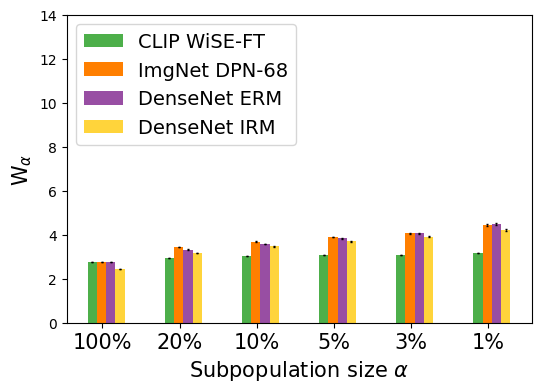}
  \centering{$Z=\{\text{Year, Cloud Cover}\}$}
\end{minipage}
\hfill
\begin{minipage}[b]{0.49\textwidth}
\centering \includegraphics[width=\textwidth, height=5.5cm]{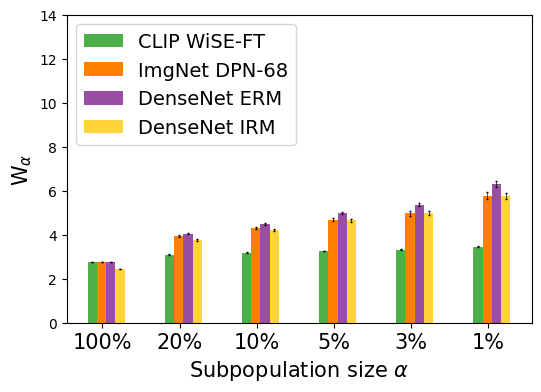}
  \centering{$Z=\{\text{Year, Region, Lat, Lon}\}$}
\end{minipage}
\caption{Worst-case subpopulation performance $W_\alpha(\theta)$ under different Z's and $\alpha's$.}
\label{fig:fmow_2}
\end{figure}

\begin{figure}[h!]
\centering
\begin{minipage}[b]{0.49\textwidth}
\centering \includegraphics[width=\textwidth, height=5.5cm]{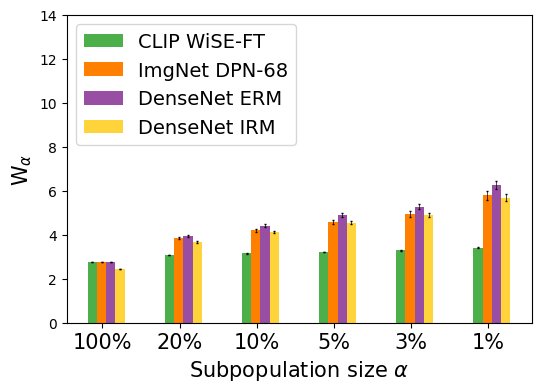}
  \centering{$Z=\{\text{Region, Lat, Lon, Cloud Cover}\}$}
\end{minipage}
\hfill
\begin{minipage}[b]{0.49\textwidth}
\centering \includegraphics[width=\textwidth, height=5.5cm]{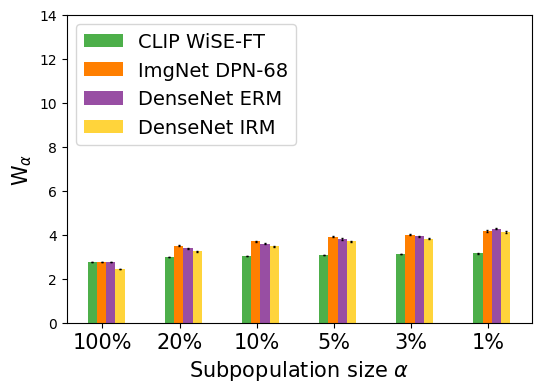}
  \centering{$Z=\{\text{Year, Region}\}$}
\end{minipage}
\caption{Worst-case subpopulation performance $W_\alpha(\theta)$ under different Z's and $\alpha's$.}
\label{fig:fmow_3}
\end{figure}

\subsection{Analysis of spatiotemporal distribution shift}
\label{section:africa}

The significant performance drop in the Africa region on data collected from
2016-2018 was also observed in~\citep{KohSaEtAl20,
  WortsmanIlLiKiHaFaNaSc21}. In
Figures~\ref{fig:africa-train}-\ref{fig:africa-test}, we plot the number of
samples collected from Africa over data splits.  In particular, we observe a
large number of single-unit and multi-unit residential instances emerge in the
OOD data.  Data collection systems are often biased against the African
continent---often as a result of remnants of colonialism---and addressing such
bias is an important topic of future research.

\begin{figure}
\includegraphics[height=18cm]{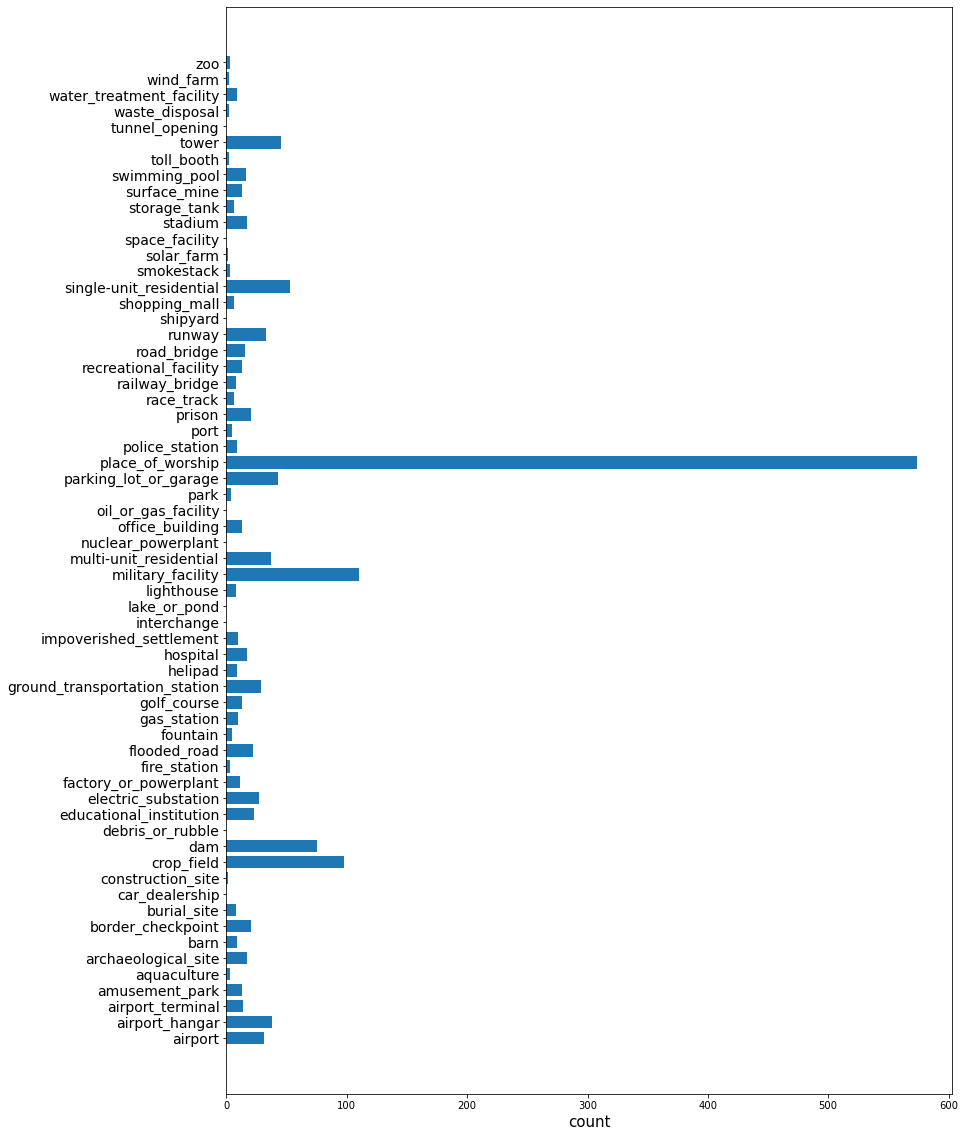}
\caption{Instances by class, ID 2002-2013, Africa}
\label{fig:africa-train}
\end{figure}

\begin{figure}
\includegraphics[height=18cm]{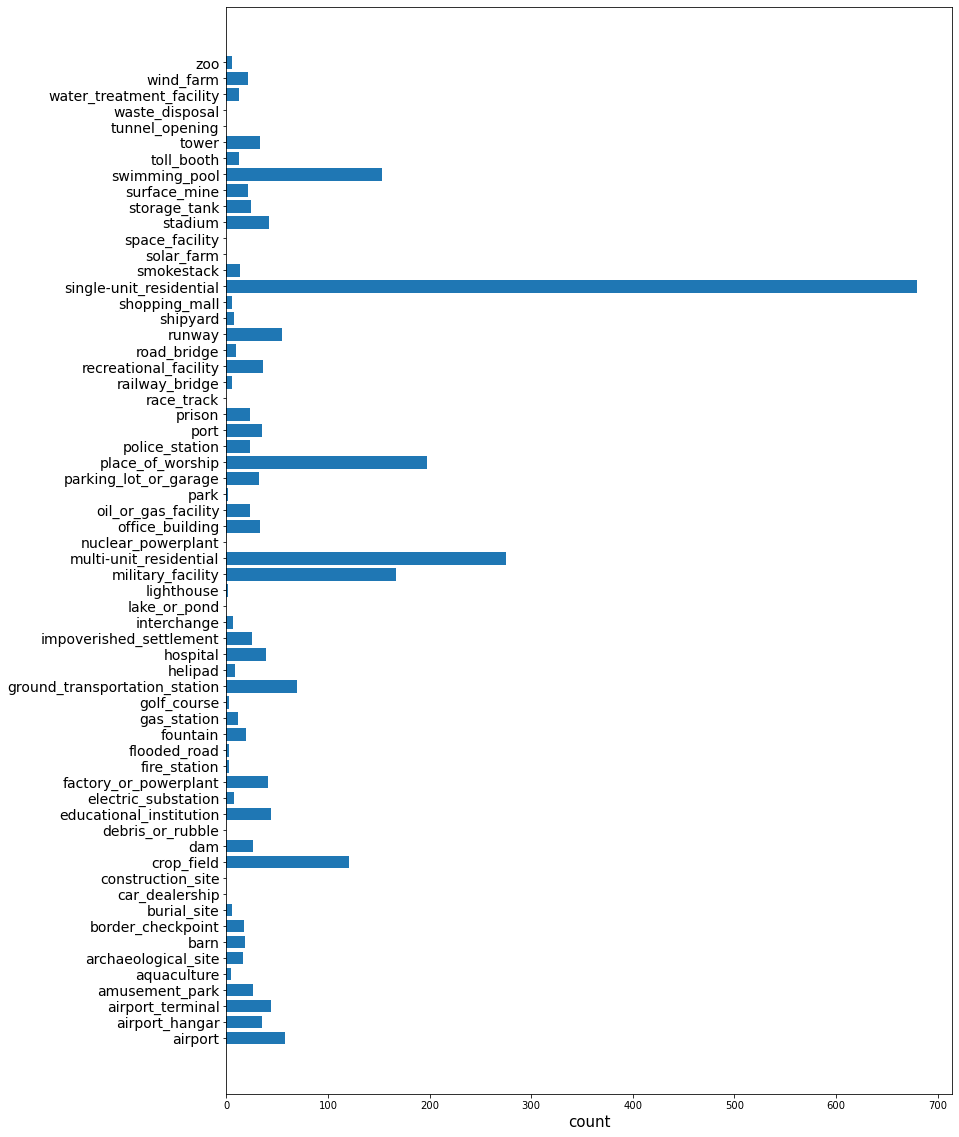}
\caption{Instances by class, test 2016-2018, Africa}
\label{fig:africa-test}
\end{figure}

\subsection{Estimator of model loss}
\label{section:h_hat}
One potential limitation of our approach is $\hat{h}$ does not always estimate
the tail losses accurately, and this is important because our approach
precisely is designed to counter ML models that perform poorly on tail
subpopulations.  Figure~\ref{fig:loss_estimator} plots a histogram of model
losses and the estimated conditional risk $\hat{h}$ for \emph{DenseNet} ERM
and \emph{CLIP WiSE}, where the y-axis is plotted on a log-scale.  It is clear
that \emph{DenseNet} ERM has more extreme losses compared to the \emph{CLIP
  WiSE} model, suggesting that at least part of the reason why \emph{DenseNet}
  ERM suffers poor loss on subpopulations: it is overly confident when it's
incorrect. While a direct comparison is not appropriate since the conditional
risk $\mu(Z)$ represent \emph{smoothed} losses, we observe that naive
estimators of $\mu(\cdot)$ may consistently underestimate. In this particular
instance, since the extent of underestimation is more severe for \emph{ImageNet}
pre-trained models, our experiments are fortuitously providing an even more
conservative comparison between the two model classes, instilling confidence
in the relative robustness of the \emph{CLIP WiSE} model.

\begin{figure}[htbp]
  \centering
  \begin{subfigure}[b]{0.48\textwidth}
    \centering
    \includegraphics[width=\textwidth]{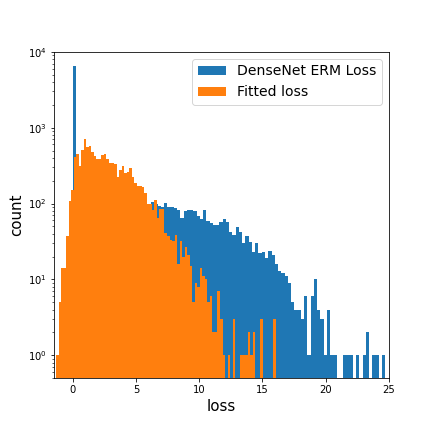}
  \end{subfigure}
  \begin{subfigure}[b]{0.48\textwidth}
    \centering
    \includegraphics[width=\textwidth]{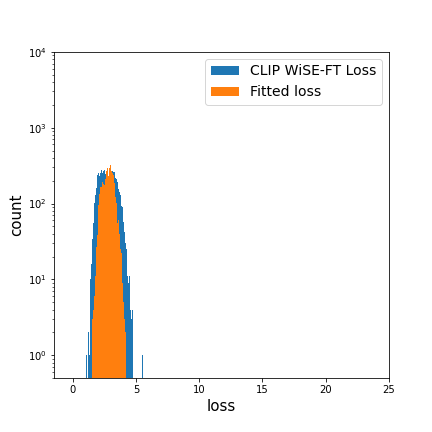}
  \end{subfigure}
  \caption{Histograms of model losses and fitted losses $\hat{h}$. Y-axis count is plotted in \textbf{log-scale}. For \emph{DenseNet} ERM model, fitted $\hat{h}$ underestimates the extreme losses (right-tail).}
  \label{fig:loss_estimator}
\end{figure}

Alternatively, we can directly define the worst-case subpopulation
performance~\eqref{eqn:cvar} using the 0-1 loss. The discrete nature of the
0-1 loss pose some challenges in estimating $\mu(\cdot)$. While we chose to
focus on the cross entropy loss that aligns with model training, we leave a
thorough study of 0-1 loss to future work.

\subsection{Additional comparisons}
\label{section:addexp}

We use the ensembled \emph{CLIP WiSE} model constructed by averaging the
network weights of \emph{CLIP zero-shot} and \emph{CLIP finetuned} models. So
far, we used proportion $\lambda = 0.4$ to match the ID validation accuracy of
\emph{CLIP WiSE} to that of \emph{DenseNet} models and
\emph{DPN}-68. In this subsection, we provide alternative choices:
\begin{enumerate}
\item $\lambda=0.24$ to match ID accuracy of \emph{ResNet}-18 of $47\%$
\item $\lambda=0.27$ to match ID accuracy of \emph{VGG}-11 of $51\%$.
\end{enumerate}
Similar to \emph{DPN}-68, \emph{ResNet}-18 and \emph{VGG}-11 are \emph{ImageNet}
pretrained models fine-tuned on FMoW as evaluated
by~\citet{MillerTaRaSaKoShLiCaSc21}. We refer to the two \emph{CLIP WiSE}
models as \emph{CLIP WiSE}-24 and \emph{CLIP WiSE}-27 respectively, and report
all model performances below. Again, we observe that our approach successfully
picks out the more robust \emph{CLIP WiSE} models, in contrast to the
non-robust models chosen by ID accuracy or ID loss.

\begin{table}[h]
\centering
\begin{tabular}{lrrr|rr}
\toprule
	& \multicolumn{3}{c|}{ID, 2002-2013} & \multicolumn{2}{c}{OOD, 2016-2018} \\
        Model &  Accuracy &  Loss &  $\mathsf{W}_{0.10} $ &  Accuracy &  Loss \\
\midrule
 \emph{CLIP WiSE}-24 &    0.47 &     2.84 &     3.47 &      0.45 &       2.85 \\
    \emph{ResNet}-18 &    0.48 &     2.84 &     5.05 &      0.40 &       3.36 \\
\midrule
\emph{CLIP WiSE}-27 &    0.51 &     3.07 &     3.54 &      0.48 &       3.08 \\
       \emph{VGG}-11 &    0.51 &     3.06 &     6.07 &      0.45 &       3.68 \\
\bottomrule
\end{tabular}
\caption{Additional experiments showcasing our approach successfully identifies more robust models.} 
\label{tbl:addexp}
\end{table}
  

\section{Proof of equivalence results}
\label{sec:proof-extension}

In this section, we discuss in detail how our notion of generalized worst-case
subpopulation performance is closely connected to distributional robustness
and coherent risk measures. The reader is recommended to refer to Section 6 of
the lectures notes by~\citet{ShapiroDeRu14} for more detail.

\subsection{Proof of Theorem~\ref{thm:wpa}}
\label{sec:proof-wpa}

To see the first claim, fix a nonempty class $\qaset$ of probability measures
on $(0,1]$. Notice $\mathsf{W}_{\qaset}(\cdot)$ is proper because $\qaset$ is
nonempty and $\mathsf{W}_{\qaset}(0) = 0 < \infty$. It is lower
semi-continuous because it is a pointwise supremum of
$\model\mapsto \int_{(0, 1]} \mathsf{W}_{\alpha}(\model) d\lambda(\alpha)$,
which are lower semi-continuous. Coherence can be shown by verifying the
definition. Clearly, $\mathsf{W}_{\qaset}$ is law-invariant because it is
defined using only $\mathsf{W}_\alpha(\cdot)$ which is law-invariant.

For the converse relation, recall the discussion preceding
Lemma~\ref{lemma:dro}. The biconjugacy relation~\eqref{eqn:biconjugacy} gives
the variational representation
\begin{equation*}
  \risk(\model) =
  \sup_{L \in \dom \risk\opt}
  \E_P[L \model].
\end{equation*}
Since $\risk$ is law-invariant, we have the tautological reformulation
\begin{equation}
  \label{eqn:risk-dual-dummy}
  \risk(\model) = \sup_{\model' \in \mathcal{L}^k} \left\{ \risk(\model'): \model' \eqd \model
  \right\}
  = \sup_{L \in \dom \risk\opt, \model' \in \mathcal{L}^k}
  \left\{
    \E_P[L \model']: \model' \eqd \model
  \right\}.
\end{equation}
Next, we use a generalization of the Hardy–Littlewood inequality.  Let
$\aq{\model}$ denote the $(1-\alpha)$-quantile of the random variable
$\model \in \mathcal{L}^k$.
\begin{lemma}[{\citet[Lemma 6.25]{ShapiroDeRu09}}]
  \label{lemma:rearrangement}
  Suppose $P$ is nonatomic.
  For $\model \in \mathcal{L}^k$ and $L\in \mathcal{L}^{k_*}$,
  \begin{equation*}
    \sup_{\model' \in \mathcal{L}^{k}} \left\{ \E_P[L \model']: \model' \eqd \model \right\}
   =  \int_0^1 P_{1-t}^{-1}(L) P_{1-t}^{-1}(\mu) dt. 
  \end{equation*}
  In particular, $\E_P[\mu] = \int_0^1 P_{1-t}^{-1}(\model) dt$.
\end{lemma}
\noindent We use the following elementary identity to rewrite
$P_{1-t}^{-1}(L)$ in the preceding display. We defer its proof to the end of
the subsection.
\begin{lemma}
  \label{lemma:quantile}  
For any random variable $L\in\mathcal{L}^{k_*}(P)$, 
  \begin{equation*}
    P_{1-t}^{-1}(L) = \int_t^1\alpha^{-1} d \qa_L(\alpha) 
    ~~\mbox{for}~~ t\in(0,1),
  \end{equation*}
  where $\qa_L(\cdot)$ is a probability distribution on $(0,1]$, if and only if 
  $L\ge0$ $P$-a.s., $\E_P[L]=1$, $\qa_L(\alpha) = \E_P \hinge{L - \aq{L}}$ 
  for $\alpha \in (0,1)$, and $\qa_L(1)=1$. 
\end{lemma}

For any fixed $L \in \mathcal{L}^{k_*}$, conclude
\begin{equation*}
\sup_{\model' \in \mathcal{L}^{k}} \left\{ \E_P[L \model']: \model' \eqd \model\right\}
=  \int_0^1 P_{1-t}^{-1}(L) P_{1-t}^{-1}(\mu) dt
= \int_0^1  \int_{(t,1]} \alpha^{-1} d\qa_L(\alpha)
  ~P_{1-t}^{-1}(\mu) dt.
\end{equation*}
Applying Fubini-Tonelli to the RHS of the preceding display, 
\begin{align*}
  \sup_{\model' \in \mathcal{L}^{k}} \left\{ \E_P[L \model']: \model' \eqd \model\right\}
 = \int_{(0,1]} \frac{1}{\alpha}\int_0^\alpha P_{1-t}^{-1}(\mu) dt  ~d\qa_L(\alpha) 
  =   \int_{(0,1]} \mathsf{W}_\alpha(\mu) d\qa_L(\alpha),
\end{align*}
where the final equality follows from the change-of-variables
reformulation~\eqref{eqn:dual}.  To obtain the
representation~\eqref{eqn:risk-dual-dummy}, we take the supremum over
$L \in \dom \risk\opt$
\begin{align*}
  \risk(\model)
  = \sup_{L \in \dom\risk\opt}
   \int_{(0,1]} \mathsf{W}_\alpha(\mu) d\qa_L(\alpha).
\end{align*}

\subsubsection{Proof of Lemma~\ref{lemma:quantile}}
\label{sec:proof-quantile}

Before proving the equivalence, we first show the following identity for any 
$L$ and $\qa_L$.
  \begin{equation}
    \label{eqn:claim}
    \int_0^\alpha [h(t)-h(\alpha)]\,\mathrm{d}t =
    \E_{P}(L-P_{1-\alpha}^{-1}(L))_+ - \qa_L(\alpha),
  \end{equation}
where we define $h(t) := P_{1-t}^{-1}(L) - \int_{(t,1]} a^{-1}\,\mathrm{d} \qa_L(a)$ 
for convenience. We separately consider the two differences in the integrand
  \begin{align*}
    \int_0^\alpha [h(t)-h(\alpha)]\,\mathrm{d}t
    =   \int_0^\alpha
    P_{1-t}^{-1}(L) - P_{1-\alpha}^{-1}(L)
    - \left(
    \int_t^1 a^{-1}\,\mathrm{d}\qa_L(a)
    -   \int_\alpha^1 a^{-1}\,\mathrm{d}\qa_L(a)
    \right)
    \,\mathrm{d} t.
  \end{align*}
  For $t\in(0,\alpha)$
  \begin{align*}
      \int_0^\alpha \left[P_{1-t}^{-1}(L) -
        P_{1-\alpha}^{-1}(L)\right]\,\mathrm{d}t
      &= \int_0^1 \left[P_{1-t}^{-1}(L) -
        P_{1-\alpha}^{-1}(L)\right]_+ \,\mathrm{d}t \\
      &= \int_0^1 P_{1-t}^{-1}\left(
        \left[L - P_{1-\alpha}^{-1}(L)\right]_+\right) \,\mathrm{d}t 
      = \E_{P} \left[L - P_{1-\alpha}^{-1}(L)\right]_+,
  \end{align*} and
  \begin{equation*}
    \int_0^\alpha \left[\int_{(t,1]} a^{-1}\,\mathrm{d} \qa_L(a) -
      \int_{(\alpha,1]} a^{-1}\,\mathrm{d} \qa_L(a) \right]\,\mathrm{d} t =
    \int_{(0,\alpha]} \left(\int_0^a \,\mathrm{d}t \right)
    a^{-1}\,\mathrm{d}\qa_L(a) = \qa_L(\alpha).
  \end{equation*}

Now we show the ``if'' part. It is clear that $\qa_L$ is nondecreasing, as $\aq$ 
is nonincreasing in $\alpha$, and $\qa_L(\alpha)\le \E_P[L]=1=\qa_L(1)$ because 
$L\ge0$ $P$-a.s. We know $\qa_L$ is right-continuous because $\aq{\cdot}$ is left continuous 
in $\alpha$. Then we conclude $\qa_L$ is a probability distribution on $(0,1]$ by 
noticing $\lim_{\alpha\downarrow0} \qa_L(\alpha) = 0$ because
\begin{align*}
  \qa_L(\alpha) = \norm{\hinge{L-\aq{L}}}_1 &\le \norm{\hinge{L-\aq{L}}}_{k_*} \\
  &\le \norm{L}_{k_*} \P\{L > \aq{L}\}^{1/k_*} \\
  &\le \norm{L}_{k_*} \alpha^{1/k_*},
\end{align*}
where the first inequality follows by H\"older's inequality, the second because 
$\hinge{L-\aq{L}} = (L-\aq{L})\textbf{1}\{L>\aq{L}\}$ and $L\ge0$ $P$-a.s., and 
the last by definition of $\aq{L}$.

By the definition of $\qa_L$, the RHS of Identity~\eqref{eqn:claim} is zero, so 
$ \alpha h(\alpha) = \int_0^\alpha h(t)\,\mathrm{d} t$. 
In particular, $h(\cdot)$ is differentiable in $(0,1)$ since
$\alpha \mapsto \int_0^\alpha h(t)\,\mathrm{d} t$ is differentiable. Taking
derivatives on both sides of the preceding display, we have
$\alpha h'(\alpha)+h(\alpha) = h(\alpha)$ and in particular, $h'(\alpha)=0$
for $\alpha\in(0,1]$.  To show that $h(t) \equiv h$ is uniformly zero, notice
\begin{equation*}
  h = 
  \int_0^1 h(t)\,\mathrm{d} t
  = \int_0^1 \left[P_{1-t}^{-1}(L)-\int_t^1 a^{-1}\,\mathrm{d}\qa_L(a)\right]\,\mathrm{d} t
  = \E_P L - 1 = 0,
\end{equation*}
where we used Fubini-Tonelli and $\qa_L(1) = 1$ in the third equality. 

Next, we show the ``only if'' part. Clearly the LHS of Identity~\eqref{eqn:claim} is zero, 
so $\qa_L(\alpha) = \E_{P}\hinge{L-\aq{L}}$. We know $L\ge0$ $P$-a.s.\ because $\qa_L\ge0$. 
Lastly, $\E_P[L] = \int_0^1 P_{1-t}^{-1}(L)dt = \int_0^1\int_t^1\alpha^{-1}d\qa_L(\alpha)dt  
= 1$ by Fubini-Tonelli.

\subsection{Proof of Proposition~\ref{prop:hcvar}}
\label{sec:proof-hcvar}

First, to see the duality result, notice the minimax theorem and H\"older's inequality imply
\begin{align*}
    \risk_k(\model) 
    &= \inf_{\eta\in\R} \{\alpha^{-1}||(\mu - \eta)_+||_k + \eta\} \\
    &= \inf_{\eta\in\R}\sup\{ \E_P [L(\model - \eta)] + \eta : L\ge0, \norm{L}_{k_*}\le \alpha^{-1}\} \\
    &= \sup \left\{ \inf_{\eta\in\R} \{ \E_P[L(\model-\eta)] + \eta\} : L \ge 0, \norm{L}_{k_*}\le \alpha^{-1} \right\} \\
    &= \sup_L \left\{ \E_P[L\model] : L \ge 0, \E_P[L]=1, \norm{L}_{k_*}\le \alpha^{-1} \right\}.
\end{align*}
This means $\dom \risk\opt_k = \{L\in\mathcal{L}^k: L\ge 0, \E_P[L]=1, \norm{L}_{k_*}\le \alpha^{-1}\}$. 
Theorem~\ref{thm:wpa} implies \[ \rho_k(\model) = \sup\left\{ \int_{(0,1]} \mathsf{W}_\alpha(\model) 
d\qa_L(\alpha): \qa_L(\alpha) = \E_P \hinge{L - \aq{L}}, L\ge0, \E_P[L]=1, 
\norm{L}_{k_*}\le\alpha^{-1} \right\}. \] 
Lemma~\ref{lemma:quantile} further implies $\{L, \qa_L: \qa_L(\alpha) = \E_P \hinge{L - \aq{L}}, 
L\ge0, \E_P[L]=1\} = \{L, \qa_L\in\Delta((0,1]): P_{1-t}^{-1}(L) = \int_t^1\alpha^{-1} d \qa_L(\alpha)\}$.
Notice $P_{1-t}^{-1}(L) = \int_t^1\alpha^{-1} d \qa_L(\alpha)$ implies \[ \norm{L}_{k_*}^{k_*} = 
\int_\Omega L^{k_*} d\P = \int_0^1 (P_{1-t}^{-1}(L))^{k_*} dt = \int_0^1 \left(\int_t^1\alpha^{-1} 
d \qa_L(\alpha) \right)^{k_*} dt,\] so
\[ \risk_k(\model) =  \sup\left\{ \int_{(0,1]} \mathsf{W}_\alpha(\model) d\qa_L(\alpha): 
\qa_L\in\qaset_k, P_{1-t}^{-1}(L) = \int_t^1\alpha^{-1} d \qa_L(\alpha) \right\}. \] 
Lastly, since $L$ can always be defined based on any $\lambda_L\in\Delta((0,1])$, we drop the 
last equality and conclude \[ \risk_k(\model) = \sup\left\{ \int_{(0,1]} \mathsf{W}_\alpha(\model) 
d\qa_L(\alpha): \qa_L\in\qaset_k \right\} = \mathsf{W}_{\qaset_k}(\model).\]

\section{Certificate of robustness}
\label{section:certificate}

 Instead of estimating the worst-case
subpopulation performance for a fixed subpopulation size $\alpha$, it may be
natural to posit a level of acceptable performance (upper bound $\bar{\loss}$
on the loss) and study $\alpha\opt$, the smallest subpopulation
size~\eqref{eqn:certificate} over which the model $\theta(\cdot)$ can
guarantee acceptable performance.
Our plug-in estimator $\what{\alpha}$ given in Eq.~\eqref{eqn:certificate-emp}
enjoys similar concentration guarantees as given above.
The following
theorem---whose proof we give in
Appendix~\ref{section:proof-certificate}---states that the true
$\alpha\opt$ is either close to our estimator $\what\alpha$ or it is
sufficiently small, certifying the robustness of the model against
subpopulation shifts.  

\begin{theorem}
  \label{theorem:certificate}
  Let Assumption~\ref{assumption:bdd} hold, let $U(\delta)>0$ be such that for
  any fixed $\alpha\in(0,1]$,
  $| \hworstsub(\what{\model}) - \worstsub(\condrisk)| \le U(\delta)/\alpha$ with
  probability at least $1-\delta$. Then given any $\underline\alpha\in(0,1]$,
  either $\alpha\opt<\underline\alpha$, or
  $$\small \left|\frac {\alpha\opt} {\what\alpha} - 1 \right| \le \frac{ U(\delta) } 
  {\what \E \left[ \what{\mu}(Z) - \what P_{1 - \underline\alpha \wedge
        \what{\alpha}} ^{-1} \left( \what{\mu}(Z) \right) \right]_+}$$ with
  probability at least $1-\delta$, where $\what\E$ and
  $\what{P}^{-1}_{1-\alpha}$ denote the expectation and the
  $(1-\alpha)$-quantile under the empirical probability measure induced by
  $I_k$.
\end{theorem}
\noindent
Our approach simultaneously provides localized Rademacher complexity bounds
and dimension-free guarantees. Our bound becomes large as
$\underline{\alpha}\to0$ and we conjecture this to be a fundamental difficulty
as the worst-case subpopulation performance~\eqref{eqn:cvar} focuses on
$\alpha$-faction of the data.

\subsection{Proof of Theorem~\ref{theorem:certificate}}
\label{section:proof-certificate}

For ease of notation, we suppress any dependence on the prediction model
$\theta(X)$ under evaluation. Consider any $\alpha_1,\alpha_2 \in(0,1]$ with
$\alpha_1<\alpha_2$. Denote by $\what\P$ the empirical probability measure induced by
$(Z_i:i\in I_k)$. For convenience denote
$\xi_1:= \what P_{1-\alpha_1}^{-1}(\what \model(Z))$ and
$\xi_2:= \what P_{1-\alpha_2}^{-1}(\what \model(Z))$, so $\xi_1\ge \xi_2$ and
$\what{\mathsf W}_{\alpha_1}(\what\model)\ge\what{\mathsf W}_{\alpha_2}(\what
h)$. Notice that
\begin{align*}
    \what\E[\what \model(Z)-\xi_2]_+ - \what \E[\what \model(Z)-\xi_1]_+ &= \what\E[\what \model(Z)-\xi_2;\what \model(Z)>\xi_2] - \what\E[\what \model(Z)-\xi_1;\what \model(Z)\ge\xi_1]\\
    &= \underbrace{\what\E[\what \model(Z)-\xi_1;\xi_2< \what \model(Z) < \xi_1]}_{\le 0} 
        + (\xi_1-\xi_2) \underbrace{\what\P(\what \model(Z)>\xi_2)}_{\le \alpha_2} \\
    &\le (\xi_1-\xi_2)\alpha_2.
\end{align*}

Hence, by Lemma~\ref{lemma:cvar-dual-optima},
\begin{align*}
  \what{\mathsf W}_{\alpha_1}(\what\model) - \what{\mathsf W}_{\alpha_2}(\what\model) 
  & = \left(\frac{\what \E[\what \model(Z)-\xi_1]_+}{\alpha_1} + \xi_1\right) - \left(\frac{\what \E[\what \model(Z)-\xi_2]_+}{\alpha_2}+\xi_2\right) \\
    & \ge \frac{\what \E[\what \model(Z)-\xi_1]_+}{\alpha_1\alpha_2}(\alpha_2-\alpha_1),
\end{align*}
meaning
$$|\alpha_1-\alpha_2| \le \frac{\alpha_1\alpha_2 |\what{\mathsf W}_{\alpha_1}(\what\model) - \what{\mathsf W}_{\alpha_2}(\what\model)|} {\what\E[\what \model(Z)-\what P^{-1}_{1-\alpha_1}(\what \model(Z))]_+}.$$

Now suppose $\alpha\opt\ge\underline{\alpha}$. Notice $\worstsub$ and $\hworstsub$ are continuous and nonincreasing in $\alpha$, so the definitions~\eqref{eqn:certificate} and~\eqref{eqn:certificate-emp} imply
$\mathsf W_{\alpha\opt}(\condrisk) = \bar\loss = \what{\mathsf{W}}_{\what\alpha}(\what\model)$. Plugging $\what\alpha$ and $\alpha\opt$ into the inequality above, we know with probability at least $1-\delta$,
\begin{equation*}
|\alpha^\star-\what\alpha| 
\le 
\frac{\what\alpha \alpha^\star \left|\what{\mathsf W}_{\alpha^\star}(\what\model) - {\mathsf W}_{\alpha\opt}(\condrisk)\right|}
{\what \E \left[ \what \model(Z) - \what P_{1 - \alpha\opt \wedge \what\alpha}
^{-1} \left( \what \model(Z) \right) \right]_+}
\le
\frac{\what\alpha U(\delta)}
{\what \E \left[ \what \model(Z) - \what P_{1 - \underline\alpha \wedge \what\alpha}
^{-1} \left( \what \model(Z) \right) \right]_+}.
\end{equation*}


\end{APPENDICES}
\else
\newpage
\appendix

\fi

\end{document}